\documentclass[runningheads]{llncs}

\usepackage{eccv}

\usepackage{svg}
\usepackage{eccvabbrv}
\usepackage{epsfig}
\usepackage{graphicx}
\usepackage{amsmath}
\usepackage{amssymb}
\usepackage{xfrac}
\usepackage[mathscr]{euscript}
\usepackage{appendix}
\usepackage{algorithm}
\usepackage{algpseudocode}
\usepackage{cancel}
\usepackage{caption} 
\usepackage{color,soul}
\usepackage{tabularx}
\usepackage{subcaption}
\usepackage{mathtools}
\usepackage{paralist} % compactenum
\usepackage{colortbl}
\usepackage{multirow}
\usepackage{graphicx}
\usepackage{booktabs}
\usepackage[algo2e]{algorithm2e}
\usepackage{amssymb}% http://ctan.org/pkg/amssymb
\usepackage{pifont}% http://ctan.org/pkg/pifont
\newcommand{\cmark}{\ding{51}}%
\newcommand{\xmark}{\ding{55}}%

\usepackage[accsupp]{axessibility}  % Improves PDF readability for those with disabilities.
\usepackage[pagebackref,breaklinks,colorlinks,citecolor=eccvblue]{hyperref}
\usepackage{orcidlink}
\usepackage{varwidth}
\usepackage{wrapfig,lipsum,booktabs}

\definecolor{yellow}{rgb}{1, 1, 0.7}
\definecolor{orange}{rgb}{1, 0.85, 0.7}
\definecolor{red}{rgb}{1, 0.7, 0.7}

\definecolor{lightyellow}{rgb}{1,1, 0.8}

\definecolor{wincolor}{rgb}{0.85, 0.0, 0.0}
\definecolor{darkyellow}{rgb}{0.8, 0.8, 0.5}
\definecolor{darkred}{rgb}{0.7, 0.3, 0.3}
\definecolor{darkgreen}{rgb}{0.3, 0.7, 0.3}
\definecolor{blue}{rgb}{0, 0, 1.0}
\definecolor{green}{rgb}{0, 1.0, 0}
\definecolor{pink}{rgb}{1, 0.4, 0.7}

\newcommand{\Mat}{\boldsymbol}

\begin{document}

\title{GAURA: Generalizable Approach for Unified Restoration and Rendering of Arbitrary Views} 
\titlerunning{GAURA: Generalizable Approach for Unified Restoration and Rendering}

% \author{Vinayak Gupta\inst{1}\orcidlink{0000-1111-2222-3333} \and
% Second Author\inst{2,3}\orcidlink{1111-2222-3333-4444} \and
% Third Author\inst{3}\orcidlink{2222--3333-4444-5555}}

\author {
    % Authors
    Vinayak Gupta\textsuperscript{\rm 1}$^*$,
    Rongali Simhachala Venkata Girish\textsuperscript{\rm 1}$^*$,
    Mukund Varma T\textsuperscript{\rm 2}$^*$,
    Ayush Tewari\textsuperscript{\rm 3},
    Kaushik Mitra\textsuperscript{\rm 1}
}
\authorrunning{V. Gupta, R. Girish, M. Varma T et al.}

\institute{\textsuperscript{1}Indian Institute of Technology Madras \enspace 
\textsuperscript{2}University of California San Diego\\
\textsuperscript{3}Massachusetts Institute of Technology
}

\maketitle

\vspace{-5mm}
\begin{figure*}[!ht]
  \includegraphics[width=\textwidth]{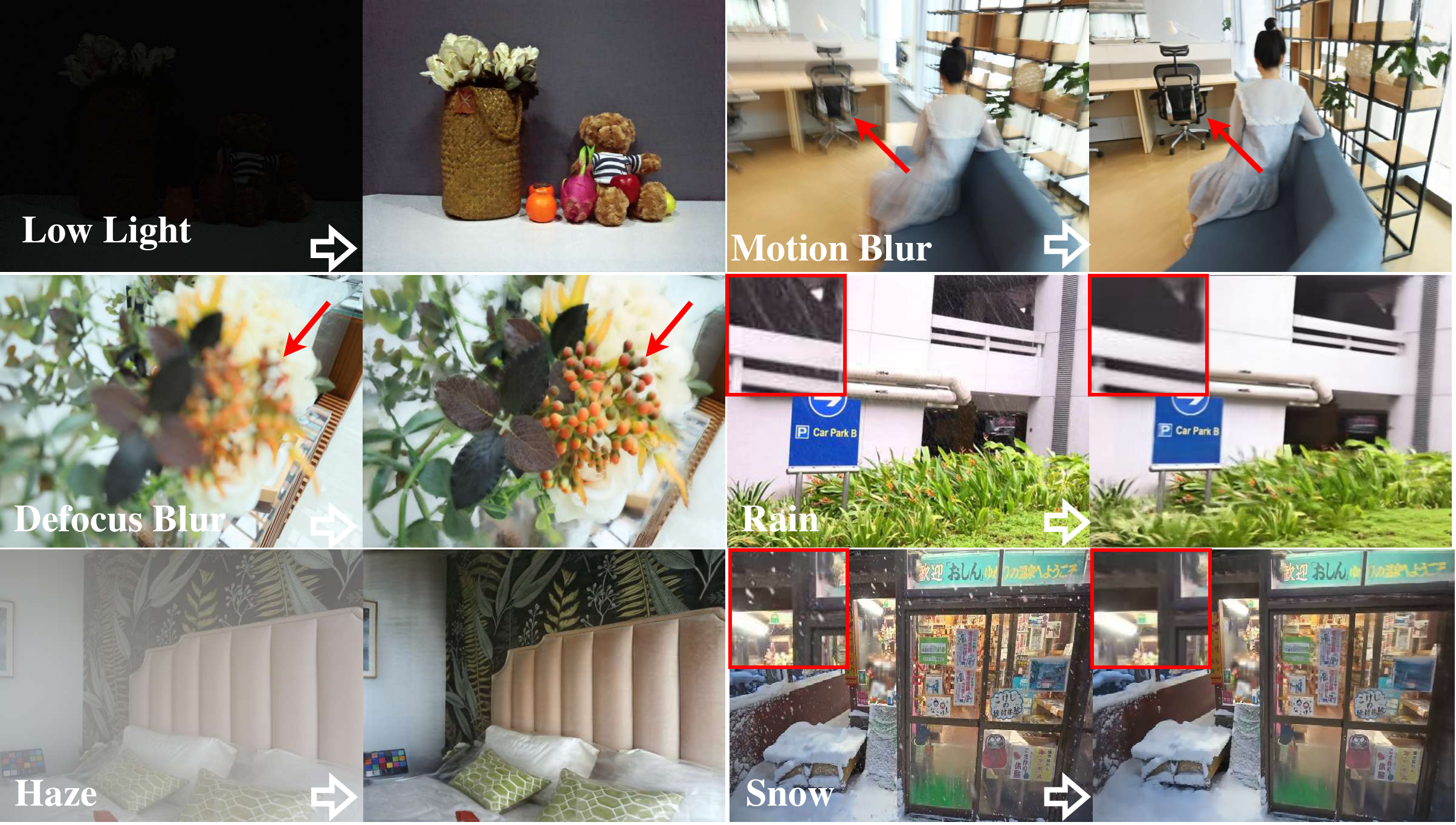}
  \caption{
  Reconstructing 3D scenes from imperfect image captures is highly desirable, e.g., in low light, but is challenging. 
  To address this, we introduce \textit{GAURA}, a technique designed to render and restore novel views from degraded input views.
  Unlike previous attempts at this problem, we demonstrate generalization to different scenes and degradation types. 
  In each example, we visualize the imperfect target capture (on the left) and its corresponding clear rendered view (on the right). 
  % \textcolor{blue}{For example, in the defocus deblurring task, we see that the flower is more visible. In the motion deblurring task, the blurred chair is sharper and in the deraining and desnowing tasks, the rain and snow particles have been removed}
  Please note that the imperfect view from the target viewing angle is not used as input for our method, and we simply visualize the same for simplicity. \textit{GAURA} faithfully recovers the underlying 3D scene with high geometric accuracy while still managing to generalize across several degradation types. 
  % Reconstructing 3D scenes from degraded captures often observed in the real world is challening. To address this, we introduce GAURA, a framework designed to simultaneously render and restore novel views. GAURA aims to learn a unified 3D restoration model, capable of restoring multiple corruptions in a single framework. Our method doesn't require scene-specific training, enabling on-the-fly restoration of unseen 3D scenes. We demonstrate that GAURA can successfully restore 3D scenes while maintaining the geometric consistency across views. For all the examples above, the left image represents the degraded input and the right results are produced by our single unified model.
  }

  % \caption{We propose GAURA, a method for consistent 3D restoration given degraded multi-view images. Our approach can render and restore novel views on-the-fly without per-scene training and across various degradations in a single framework. For all examples above, the left image corresponds to the degraded input and the right results are produced by our single unified model.}
  \label{fig:teaser}
  \vspace{-3em}
\end{figure*}

\begin{abstract}
Neural rendering methods can achieve near-photorealistic image synthesis of scenes from posed input images. However, when the images are imperfect, e.g., captured in very low-light conditions, state-of-the-art methods fail to reconstruct high-quality 3D scenes. Recent approaches have tried to address this limitation by modeling various degradation processes in the image formation model; however, this limits them to specific image degradations. In this paper, we propose a generalizable neural rendering method that can perform high-fidelity novel view synthesis under several degradations. Our method, \textit{GAURA}, is learning-based and does not require any test-time scene-specific optimization. It is trained on a synthetic dataset that includes several degradation types.
GAURA outperforms state-of-the-art methods on several benchmarks for low-light enhancement, dehazing, deraining, and on-par for motion deblurring. Further, our model can be efficiently fine-tuned to any new incoming degradation using minimal data. We thus demonstrate adaptation results on two unseen degradations, desnowing and removing defocus blur. Code and video results are available at \href{https://vinayak-vg.github.io/GAURA/}{vinayak-vg.github.io/GAURA}.
  % \keywords{\textcolor{blue}{Neural Radiance Fields \and 3D Restoration \and Generalisable Rendering}}
\end{abstract}
\vspace{-1em}
\section{Introduction}
\label{sec:intro}

Recent progress in Neural Radiance Fields (NeRF)~\cite{mildenhall2020nerf} and other implicit representations~\cite{chen2019learning,genova2020local,niemeyer2020differentiable,sitzmann2019scene} has enabled the transformation of multi-view captures of real-world scenes into 3D models that produce photo-realistic and consistent renderings from novel viewpoints. 
However, methods like NeRF still require images captured in perfect conditions, which is often hard to achieve in the real world. 
Imperfections in the captured images can either arise from challenging scenes, e.g., lack of sufficient illumination, or from the capturing device, e.g., motion or defocus blur from commodity hand-held devices. 
%
% These imperfections can violate the multi-view consistency of training images, resulting in poor rendering quality from novel viewpoints.
These imperfections violate the image formation model of neural radiance fields, resulting in poor rendering quality. 

Existing methods for novel view synthesis from degraded image captures, typically modify the image formation process by modeling the physical degradation. 
As examples, Seathru-NeRF \cite{levy2023seathru} incorporates the underwater image formation model into their rendering process, and Deblur-NeRF \cite{ma2022deblur} leverages a kernel at each spatial location to simulate blurring. 
%
% and hence recover clean images from their corresponding degraded versions. 
% , they typically require a highly creative method 
% \AT{I would remove highly creative and just mention that they are are specialized and not general. } 
However, explicitly modeling the physical degradation process results in more challenging inverse problems and limits these approaches to very specific degradations, limiting practical applications of these methods in the real world. 
%

% This results in specialization of these models to specific degradation types \cite{cui_aleth_nerf,ma2022deblur,chen2023dehazenerf}. 
% \AT{More than violating multi-view consistency, the main point should be that volume rendering in NeRF does not account for any such physical degradation process and can thus not reconstruct accurate 3D scenes in their presence.}
% In this paper, we ask: is it possible to account for these imperfections in the input image captures, and simultaneously restore them while rendering novel viewpoints?
% \AT{I see--I got confused reading this paragraph. It only talks about NeRF and other methods that are optimization-based and are not generalizable. So the concept of training images is unexpected. My earlier comment was based on this reading that you cannot optimize a NeRF from degraded images. Perhaps, it is better to first introduce generalizable NeRFs and then talk about the distribution shift.}
% \AT{The generalization vs. per-scene optimization point is still confusing. Can we discuss them separately? We can mention that (1) Per-scene optimization requires modeling the degradation process and it is difficult to model all possible degradations, and (2) learning-based methods would avoid this problem (inspired by 2D methods). Are we the first learning-based method in this setting? If so, we can claim it upfront.}

In this paper, we propose a novel method that \textit{learns} to reconstruct scenes under any general \textit{real-world} degradation, enabling the rendering of clean images from arbitrary viewpoints. 
We take inspiration from state-of-the-art 2D image restoration (IR) methods that leverage large networks trained on abundant paired data, i.e., degraded and their corresponding clean image pairs captured from the real world or generated synthetically\cite{li2020all,potlapalli2023promptir,airnet}. 

Our learning-based reconstruction method builds on the progress in generalizable NeRF-based scene reconstruction  \cite{wang2021ibrnet,wang2022generalizable,gnt}. 
Unlike per-scene optimization approaches \cite{mildenhall2020nerf}, generalizable methods can reconstruct scenes from sparse input images using feed-forward network operations, which can be used to render images from novel viewpoints.
%
% More recently, generalizable NeRF (Gen-NeRF) methods\cite{wang2021ibrnet,wang2022generalizable,gnt} have become increasingly competitive, achieving performance similar to per-scene optimized techniques and can bake the underlying rendering prior into large networks e.g., transformers\cite{vaswani2017attention}, when trained on sufficiently large datasets. 
% These generalizable methods learn to render novel views from sparse input images while strictly following epipolar geometry. 
We extend the idea of generalization across scenes to generalization across degradations as well. 
We call our technique GAURA: \textit{Generalizable Approach for Unified Restoration and Rendering of Arbitrary Views}, a model capable of restoring many different degradations. 
% , by implicitly encoding several degradation-specific rendering models onto a single Gen-NeRF network, but achieving this without any additional prior can be challenging. 
% Drawing inspiration from these observations, we could implicitly encode several degradation-specific rendering models onto a single Gen-NeRF network trained on such paired multi-view data but can be challenging with no additional prior. 

We base our method on the state-of-the-art generalizable model GNT \cite{gnt}, which leverages two transformers - \textit{view transformer} for epipolar feature aggregation and \textit{ray transformer} for learned ray-based rendering.  
We propose to condition the epipolar feature aggregation and learned rendering on degradation-aware latent codes.
These codes are learnable parameters that encode discriminative information about different image degradations, allowing the network to dynamically adapt its behavior to different degradations. 
%
% We base our method on the SOTA Gen-NeRF technique GNT\cite{gnt}, which leverages two transformers - \textit{view transformer} for epipolar feature aggregation and \textit{ray transformer} for learned ray-based rendering.  
% \AT{It might be possible to restructure it a little, such that generalizable NeRFs are already discussed before, along with their limitations (they can't deal with degradations). This paragraph could then be worded as we address this limitation in these models. }
GAURA casts a ray for each pixel in the target image plane, samples, and project points onto the degraded input views to fetch RGB features based on epipolar correspondences. 
The view transformer estimates coordinate-wise features from epipolar lines, which are additionally now conditioned on the input degradation type. 
For rendering a clean novel view, the ray transformer composes these point features to color while simultaneously incorporating the input degradation prior onto its learned rendering process. 
Our key insight is that our degradation-aware latent codes can interact with the input features and dynamically adjust their representations to adapt the restoration and rendering process for the relevant degradation.
The entire pipeline is trained end-to-end on degraded and clean image pairs of several degradation types, synthetically generated from existing multi-view datasets\cite{wang2021ibrnet,mildenhall2020nerf,mildenhall2019llff}. 

Even though GAURA is trained on synthetic data, its performance transfers to real-world scenes, where it can simultaneously render and restore novel views from arbitrary viewpoints. 
Further, our method can be easily tuned to any unseen degradation with very little paired data (as little as eight scenes) by adapting the learned scene and rendering priors.
On various rendering and restoration benchmarks like low-light enhancement, dehazing, and motion deblurring, the performance of our generalizable restoration technique is comparable and sometimes even better than methods that use per-scene optimization and have been designed for the specific degradation type. 
In other degradation types, like deraining and desnowing, we present the first 3D restoration technique to model these 2 corruptions effectively. 

In summary, by incorporating learnable degradation latent codes in both the epipolar feature aggregation and rendering steps, GAURA can universally render and restore novel views from degraded input views and demonstrates versatility to several degradation types. 
Such a data-driven pipeline enables zero-shot inference on any new scene containing real-world degradations from any degradation types seen during training without any extra optimization.
Further, the useful priors learned by the trained model can help adapt GAURA to any unseen degradation using a fast and efficient fine-tuning strategy. 
To the best of our knowledge, our method is the first that does not need to deal with each degradation type separately in a 3D setting.

% In summary, GAURA can universally render and restore novel views from degraded input views and demonstrates versatility to several degradation types. 
% Our technical contributions are:
% \begin{itemize}
%     \item A degradation latent module, a plug-in module incorporated in the feature extraction, aggregation and in the rendering steps to facilitate the restoration process across multiple degradation.
%     \item A data-driven strategy for 3D restoration which allows scalability to accommodate several degradation without requiring manual crafting for each degradation.
%     \item An adaptive module which encodes the intensity of degradation, allowing the model to adapt to various intensities enabling generalisation to real-world scenes. 
%     \item An efficent finetuning strategy, which allows us to adapt to new degradation quickly and efficiently with few data samples.
%     \item We contribute a view synthesis dataset containing 6 scenes for restoration of rain and snow degradations in a 3D setting.
%     \item Our comprehensive experiments demonstrate the dynamic adaptation behavior of GAURA
% by achieving state-of-the-art performance on various image restoration tasks, including
% low-light enhancement, motion-deblurring and dehazing 
    
% \end{itemize}

\vspace{-1em}
\section{Related Works}
\label{sec:introduction}
\vspace{-0.6em}
\paragraph{\textbf{2D Image Restoration.} }
Image Restoration (IR), a long-standing problem in computer vision, attempts to reconstruct a high-quality, clean image from a degraded observation. The 2D image restoration domain boasts of several benchmark datasets~\cite{yang2017deep,Chen2018Retinex,rim_2020_ECCV,zhang2021learning}, and performant methods for individual corruption types, e.g., for deblurring\cite{cho2021rethinking,liu2018learning}, low-light enhancement\cite{wang2022low,ma2022toward}, dehazing~\cite{zhang2018densely,li2021you} and many others~\cite{DerainRLNet2021, tian2020image, aharon2023hypernetwork}.
In contrast to these degradation-specific techniques, a recent line of work~\cite{airnet,potlapalli2023promptir,daclip} tries to build a single unified model that can generalize to several input imperfections and has showcased promising results. 
Naively applying these techniques to 3D data, i.e., multi-view images, yields inconsistent predictions as they are originally only intended for single-image restoration.

\vspace{-0.6em}
\paragraph{\textbf{Novel View Synthesis.} }
A pioneering work, Neural Radiance Field or NeRF~\cite{mildenhall2020nerf} 
synthesizes highly realistic and consistent novel views by fitting each scene as a continuous 5D radiance field parameterized by an MLP. 
Subsequent works have improved NeRF's rendering quality~\cite{barron2022mip, Wizadwongsa2021NeX, wang2021neus, verbin2021ref}. 
Recently, there have been efforts to generalize NeRFs to arbitrary scenes by learning priors from large-scale multi-view image datasets~\cite{wang2021ibrnet, suhail2022generalizable, gupta2024gsn}.
However, all these methods are rarely applied to input images captured in imperfect conditions (e.g., lack of sufficient illumination, motion blur, etc), thereby violating the image formation model and resulting in poor novel view synthesis quality. 
In this paper, we attempt to reconstruct a 3D scene under any real-world degradation and render clean images from arbitrary viewpoints without any additional optimization. 

% Several studies have investigated the possibility of training NeRF using non-ideal input. For instance, \cite{lin2021barf, wang2021nerf, meng2021gnerf} have attempted to train NeRF without camera poses. To tackle NeRF training under uncontrolled, in-the-wild photographs, \cite{martin2021nerf, chen2022hallucinated} introduces several extensions to NeRF that effectively model the inconsistent appearance variations and transient objects across views. However, these methods lack generalization across scenes and require per-scene training. We discuss generalized neural rendering in Sec. \ref{sec:preliminary}.

% For instance, Mip-NeRF\cite{barron2021mip, barron2022mip} addresses the issue of object scales in unbounded scenes, while Nex~\cite{Wizadwongsa2021NeX} tackles large view-dependent effects. Other works, such as \cite{oechsle2021unisurf, yariv2021volume, wang2021neus} improve surface representation, extend to dynamic scenes ~\cite{park2021nerfies, park2021hypernerf, pumarola2021d}, and introduce lighting and reflection modeling \cite{chen2021nerv, verbin2021ref}. However, these methods lack generalization across scenes and require per-scene training.
 
% \textbf{Generalizable Neural Radiance Fields.}
\vspace{-0.6em}
\paragraph{\textbf{3D Restoration. } }
Some recent efforts have attempted to reconstruct a radiance field using imperfect image captures, e.g., from blurry~\cite{ma2022deblur, wang2023badnerf}, and low-light corruptions~\cite{wang2023lighting, cui_aleth_nerf,Mildenhall_2022_CVPR}. 
These methods explicitly simulate the degradation process by modifying the rendering process in NeRF. 
This requires significant creativity, is challenging for more complex imperfections, and limits approaches to a single degradation type, affecting the practical applications of these techniques in the real world. 
In contrast, our method GAURA implicitly encodes the degradation process into learned modules, enabling generalization - to both scenes and corruption types. 

\vspace{-1em}
\section{Preliminary: Generalizable Novel View Synthesis}
\label{sec:preliminary}

We formulate our technique around Generalizable Novel View Synthesis (GNVS) to render clean images from imperfect input views and introduce the background below.
Given $N$ calibrated input (or source) views with known pose information $\{\Mat{I}_{i}, \Mat{P}_{i}\}_{i=1}^{N}$, the goal of GNVS is to synthesize a target novel view $\Mat{I}_{T}$ based on these source views, even for scenes not observed during training, thereby achieving generalizability. 
These methods first extract deep convolutional features $\Mat{F}_{i} = \mathcal{F}_{conv}(\Mat{I}_{i})$ from each input view. 
To render a target view, several rays are cast into the scene, and $K$ points $\{ \Mat{r}(t_{i})\}_{i=1}^{K}$ are sampled along each ray $\Mat{r}(t) = \Mat{o} + t\Mat{d}$, where $\Mat{o}$ is the camera center and $\Mat{d}$ is the ray's unit direction vector. 
Each point is then projected onto source view $\Mat{I}_{i}$ using projection function $\Pi_{i}$, and the nearest feature in the image plane is queried. 
These multi-view features are aggregated into point feature $\Mat{f}$ as follows:
\begin{equation}
    \Mat{f}(t) = \mathcal{F}_{view}(\{\Mat{F}_{\Pi_{i}(\Mat{r}(t))}\}_{i=1}^{N})
\end{equation}
where $\mathcal{F}_{view}$ is a permutation-invariant aggregation mapping that learns to be occlusion-aware to combine epipolar features.
Finally, these point features are accumulated into color $\Mat{c}$ using aggregation function $\mathcal{F}_{point}$:
\begin{equation}
    \Mat{c}(r) = \mathcal{F}_{point}(\{\Mat{f}(t_{i})\}_{i=1}^{K})
\end{equation}
Early works like IBRNet~\cite{wang2021ibrnet} adopt simple MLPs and an explicit volume rendering as aggregation functions $\mathcal{F}_{view}$ and $\mathcal{F}_{point}$ respectively, while more recent techniques~\cite{suhail2022generalizable, gnt, johari2022geonerf} use complex transformer mechanisms due to their stronger representation capacity. 
In this work, we adapt GNT~\cite{gnt} that specifically employs a \textit{view transformer} to aggregate epipolar features and a \textit{ray transformer} for ``attention-based'' rendering. 

\begin{figure*}[!ht]
  \centering
  \includegraphics[width=\textwidth]{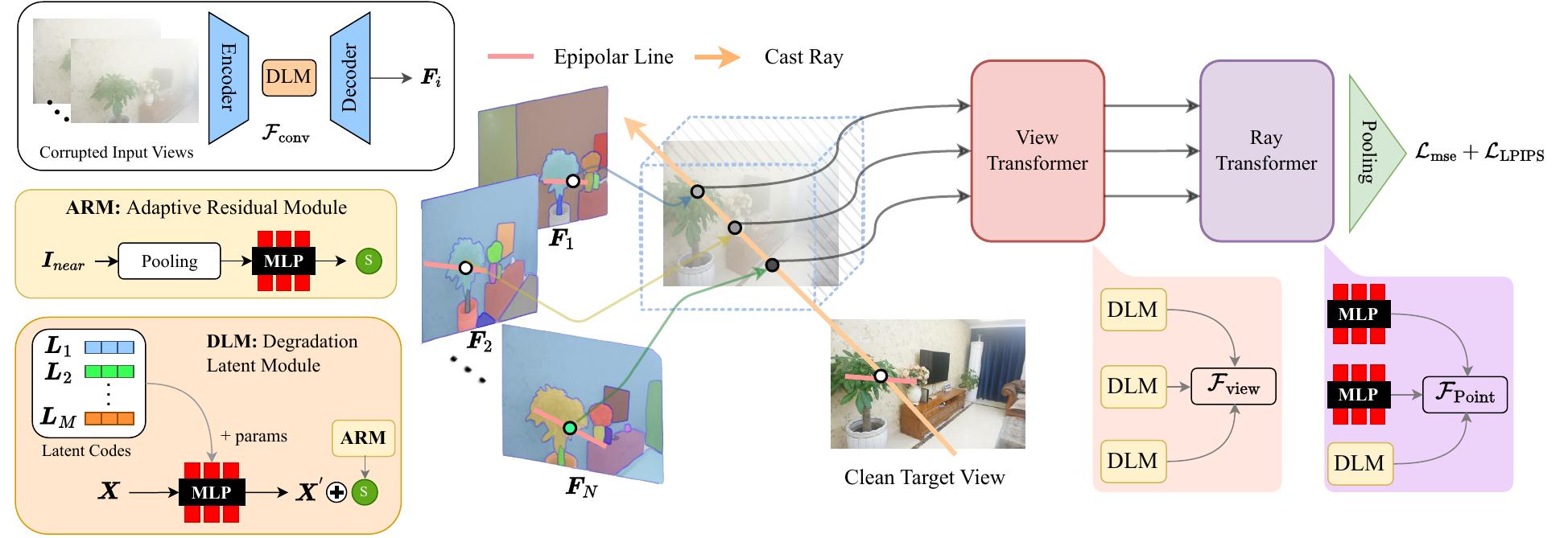}
  \caption{Overview of \textit{GAURA}: 1) Given multi-view images of a scene captured in imperfect conditions, we first extract deep convolutional features for each input view, 2) Using the source view features, we estimate the target clean rendered view via an extended epipolar-based rendering pipeline conditioned on the input imperfection type, 3) By supervising this pipeline end-to-end on paired synthetic data, our degradation-specific latent codes encode discriminative information about each imperfection type and dynamically adapts the rendering modules based on the input corruption type. Our method can directly generalize to any new scene containing real-world degradations during inference. 
  % a set of clean multi-view images, we first corrupt these images with a randomly chosen degradation type and intensity at each iteration, 2) Subsequently, we employ a degradation-aware feature extractor conditioned on input degradation-label to encode feature maps for each degraded view independently, 3) Using the source view features, we estimate the target view restored pixel colour via a series of View and Ray transformers, embedded with the Degradation Latent Module that allows us to correct imperfections and fuse degradation information while aggregating features along views and ray. 
  % 4) A Adaptive Residual is utilised to explicitly learn the variation levels in the input images, making the model robust to real-world degradations.
  }
  \label{fig:arch}
  \vspace{-4em}
\end{figure*}
\vspace{1em}
\section{Method}
\label{sec:overview}
\vspace{-0.7em}
\paragraph{\textbf{Overview. }} 
We introduce \textit{GAURA}, a novel framework to perform generalizable novel view synthesis from imperfect input captures, as illustrated in Fig.~\ref{fig:arch}. 
Formally, given $N$ calibrated multi-view images of a scene with known camera poses $\{\Mat{I}_{i}, \Mat{P}_{i}\}_{i=1}^{N}$ captured in an imperfect environment containing a given degradation type $\Mat{D}$\footnote{we assume a single degradation type for simplicity, while in practice it could be a combination of multiple (see Sec.~\ref{sec:multi-corruption})}, our goal is to synthesize clear novel views from arbitrary angles and simultaneously generalize to different scenes and corruption types. 
Our method builds upon existing GNVS technique~\cite{gnt} but conditions the scene representation and rendering processes upon some learned degradation-aware latent codes (Sec.~\ref{sec:degradationmodule}, ~\ref{sec:adaptiveres}). 
These latent codes encode discriminative information about each degradation type and dynamically adapt the remaining parts of the network based on the input imperfection (Sec. \ref{sec:degradationnvs}). 
Such a pipeline can be trained end-to-end on manually altered scenes and directly transferred to any real scene without any additional optimization (Sec. \ref{sec:trainandinf}). 
Our degradation-aware latents are fully disentangled from the remaining parts of the network, implying that they can be easily expanded for any new imperfection type and therefore be fine-tuned with minimal training (Sec. \ref{sec:trainandinf}). 
This further underscores the practical applicability of our technique, i.e., GAURA acts as an \textit{all-in-one unified restoration and rendering pipeline}. 
% \textcolor{blue}{all-in-one or generalizable?}

\subsection{Degradation-aware Latent Module}
\label{sec:degradationmodule}
Existing work for novel view synthesis and restoration often explicitly models the physical degradation process into the rendering equation~\cite{cui_aleth_nerf, ma2022deblur}. 
Although the image formation process in imperfect environments is well-studied~\cite{koschmieder1924theorie, land1977retinex, sun2013edge}, it still requires significant creativity to tailor the rendering process for each corruption type and forces no generalization to different degradation types. 
In this work, we take an alternate route and attempt to implicitly model the image formation process under several corruption types (assuming $M$ different corruption types) into learnable parameters.
% , specifically $M$ latent codes $\{\Mat{L}_{i}\}_{i=1}^{M}$. 
A naive strategy would be to create $M$ different network clones and individually allow each to model the rendering process under different degradation types, but this is inefficient. 
Moreover, the image formation processes under different imperfect environments sometimes share several similarities~\cite{cheng2023drm}. Motivated by this observation, we encode degradation-specific information into $M$ latent codes $\{\Mat{L}_{i}\}_{i=1}^{M}$ that interact with the remaining parts of the network whose parameters are shared irrespective of the input imperfection type. 
Drawing inspiration from HyperNetworks~\cite{ha2016hypernetworks}, these latent codes are mapped to weight $\Mat{W}$ of an MLP that performs a degradation-specific transformation on the input data $\Mat{X}$. 
This module, dubbed \textit{Degradation-aware Latent Module} (or DLM) can be formally described as:
\begin{equation}
\label{eq:dlm}
    \textit{DLM}(\Mat{X}, \Mat{D}) = W \cdot \Mat{X}\text{, where } W = \mathcal{F}_{\text{latent}}(\Mat{L}_{\Mat{D}} | \{\Mat{L}_{i}\}_{i=1}^{M})
\end{equation}
where $\mathcal{F}_{\text{latent}}$ denotes a fully-connected layer, and $\Mat{L}_{\Mat{D}}$ represents the latent code for the degradation type $\Mat{D}$. 
Such a technique is efficient, as the latent codes contribute to a small portion of extra parameters while still ensuring sufficient capacity to represent individual imperfection types. 

\subsection{Adaptive Residual Module}
\label{sec:adaptiveres}
Although we use a set of degradation-aware latents to encode imperfection-specific information, they are independent of the actual input captures. 
We argue that several variations exist within each corruption type, and the previous static approach might yield rather sub-optimal results. 
Therefore, we propose to add a residual feature $\Mat{S}$ obtained from the input view closest to the target view to be rendered and reformulate Eq. \ref{eq:dlm} to:
\begin{equation}
    \textit{DLM}_{\text{w/ residue}}(\Mat{X}, \Mat{D}) = \textit{DLM}(\Mat{X}, \Mat{D}) + \Mat{S}\text{, where } \Mat{S} = \textit{pool}(\mathcal{F}_{\text{residue}}(\Mat{I}_{\text{nearest}}))
\end{equation}
where $\textit{pool}(.)$ denotes the global average pooling operation, $\mathcal{F}_{\text{residue}}$ denotes a tiny convolutional network that encodes the input view closest to the target viewing angle $\Mat{I}_{\text{nearest}}$. 
In the remainder of this paper, we refer to the DLM module with \textit{Adaptive Residual Module} (ARM) as simply DLM for the sake of simplicity unless stated otherwise. 
\vspace{-1em}

\subsection{Degradation-aware Generalizable Novel View Synthesis}
\label{sec:degradationnvs}

We formulate our method around the recent SOTA GNVS technique - GNT~\cite{gnt}, that follows a three-stage pipeline as highlighted in Sec. \ref{sec:preliminary}. 
More specifically, GNT leverages a UNet-based feature extractor ($\mathcal{F}_{\text{conv}}$) to obtain geometry and appearance features from input images that are aggregated into point features using a view transformer ($\mathcal{F}_{\text{view}}$) and then accumulated along the ray to estimate ray color using the ray transformer ($\mathcal{F}_{\text{point}}$). 
This section details the different strategies employed to augment these components with degradation-specific priors that constitute \textit{GAURA}. 

The first key challenge is extracting meaningful features from imperfect input views that can later be used to embed a scene representation and render a clean novel view. 
Therefore, we incorporate the DLM blocks in the feature UNet, specifically at each decoder level, before increasing the spatial resolution of the intermediate feature maps. 
At each level, the DLM implicitly enriches the input features with information about the input degradation for guided recovery. 
To obtain point feature $\Mat{F}(t)$ at the $t$-th point sampled along the ray, we aggregate features along the epipolar line using the view transformer. 
However, this is not trivial since the features on the image plane are not necessarily multi-view consistent due to imperfections in the input captures and need to be rectified. 
Following our recurrent theme of implicitly capturing such imperfection-specific aggregation processes, we replace the vanilla MLP query, key, and value feature mapping with DLM modules and, by doing so, incorporate a degradation prior onto the scene representation phase. 
Finally, the ray transformer accumulates these point features along the ray and then decodes them into color. 
% \textcolor{blue}{I was just thinking that here you can also say that since Ray Transformer aggregates information across points in the ray, the aggregator needs to have degradation awareness for better restoration like in the case of haze, rain and snow. It is kind of important for the ray transformer to skip these particles during rendering process and just render the important objects}
By viewing the attention scores as blending weights and values as color features, such a learned aggregation represents a "generalized" volume rendering process. 
% \textcolor{blue}{Don't you think the above statement came in without any context? Like suddenly it looks like from where did attention scores, blending weights come from? }
Therefore, we only replace the value feature mapping with DLM modules in the ray transformer since the underlying scene geometry represented as attention scores obtained from the query and key features are independent of input imperfection type. 
\begin{equation}
\begin{aligned}
    \Mat{F}_{i} = \mathcal{F}_{\text{dec}}(\mathcal{F}_{\text{enc}}(\Mat{I}_{i}), \textit{DLM}(\Mat{I}_{i}, D))\text{, where } \mathcal{F}_{\text{dec}} \circ \mathcal{F}_{\text{enc}} := \mathcal{F}_{\text{conv}}  \\
    \Mat{f}(t) = \mathcal{F}_{view}(\{\textit{DLM}(\Mat{G})_{\Mat{q}}, \textit{DLM}(\Mat{G})_{\Mat{k}}, \textit{DLM}(\Mat{G})_{\Mat{v}}\})\text{, where } \Mat{G} = \{\Mat{F}_{\Pi_{i}(\Mat{r}(t))}\}_{i=1}^{N}\\
    \Mat{c}(r) = \mathcal{F}_{point}(\{\Mat{H}_{\Mat{q}}, \Mat{H}_{\Mat{k}}, \textit{DLM}(\Mat{H})_{\Mat{v}}\})\text{, where } \Mat{H} = \{\Mat{f}(t_{i})\}_{i=1}^{K}
\end{aligned}
\end{equation}
% \textcolor{blue}{In this section you have both DLM and DRM, aren't they both same or am I missing something? You don't seem to have defined DRM anywhere}
Since $\mathcal{F}_{\text{view}}$, $\mathcal{F}_{\text{point}}$ both represent transformer mechanisms, the operands within each explicitly indicate the query ($\Mat{q}$), key ($\Mat{k}$), value ($\Mat{v}$) features with their respective subscripts. 
We defer more implementation details to the supplementary. 

\subsection{Training and Inference}
\label{sec:trainandinf}

The entire pipeline can be end-to-end supervised across multiple scenes with multi-view images. 
At each training iteration, we sample from the $M$ degradation types, apply the same to the input images (with reasonable intra-corruption variability), and supervise the rendered view with its corresponding clean ground truth image. 
\begin{equation}
    \mathcal{L} = \mathcal{L}_{\text{MSE}} (\widehat{\Mat{I}}_{target}, \Mat{I}_{target}) + \mathcal{L}_{\text{LPIPS}} (\widehat{\Mat{I}}_{target}, \Mat{I}_{target})
\end{equation}
where $\mathcal{L}_{\text{MSE}}$, $\mathcal{L}_{\text{LPIPS}}$ denote the mean-squared error and LPIPS feature distance~\cite{zhang2018perceptual} criterion respectively, and $\Mat{I}_{target}$ represents the rendered target view and its corresponding ground truth view $\widehat{\Mat{I}}_{target}$. 
Given sufficient training data, the degradation-aware latent codes must encode specific information regarding each imperfection type and guide the recovery process.
We observe that by simply training on scenes containing synthetic realizations of several corruption types, \textit{GAURA} learns to generalize to scenes containing real-world imperfections.
Similar results have been seen in 2D image restoration~\cite{lv2021attention, lou2023simhaze, wang2022removing, zhang2020deblurring} and confirms the hypothesis in the 3D domain. 
More significantly, our DLM blocks are plug-and-play and can be easily expanded to accommodate new imperfection types. 
Being disentangled from other parts of the network, the new latent code can be efficiently fine-tuned with limited training (as little as 6-8 synthetic scenes) by re-purposing the priors learned during pretraining.

\vspace{-1em}
\section{Results}
\label{sec:results}
To demonstrate the effectiveness of \textit{GAURA}, we evaluate several image restoration tasks, including low-light enhancement, motion blur removal, image dehazing, and deraining. 
The useful priors learned by the trained model can help adapt \textit{GAURA} to any unseen degradation using an efficient finetuning strategy (discussed in Sec. \ref{sec:finetuning}), and we verify the same in two representative image restoration tasks: defocus blur removal and image desnowing. 
% \textcolor{blue}{Additionally, our method can also handle clean images and generate novel views without disturbing the appearance of the image. } 
In some of these restoration tasks, like image desnowing and deraining, we present the first 3D extension with no comparable previous 3D method. 
Our method is never fine-tuned on each scene for all the experiments listed below, and we only perform inferences using a trained model. 

\subsection{Implementation Details}
\label{sec:implementation}
We train our entire pipeline end-to-end on datasets of multi-view posed images. 
Specifically, we use the training data from IBRNet~\cite{wang2021ibrnet} and LLFF\cite{mildenhall2019llff} and synthetically alter the original clean scene to contain imperfections. 
At every training iteration, we sample from one of 4 different corruption types (i.e., low-light, motion blur, haze, or rain) and apply it to the input images. 
To simulate real-world data, we incorporate variations within each corruption type by randomly sampling the strength of the corruption. This helps the network generalize better (more implementation details in the supplementary materials).
We follow the same input view sampling strategy from IBRNet~\cite{wang2021ibrnet}, and select between 8-12 source views during training and a fixed number of 10 views during inference on any new scene. 
Rather than training from scratch, we reuse the pretrained checkpoint from GNT~\cite{gnt} and add our proposed degradation-aware modules to the necessary parts of the network. 
Our method is optimized to render clean images using the Adam optimizer with an initial learning rate of $5\times10^{-4}$, decayed over the course of 400,000 training steps.

\subsection{Comparisons}
\label{sec:comparisons}
Since no previous works deal with generalizable restoration and view synthesis, we devise several reasonable baselines for comparison.

\vspace{-0.7em}
\paragraph{\textbf{NeRF--Restore.}} refers to a Vanilla NeRF~\cite{mildenhall2020nerf} trained on images restored using a single-image restoration technique, specialized for the given degradation type. 
The restoration technique in the study is different for different input imperfections.
This baseline does not generalize to scenes or corruption types. 

\vspace{-0.7em}
\paragraph{\textbf{GNT--Restore.}} uses a similar strategy as Restore-NeRF but replaces the Vanilla NeRF with a generalizable novel view synthesis technique GNT~\cite{gnt}. 
Such a baseline can generalize across scenes but is specific to a given restoration type. 

\vspace{-0.7em}
\paragraph{\textbf{GNT-(All-in-One) Restore.}} Recent single-image restoration methods are All-in-One i.e., can handle multiple degradation types.
We leverage state-of-the-art all-in-one methods~\cite{daclip, potlapalli2023promptir, jiang2023autodir, airnet} followed by GNT for novel view synthesis. 
Such a baseline is neither scene nor corruption-specific and is, therefore, a more suitable comparison to our method. 

\vspace{-0.7em}
\paragraph{\textbf{3D Restore.}} We also compare against specialized NeRF techniques for the given degradation type, wherever available. 
% Specific details on the method are deferred to the respective subsections. 
Similar to NeRF-Restore, this baseline is both scene and degradation-specific. 

\noindent The best-performing method choices for each baseline across individual degradation types are discussed in the supplementary materials. 

% \textcolor{blue}{Details regarding the best performing method in each corruption chosen for evaluation is provided in the supplementary. }

\begin{table}[t]
  \centering
  \caption{
  Quantitative results on scenes containing \textit{real-world} degradations - specifically low-light enhancement, motion blur removal, and dehazing. The \sethlcolor{black!30}\hl{best} scores and \sethlcolor{black!15}\hl{second best} scores are highlighted with their respective colors only amongst the generalizable methods. 
  }
  \resizebox{1.0\columnwidth}{!}{
  \begin{tabular}{lccccc|ccc|ccc}
  
    \toprule
    \multirow{2}{*}{Models} & \multicolumn{2}{c}{Generalize to} & \multicolumn{3}{c|}{Low-Light} & \multicolumn{3}{c|}{Motion Blur} & \multicolumn{3}{c}{Haze}\\
    \cmidrule(r){2-12}
    & Scene & Corr. & \hspace{0.2em}PSNR$\uparrow$\hspace{0.2em} & SSIM$\uparrow$\hspace{0.2em} & LPIPS$\downarrow$\hspace{0.2em} & \hspace{0.2em}PSNR$\uparrow$\hspace{0.2em} & SSIM$\uparrow$\hspace{0.2em} & LPIPS$\downarrow$\hspace{0.2em} & \hspace{0.2em}PSNR$\uparrow$\hspace{0.2em} & SSIM$\uparrow$\hspace{0.2em} & LPIPS$\downarrow$\\
    \midrule
    NeRF--Restore & \xmark & \xmark & 15.42 & 0.702 & 0.393 & 23.27 & 0.717 & 0.331 & 13.87 & 0.690 & 0.333\\ 
    3D Restore & \xmark & \xmark & 17.64 & 0.736 & 0.415 & 25.65 & 0.758 & 0.181 & - & - & -\\ 
    \midrule
    GNT--Restore & \cmark & \xmark & 16.36 & \cellcolor{black!15}0.730 & \cellcolor{black!30}0.344 & \cellcolor{black!15}21.97 & \cellcolor{black!15}0.705 & \cellcolor{black!30}0.392 & 14.16 & 0.678 & \cellcolor{black!30}0.279\\ 
    GNT--(All-in-one) Restore & \cmark & \cmark & \cellcolor{black!15}17.90 & 0.573 & 0.354 & 20.88 & 0.632 & 0.41 & \cellcolor{black!15}16.68 & \cellcolor{black!15}0.729 & 0.300\\ 
    \midrule
    Ours & \cmark & \cmark & \cellcolor{black!30}19.91 & \cellcolor{black!30}0.738 & \cellcolor{black!15}0.352 & \cellcolor{black!30}22.12 & \cellcolor{black!30}0.712 & \cellcolor{black!15}0.346 & \cellcolor{black!30}16.82 & \cellcolor{black!30}0.759 & \cellcolor{black!15}0.288\\ 
    \bottomrule
  \end{tabular}
  }
  
  \label{tab:realscenes}
  \vspace{-1em}
\end{table}

\begin{table}[t]
  \centering
  \caption{
  Quantitative results on the LLFF-Corrupted benchmark, containing synthetically altered scenes of several imperfection types.}

  \resizebox{1.0\columnwidth}{!}{
  \begin{tabular}{lccc|ccc|ccc|ccc}
    \toprule
    \multirow{2}{*}{Models} & \multicolumn{3}{c|}{Low-Light} & \multicolumn{3}{c|}{Motion Blur} & \multicolumn{3}{c|}{Haze} & \multicolumn{3}{c}{Rain}\\
    \cmidrule(r){2-13}
     & \hspace{0.2em}PSNR$\uparrow$\hspace{0.2em} & SSIM$\uparrow$\hspace{0.2em} & LPIPS$\downarrow$\hspace{0.2em} & \hspace{0.2em}PSNR$\uparrow$\hspace{0.2em} & SSIM$\uparrow$\hspace{0.2em} & LPIPS$\downarrow$\hspace{0.2em} & \hspace{0.2em}PSNR$\uparrow$\hspace{0.2em} & SSIM$\uparrow$\hspace{0.2em} & LPIPS$\downarrow$ & \hspace{0.2em}PSNR$\uparrow$\hspace{0.2em} & SSIM$\uparrow$\hspace{0.2em} & LPIPS$\downarrow$\\
    \midrule
    GNT--Airnet\cite{airnet} & \cellcolor{black!15}18.2 & \cellcolor{black!15}0.590 & \cellcolor{black!15}0.405 & 21.08 & 0.604 & 0.410 & 14.55 & 0.501 & 0.400 & 20.71 & \cellcolor{black!15}0.612 & \cellcolor{black!30}0.319\\ 
    GNT--PromptIR\cite{potlapalli2023promptir} & 17.67 & 0.571 & \cellcolor{black!30}0.389 & 21.01 & 0.604 & \cellcolor{black!15}0.378 & \cellcolor{black!15}15.81 & \cellcolor{black!15}0.514 & \cellcolor{black!30}0.361 & \cellcolor{black!15}20.73 & \cellcolor{black!15}0.612 & \cellcolor{black!15 }0.276 \\ 
    GNT--DA-CLIP\cite{daclip} & 12.458 & 0.527 & 0.473 & \cellcolor{black!15}21.96 & \cellcolor{black!15}0.634 & 0.412 & 8.36 & 0.443 & 0.563 & 20.24 & 0.608 & 0.434 \\ 
    GNT--AutoDIR\cite{jiang2023autodir} & 10.48 & 0.453 & 0.552 & 21.85 & 0.628 & 0.427 & 12.59 & 0.511 & 0.538 & 19.25 & 0.558 & 0.501 \\ 
    \midrule
    Ours & \cellcolor{black!30}21.98 & \cellcolor{black!30}0.598 & 0.464 & \cellcolor{black!30}22.61 & \cellcolor{black!30}0.680 & \cellcolor{black!30}0.349 & \cellcolor{black!30}18.95 & \cellcolor{black!30}0.642 & \cellcolor{black!15}0.383 & \cellcolor{black!30}22.61 & \cellcolor{black!30}0.710 & 0.342 \\ 
    \bottomrule
  \end{tabular}
  }
  
  \label{tab:syntheticscenes}
  % \vspace{-2em}
\end{table}

\begin{figure*}[t]
\centering
\captionsetup[subfigure]{labelformat=empty}
% \rotatebox[origin=lc]{90}{\scriptsize{Low-Light}}
\rotatebox{90}{\quad \tiny{Low-Light}}
\begin{subfigure}[t]{0.19\textwidth}
  \centering
  \includegraphics[width=1.0\linewidth]{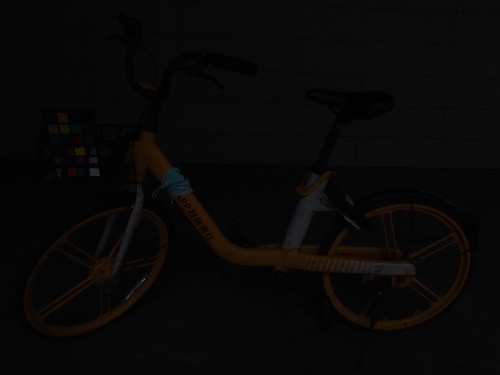}
\end{subfigure}%
\begin{subfigure}[t]{0.19\textwidth}
  \centering
  \includegraphics[width=1.0\linewidth]{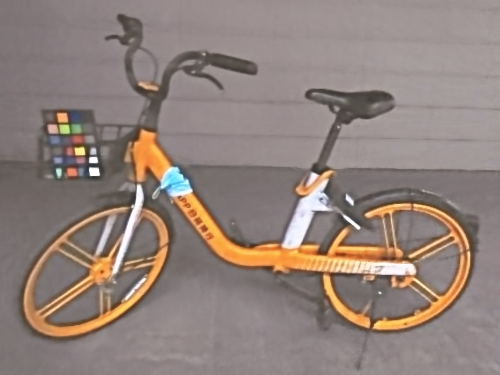}
\end{subfigure}%
\begin{subfigure}[t]{0.19\textwidth}
  \centering
  \includegraphics[width=1.0\linewidth]{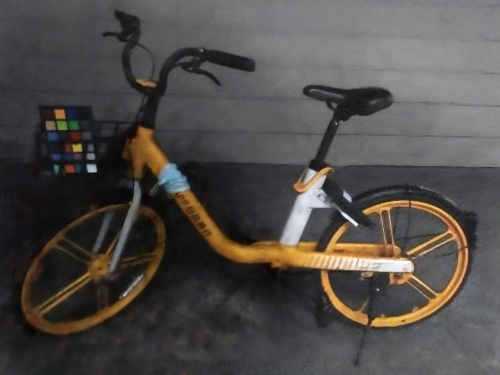}
\end{subfigure}%
\begin{subfigure}[t]{0.19\textwidth}
  \centering
  \includegraphics[width=1.0\linewidth]{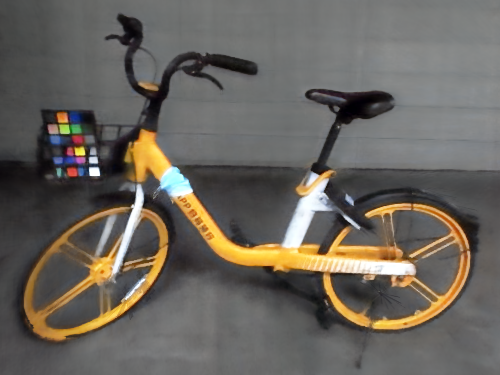}
\end{subfigure}%
\begin{subfigure}[t]{0.19\textwidth}
  \centering
  \includegraphics[width=1.0\linewidth]{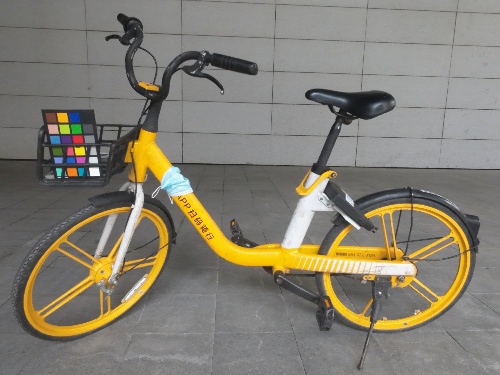}
\end{subfigure}%
\setcounter{subfigure}{0}

\rotatebox{90}{\, \tiny{Motion Blur}}
\begin{subfigure}[t]{0.19\textwidth}
  \centering
  \includegraphics[width=1.0\linewidth]{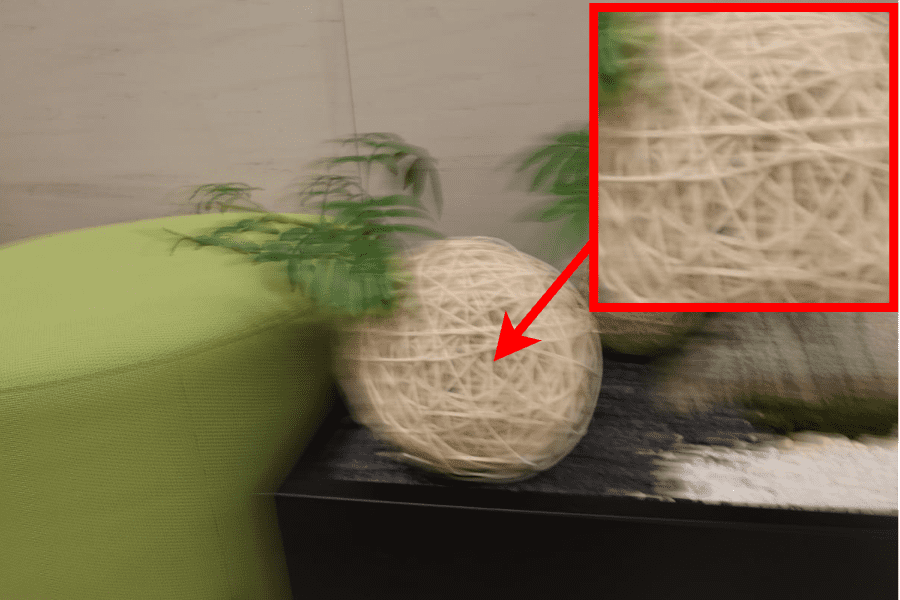}
\end{subfigure}%
\begin{subfigure}[t]{0.19\textwidth}
  \centering
  \includegraphics[width=1.0\linewidth]{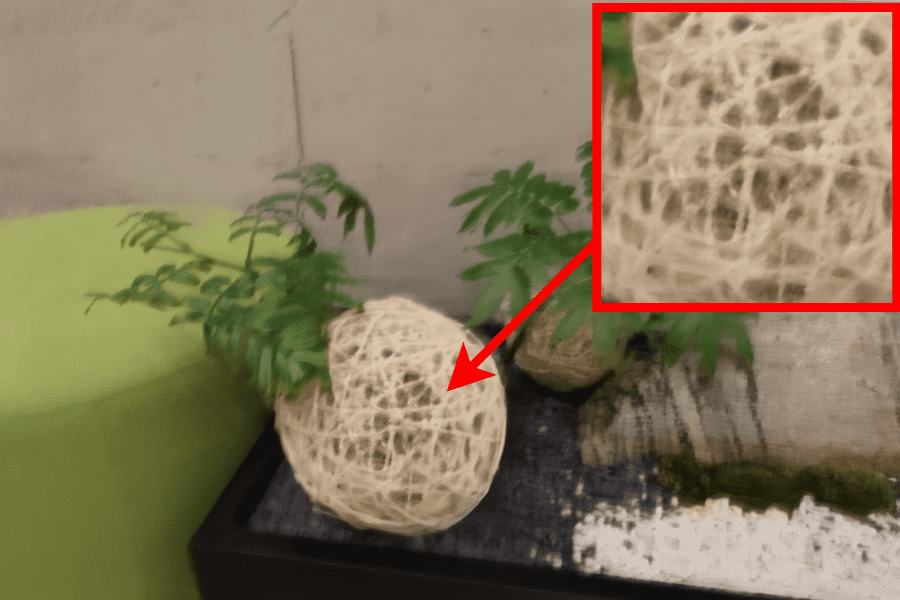}
\end{subfigure}%
\begin{subfigure}[t]{0.19\textwidth}
  \centering
  \includegraphics[width=1.0\linewidth]{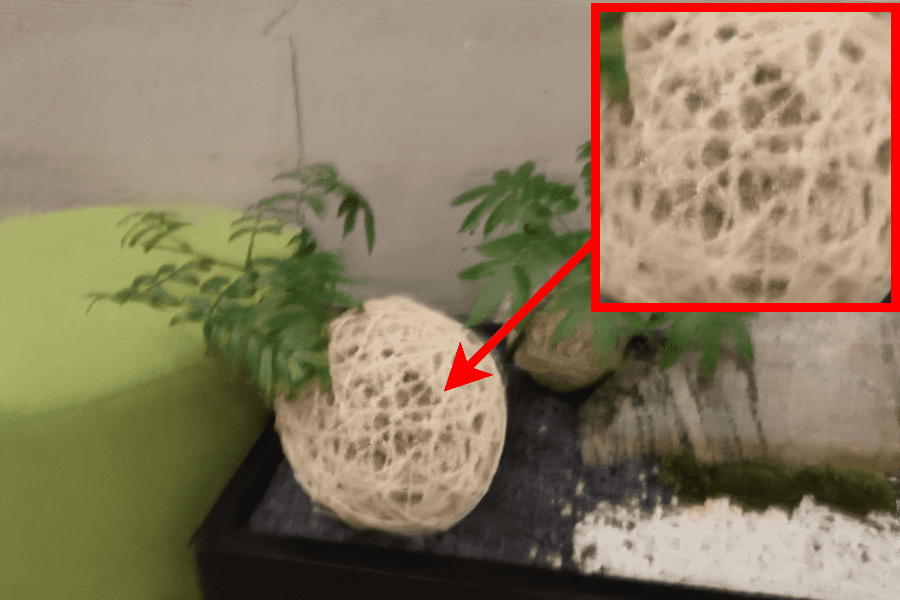}
\end{subfigure}%
\begin{subfigure}[t]{0.19\textwidth}
  \centering
  \includegraphics[width=1.0\linewidth]{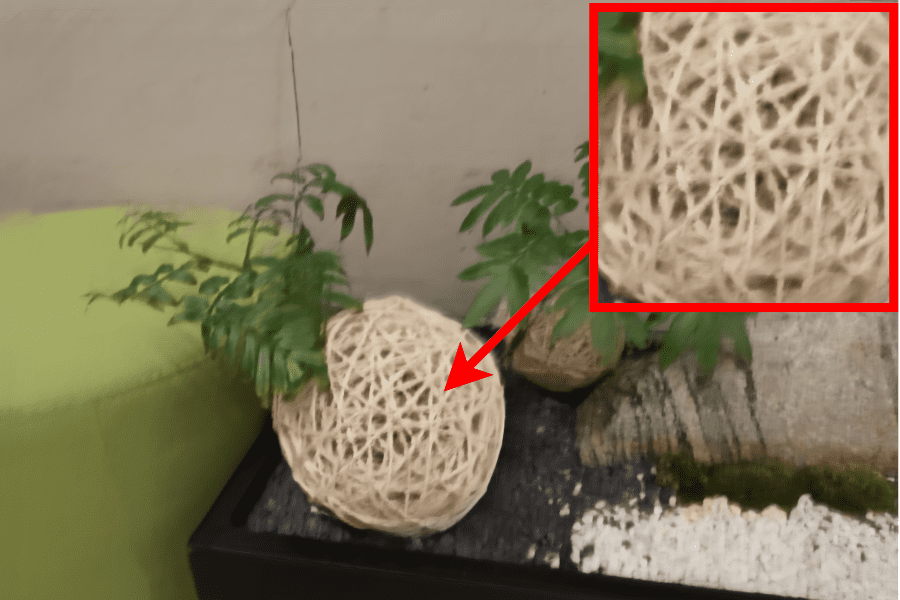}
\end{subfigure}%
\begin{subfigure}[t]{0.19\textwidth}
  \centering
  \includegraphics[width=1.0\linewidth]{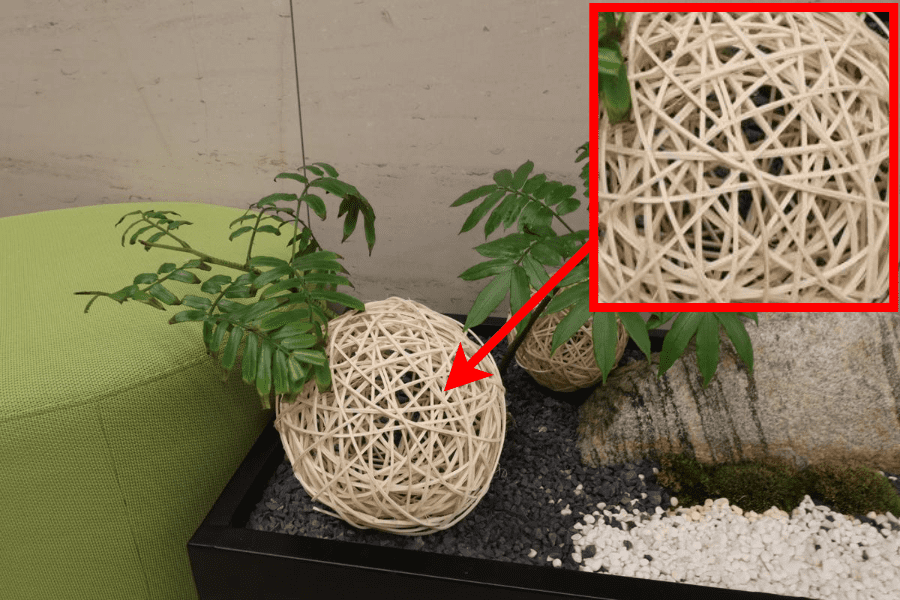}
\end{subfigure}%
\setcounter{subfigure}{0}

\rotatebox{90}{\quad\, \tiny{Haze}}
\begin{subfigure}[t]{0.19\textwidth}
  \centering
  \includegraphics[width=1.0\linewidth]{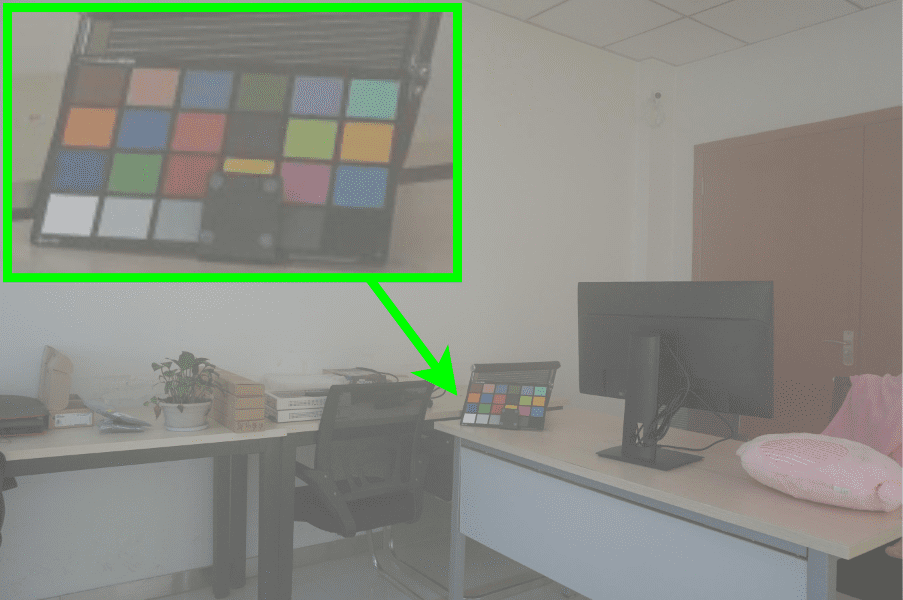}
\caption{\tiny Corrupted View}
\end{subfigure}%
\begin{subfigure}[t]{0.19\textwidth}
  \centering
  \includegraphics[width=1.0\linewidth]{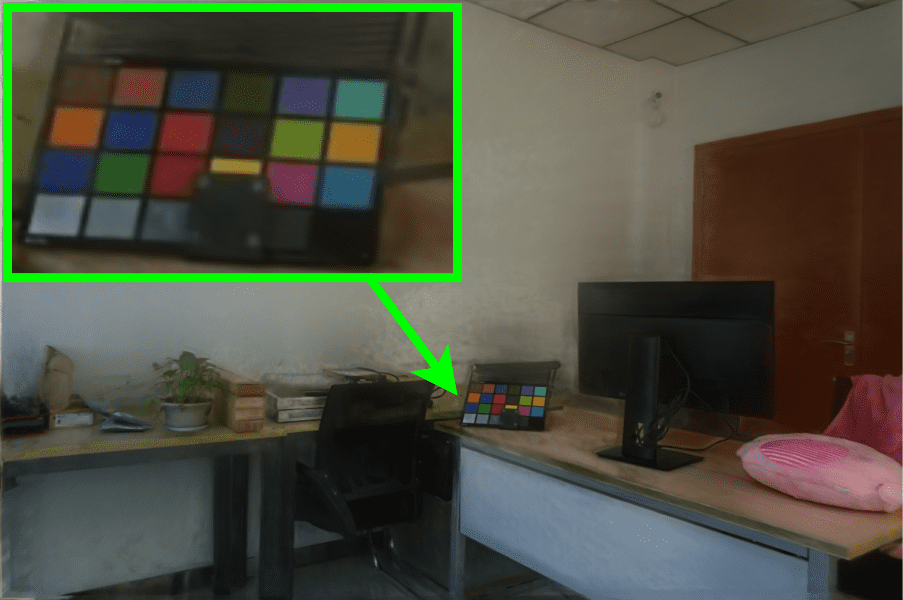}
\caption{\tiny GNT--Restore}
\end{subfigure}%
\begin{subfigure}[t]{0.19\textwidth}
  \centering
  \includegraphics[width=1.0\linewidth]{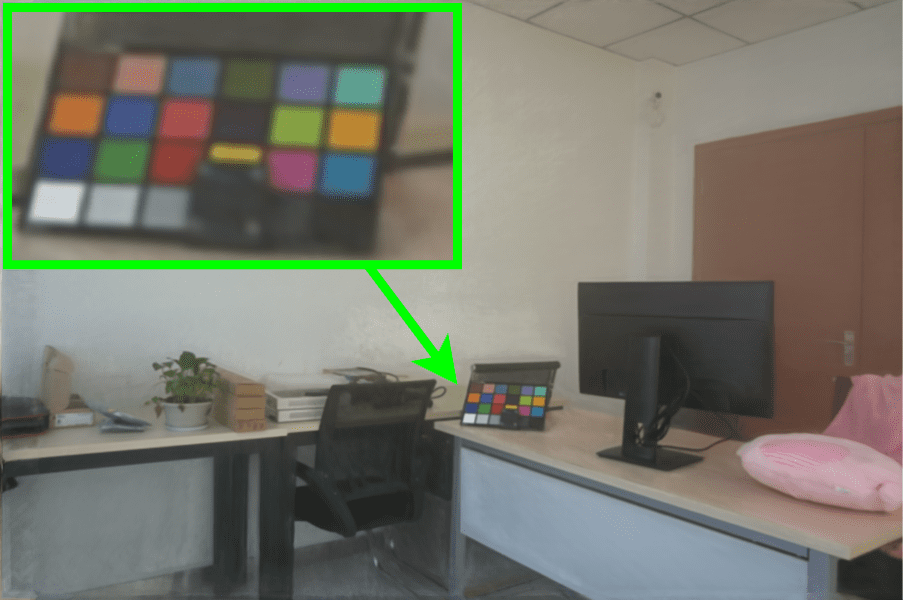}
% \captionsetup{font={tiny}}

% \caption{\scalebox{0.6}{(All-in-One) Restore--GNT}}
\caption{\scalebox{0.6}{GNT--(All-in-One) Restore}}
\end{subfigure}%
\begin{subfigure}[t]{0.19\textwidth}
  \centering
  \includegraphics[width=1.0\linewidth]{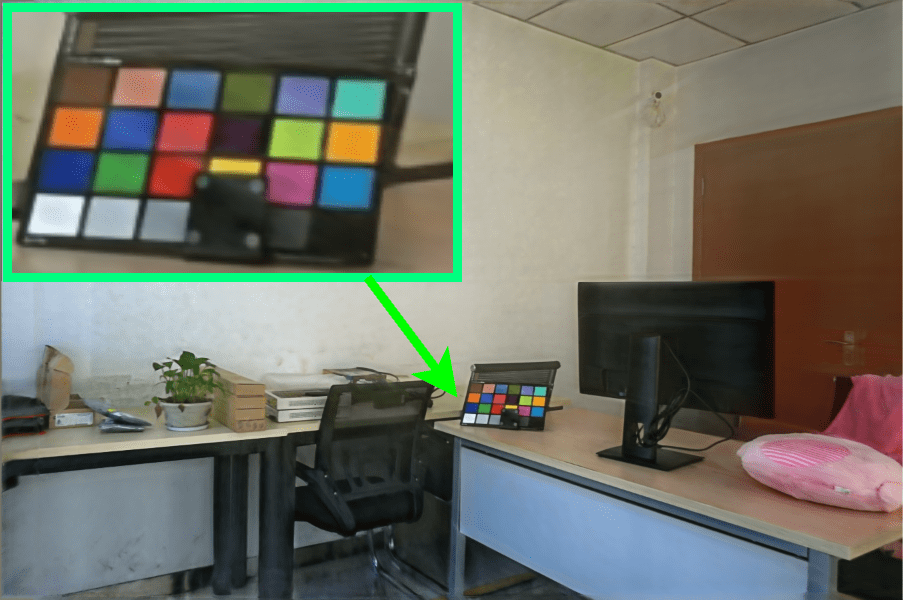}
\caption{\tiny Ours}
\end{subfigure}%
\begin{subfigure}[t]{0.19\textwidth}
  \centering
  \includegraphics[width=1.0\linewidth]{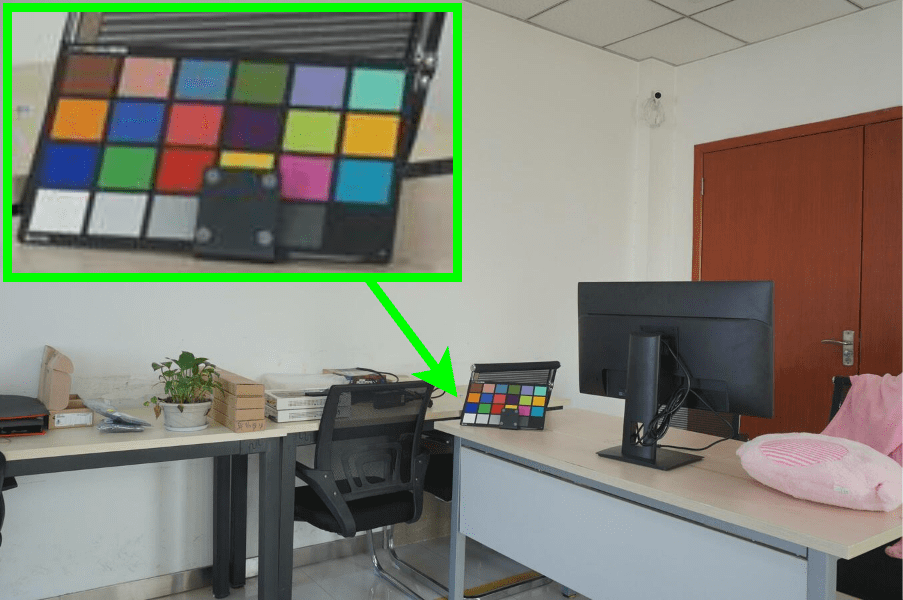}
\caption{\tiny Ground Truth}
\end{subfigure}%

\caption{Qualitative results on three restoration tasks: low-light enhancement, motion blur removal, and dehazing on \textit{real-world} datasets. In the first row, our method successfully recovers the underlying 3D scene from poorly lit images much closer to the ground truth than the baselines. In the second row, our method can reconstruct fine details (woven patterns on the ball) with higher fidelity while in the third row, GAURA successfully removes haze from the synthesized views and can accurately match the colors on the palette more closely to the ground truth color. 
% We present results from our model and the baselines on 3 restoration tasks: low-light enhancement, motion deblurring and dehazing. In the first row, the baselines struggle to get the appearance correct and produces artifacts. In the second row, we observe that our method recovers the holes in the ball better compared to the baselines. In the third row, the haze from the colour pallette is removed successfully by our model. 
}
\label{fig:dark_motion_haze}
\vspace{-1em}
\end{figure*}

%%%%% 

\begin{figure*}[t]
\centering
\captionsetup[subfigure]{labelformat=empty}
\rotatebox{90}{\quad\;\;\; \scriptsize{Rain}}
\begin{subfigure}[t]{0.32\textwidth}
  \centering
  \includegraphics[width=1.0\linewidth]{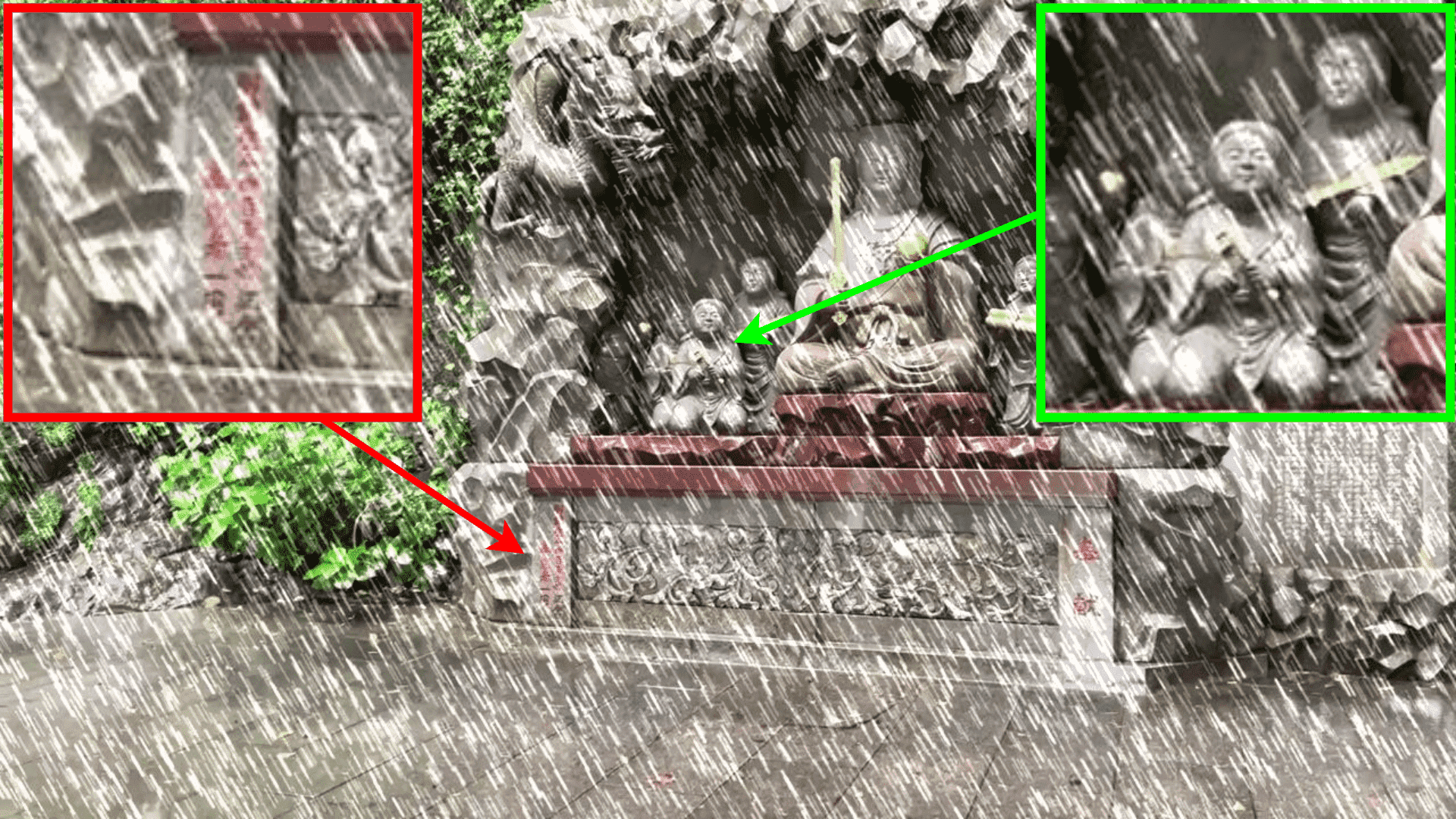}
\end{subfigure}%
\begin{subfigure}[t]{0.32\textwidth}
  \centering
  \includegraphics[width=1.0\linewidth]{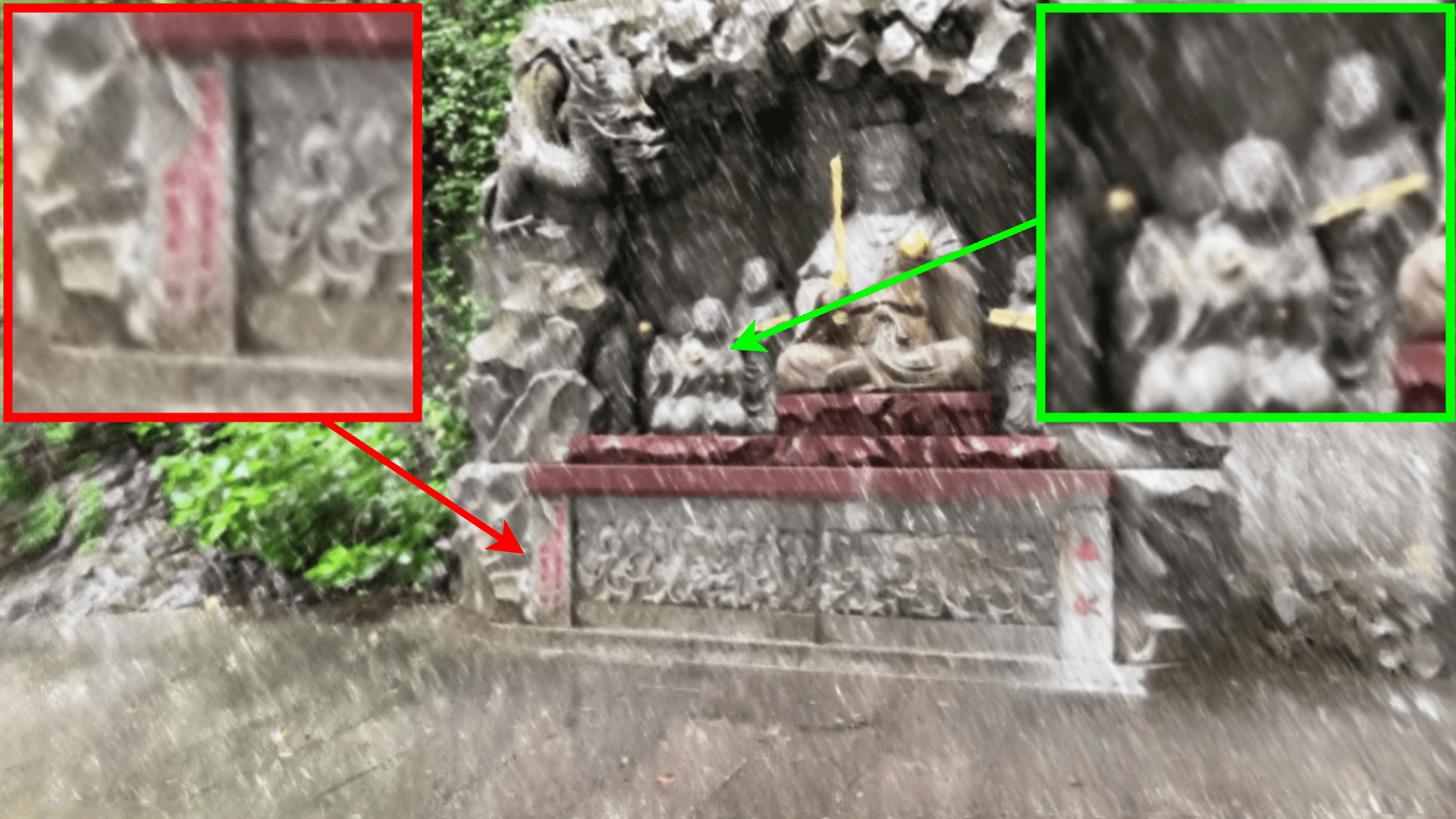}
\end{subfigure}%
\begin{subfigure}[t]{0.32\textwidth}
  \centering
  \includegraphics[width=1.0\linewidth]{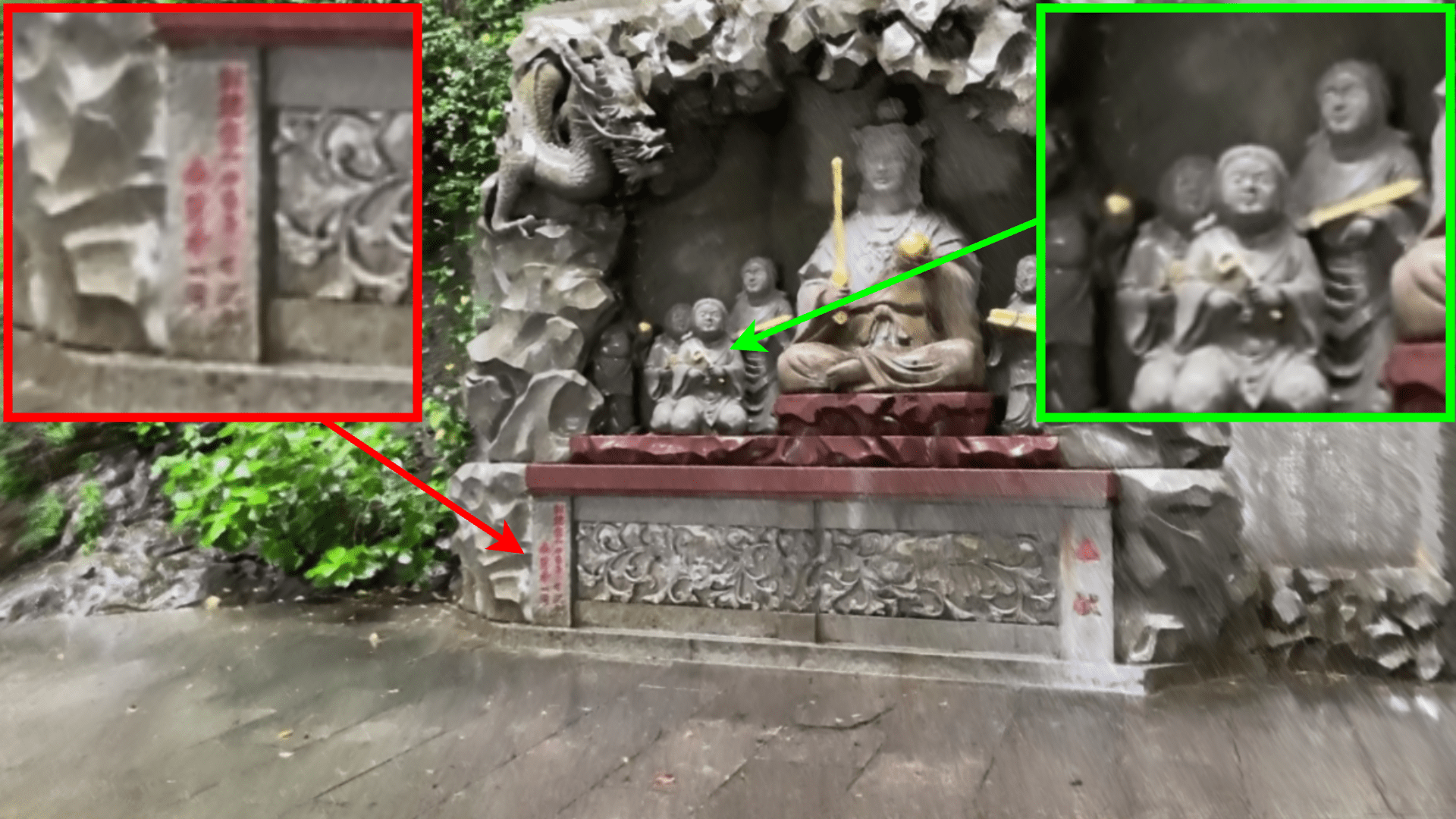}
\end{subfigure}%
\setcounter{subfigure}{0}

\rotatebox{90}{\quad\;\;\; \scriptsize{Snow}}
\begin{subfigure}[t]{0.32\textwidth}
  \centering
  \includegraphics[width=1.0\linewidth]{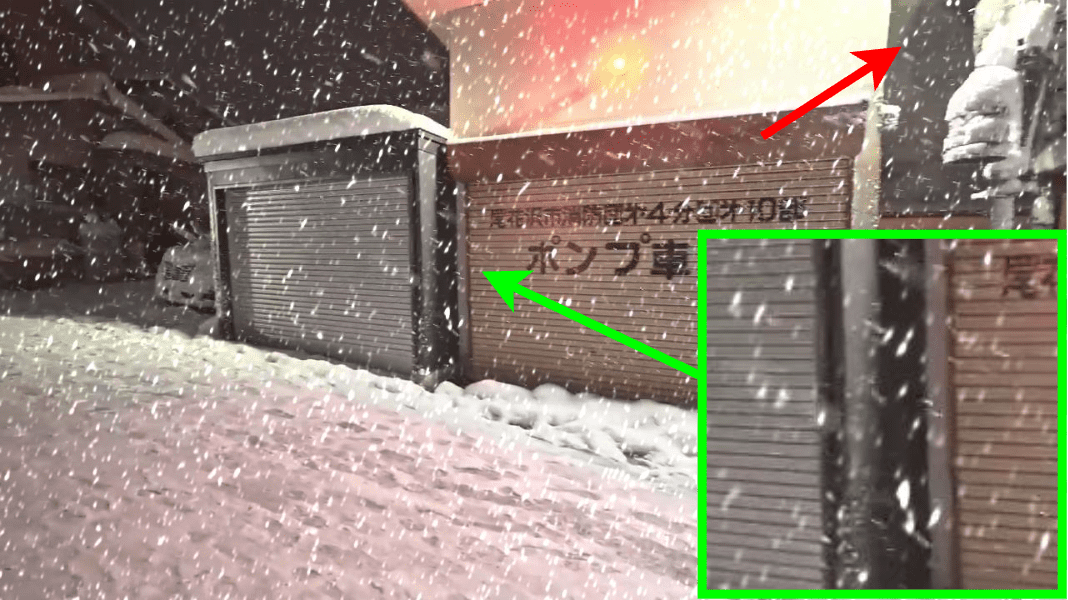}
\end{subfigure}%
\begin{subfigure}[t]{0.32\textwidth}
  \centering
  \includegraphics[width=1.0\linewidth]{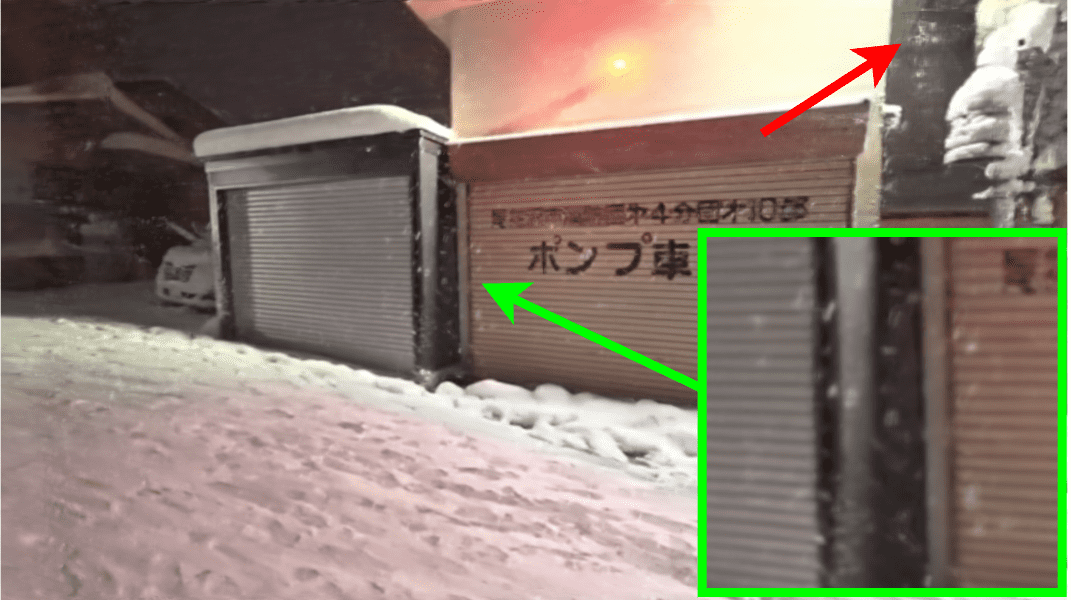}
\end{subfigure}%
\begin{subfigure}[t]{0.32\textwidth}
  \centering
  \includegraphics[width=1.0\linewidth]{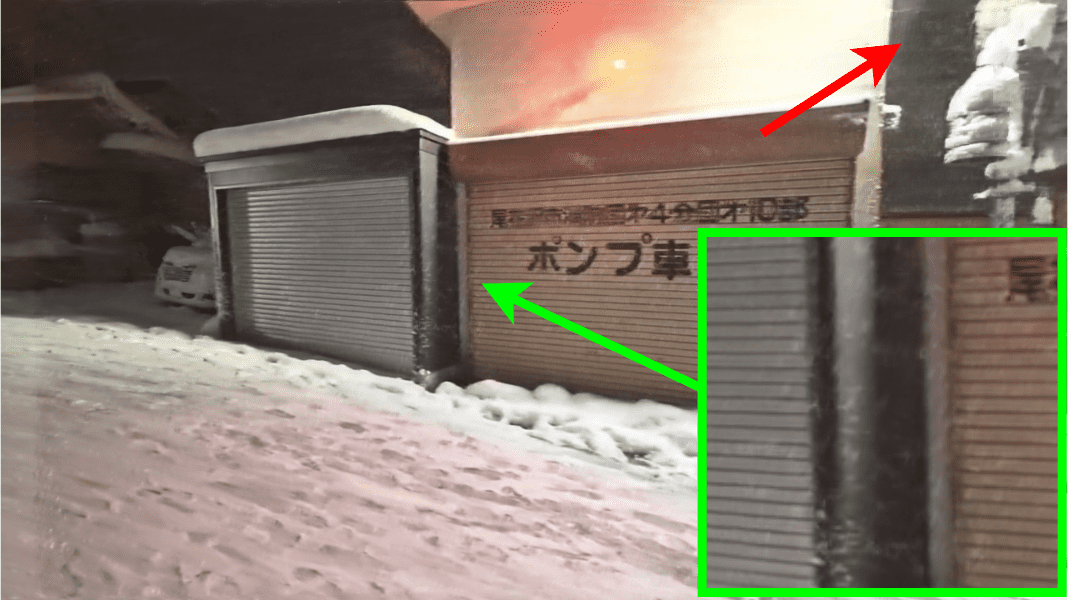}
\end{subfigure}%

\setcounter{subfigure}{0}

\rotatebox{90}{\;\;\; \scriptsize{Defocus Blur}}
\begin{subfigure}[t]{0.32\textwidth}
  \centering
  \includegraphics[width=1.0\linewidth]{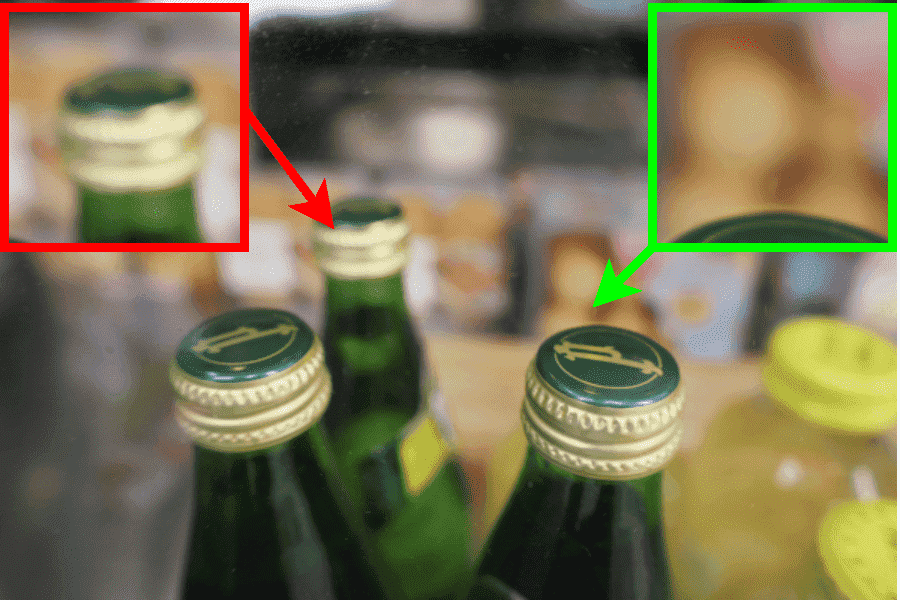}
\caption{\scriptsize Corrupted View}
\end{subfigure}%
\begin{subfigure}[t]{0.32\textwidth}
  \centering
  \includegraphics[width=1.0\linewidth]{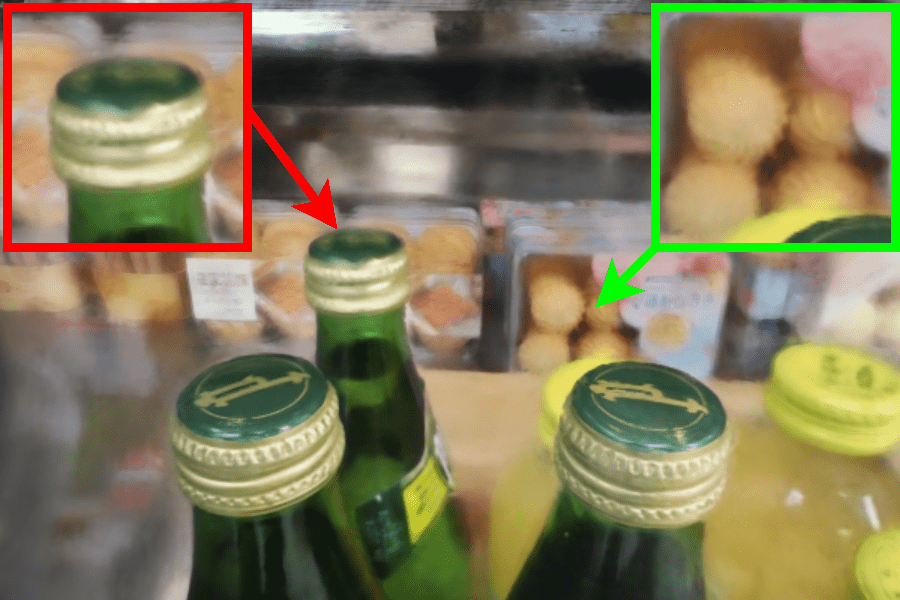}
\caption{\scriptsize GNT--Restore}
\end{subfigure}%
\begin{subfigure}[t]{0.32\textwidth}
  \centering
  \includegraphics[width=1.0\linewidth]{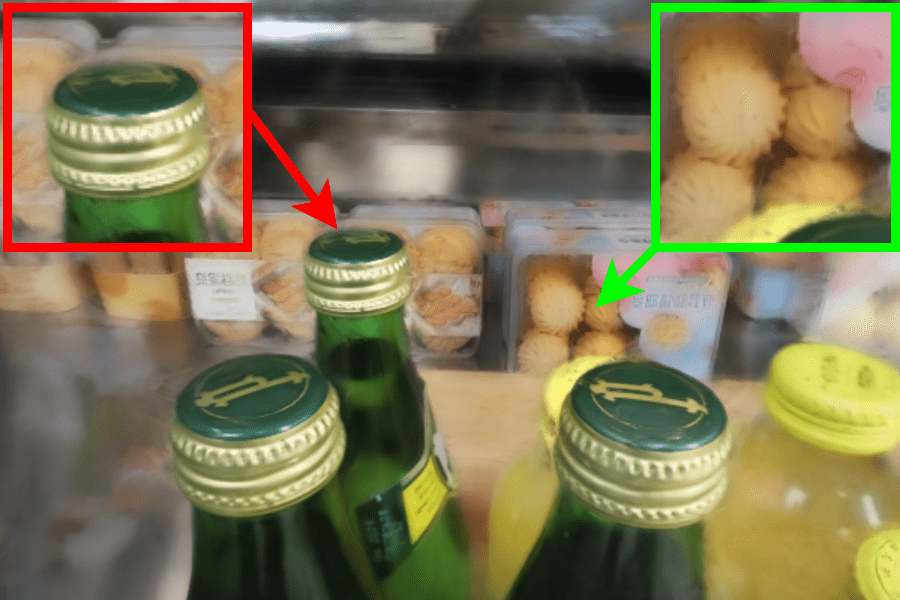}
\caption{\scriptsize Ours}
\end{subfigure}%

\caption{
Qualitative results of GAURA on additional restoration tasks. Our pretrained model can remove imperfections caused due to rain particles (row 1). With limited finetuning, GAURA can learn to successfully restore any new degradation type (unseen during training). For representation purposes, we visualize results on snow (row 2) and defocus blur (row 3), and our method outperforms degradation-specific restoration baselines. 
% Qualitative results on deraining, desnowing and defocus blurring tasks. In the first 2 rows, our method consistently eliminates rain and snow from the image while preserving the original scene geometry compared to baseline methods. In the last row, our method notably excels in recovering fine details of the scene, such as the bottle cap and cookies
}

\label{fig:rain_snow_defocus}
\vspace{-1em}
\end{figure*}

\subsection{Generalization to Multiple Degradations}
\label{sec:generalizedeg}
% \vspace{-0.7em}
\paragraph{\textbf{Low-light Enhancement. }}
Given multi-view images of a scene captured in low light, we aim to render images from arbitrary viewpoints as if they were viewed with sufficient illumination. 
We evaluate our method and the baselines discussed in Sec. \ref{sec:comparisons} on the dataset from AlethNeRF~\cite{cui_aleth_nerf}, containing in total 5 scenes and discuss results in Table~\ref{tab:realscenes}.
In the case of scene-specific techniques, we optimize a radiance field for each scene (i.e., 2D Restore-NeRF and 3D-Restore), while the generalizable methods, including ours, are directly evaluated on the benchmark dataset, not observed during training. 
GAURA, although neither scene nor corruption-specific, manages to outperform other comparison methods significantly, even 3D restoration methods tuned specifically for low-light enhancement~\cite{cui_aleth_nerf} (3D Restore in Table~\ref{tab:realscenes})
This indicates the effectiveness of the learned rendering process compared to explicitly modeling the degradation in the image formation process. 
In Fig. \ref{fig:dark_motion_haze} (row 1), we observe that GAURA can faithfully recover the scene much closer to the ground truth from poorly lit scenes. 

% \vspace{-0.7em}
\paragraph{\textbf{Deblurring. }} 
We evaluate our method on the task of blur-removal, particularly motion blur on the scenes from the DeblurNeRF dataset~\cite{ma2022deblur}, and average results across 10 different scenes in Table~\ref{tab:realscenes}. 
We observe that our method performs on par with methods that require any optimization and significantly better than a fully generalizable baseline~\cite{daclip} indicated as All-in-one-GNT in Table~\ref{tab:realscenes} ($\sim$10\% $\uparrow$ PSNR).
% Naively combining novel view synthesis~\cite{} with pretrained All-in-One single image restoration techniques~\cite{}, though generalizable like ours, often leads to multi-view inconsistency, while GAURA ensures view-consistent and high-quality restoration.
We wish to reiterate that although the 3D Restore technique~\cite{ma2022deblur} performs better than GAURA, it requires per-scene optimization and is specific to the task of deblurring, while our method is generalizable. 

Fig. \ref{fig:dark_motion_haze} (row 2) visualizes the rendered novel views, and our method can recover fine details in the scene (e.g., woven patterns in the ball) with superior quality compared to other baselines.

\vspace{-0.7em}
\paragraph{\textbf{Haze Removal. }} In this task, we look at novel view synthesis from input images captured in a hazy environment. 
We compare GAURA against different baselines on the dataset from~\cite{zhang2021learning} containing 5 scenes, initially not intended for novel view synthesis. 
We run COLMAP on the video frames given in the dataset to extract camera poses and test on every 8th image in the dataset. 
% We sample 30-40 frames from the continuously captured hazy videos, run COLMAP to extract camera poses, and test on five held-out views.
Unlike the previous two degradations, there are no existing 3D restoration techniques, perhaps indicating the complexity of explicitly modeling the degradation process and further supporting the practical applicability of our method.
% \textcolor{blue}{(Actually there are 3 Haze+NeRF methods, but none of them have released code and hence we cannot compare. So I don't know whether it is appropriate to call it as no existing 3D restoration method. Also we have cited on of the Dehaze-NeRF papers in the introduction and explained what it does in brief. So do you think we should call here as no existing 3d methods?}  
From Table~\ref{tab:realscenes}, we see that our method showcases better quantitative results when compared to other baselines. 
From Fig.~\ref{fig:dark_motion_haze} (row 3), we observe that \textit{GAURA} successfully removes haze from the synthesized views and accurately matches the color palette more closely to the ground truth color. 

\vspace{-0.1em}
\paragraph{\textbf{Results on LLFF-Corrupted. }} In addition to the results discussed above on existing benchmarks, we manually alter all eight scenes from  LLFF~\cite{mildenhall2019llff} into several corrupted versions, containing one low-light, motion blur, haze, or rain imperfections.
We follow the same train-test split from the original LLFF dataset and evaluate GAURA's generalization capabilities on this benchmark (dubbed LLFF-Corrupted).  
We compare against a similar generalizable baseline as ours, an All-in-One restoration technique~\cite{potlapalli2023promptir,jiang2023autodir,airnet,daclip} followed by generalizable rendering~\cite{gnt}, and retrain both restoration and rendering components on the same synthetic training dataset as GAURA to ensure a fair comparison.
Table~\ref{tab:syntheticscenes} shows that GAURA significantly outperforms all baselines across all imperfection types. 
This indicates the capacity of our model to implicitly encode different degradation processes into neural networks, in this case, transformers, and that such a learned renderer is performing well. 
We visualize qualitative results in the supplementary materials. 

% \begin{figure*}[t]
%   \includegraphics[width=\textwidth]{figures/multi-corr.png}
%     \caption{\textcolor{red}{Comaprison with all-in-models which have GT}}
%     \label{fig:visual_comparison}
% \end{figure*}

\begin{figure*}[t]
\centering
\captionsetup[subfigure]{labelformat=empty}
% \rotatebox[origin=c]{90}{Scale=6}
\rotatebox{90}{\quad Rain+Dark}
\begin{subfigure}[t]{0.31\textwidth}
  \centering
  \includegraphics[width=1.0\linewidth]{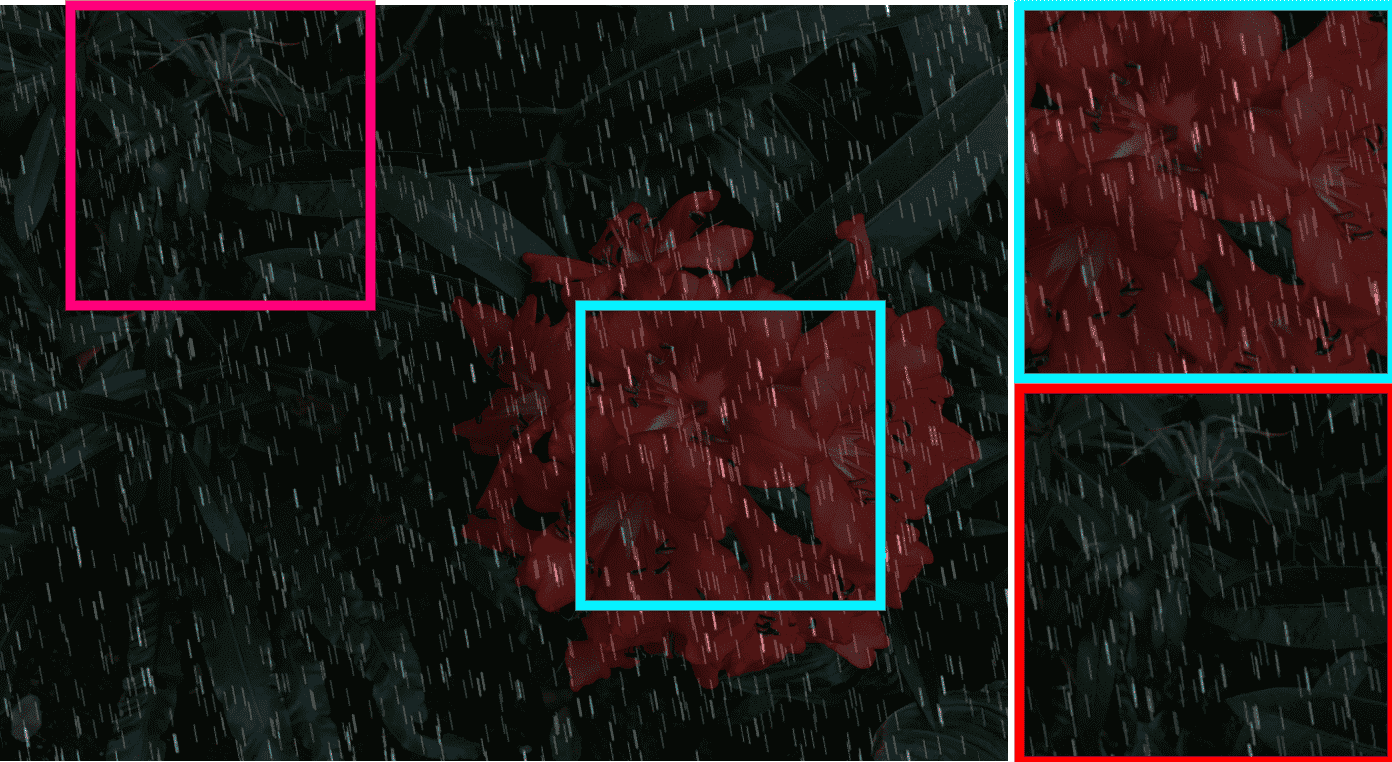}
\end{subfigure}%
\begin{subfigure}[t]{0.31\textwidth}
  \centering
  \includegraphics[width=1.0\linewidth]{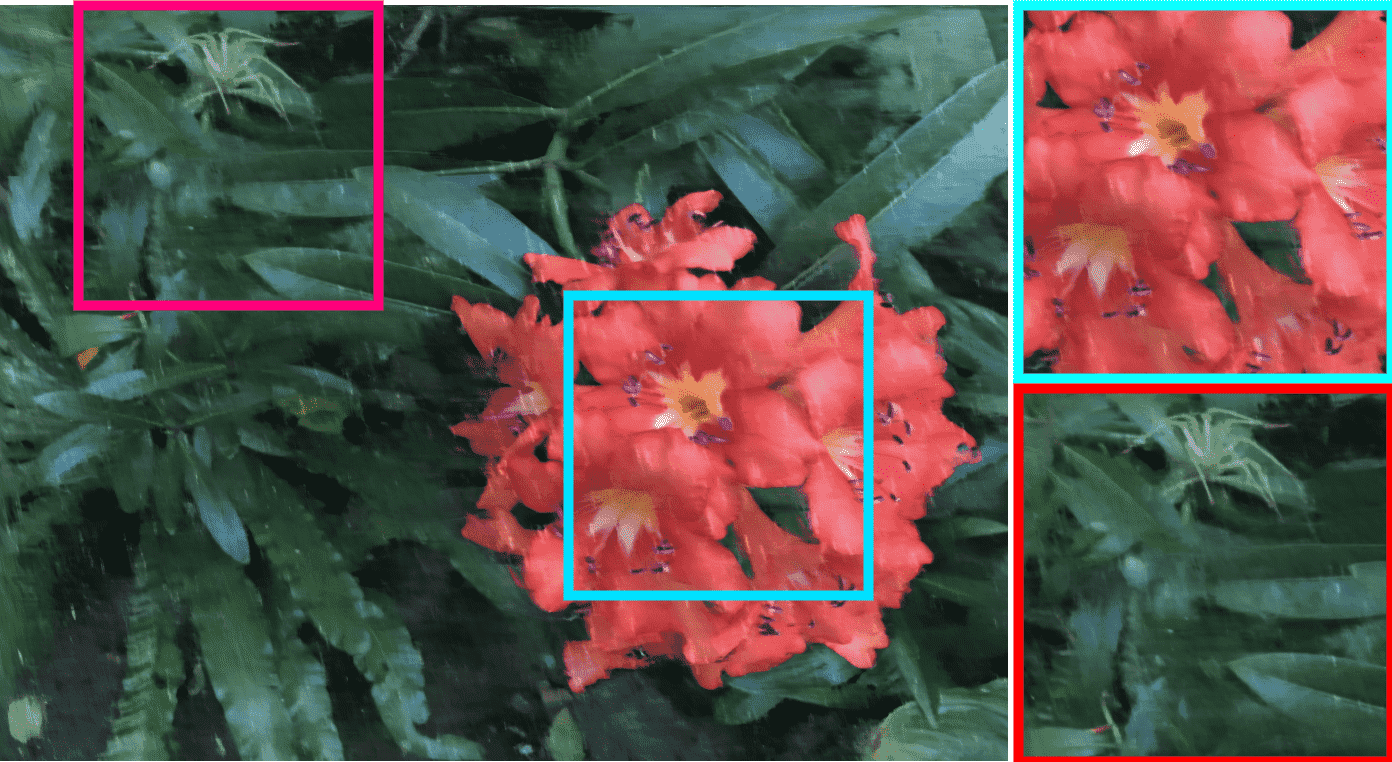}
\end{subfigure}%
\begin{subfigure}[t]{0.31\textwidth}
  \centering
  \includegraphics[width=1.0\linewidth]{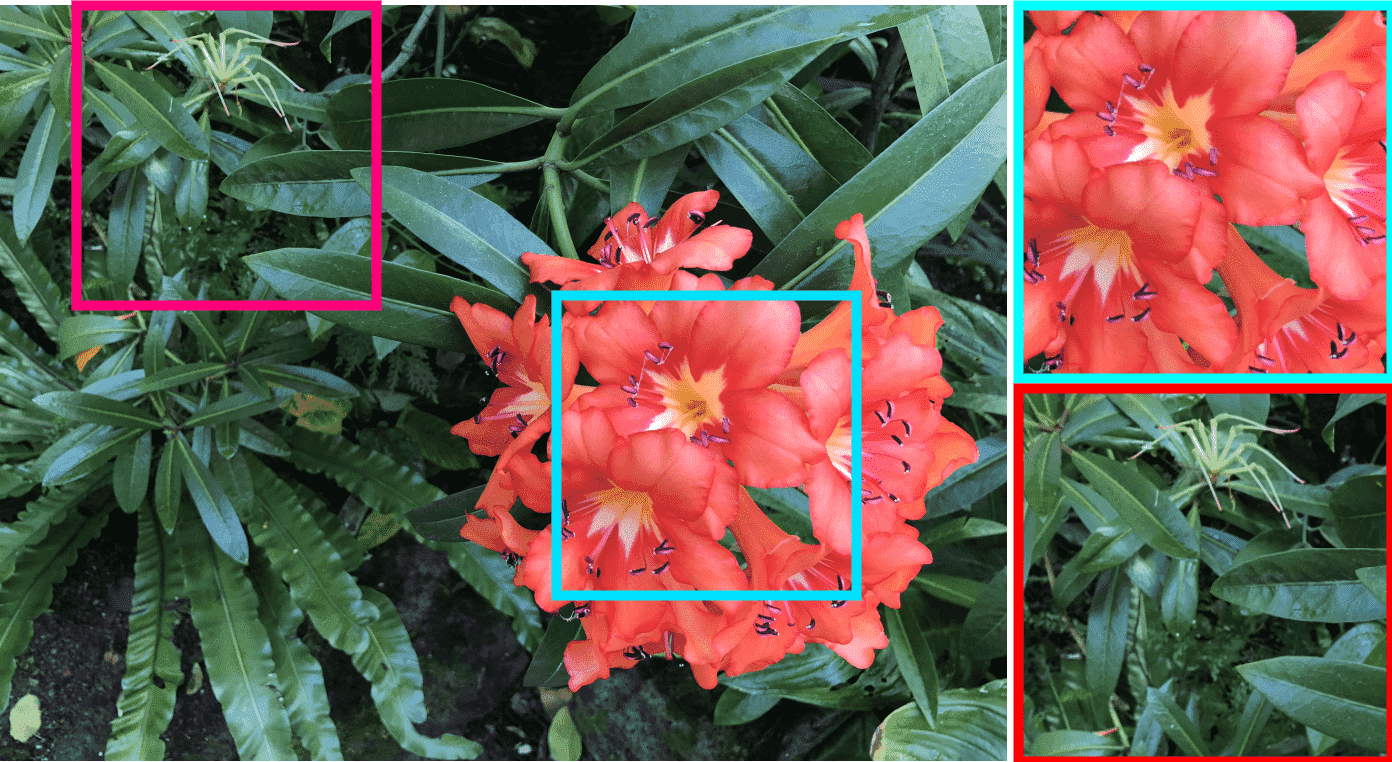}
\end{subfigure}%
\setcounter{subfigure}{0}

% \rotatebox[origin=c]{90}{Scale=30}
\rotatebox{90}{\,\,\, Snow+Haze}
\begin{subfigure}[t]{0.31\textwidth}
  \centering
  \includegraphics[width=1.0\linewidth]{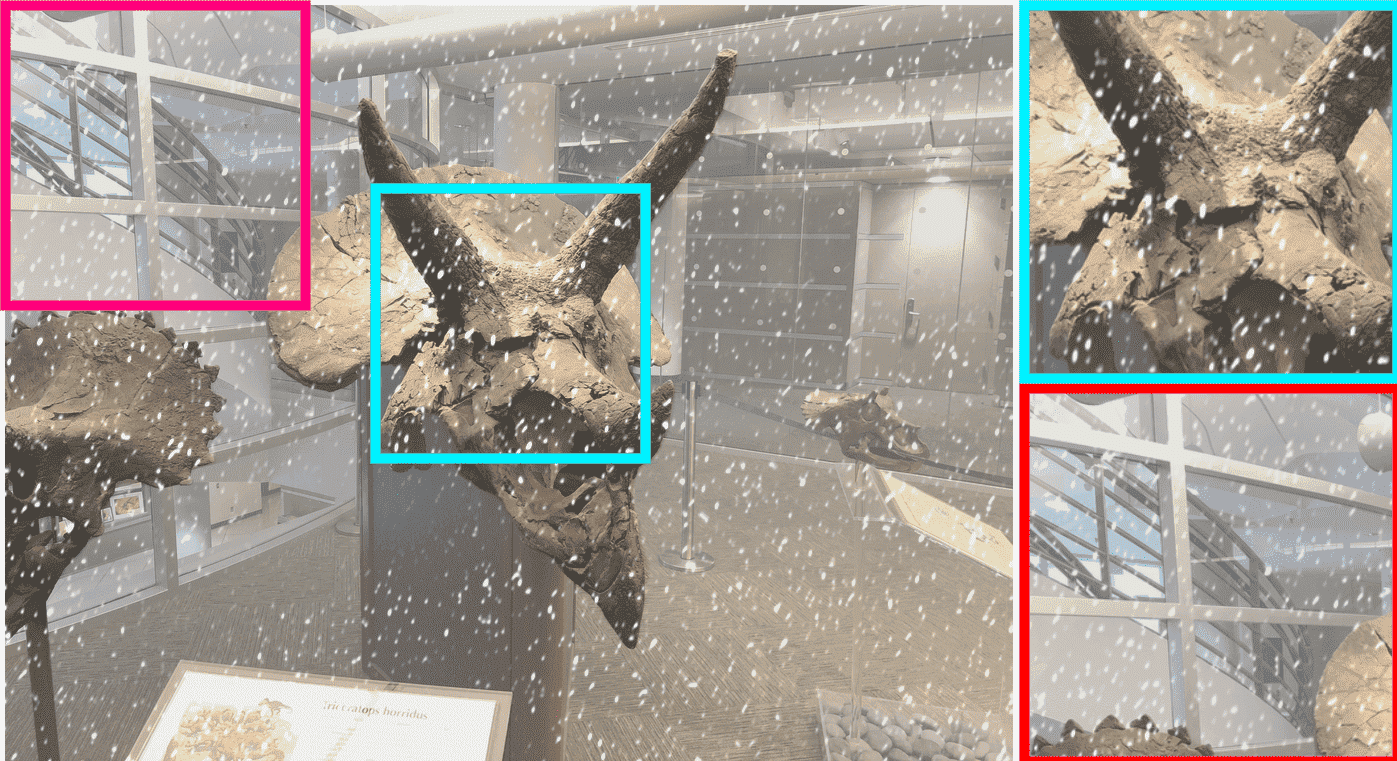}
\caption{Corrupted View}
\end{subfigure}%
\begin{subfigure}[t]{0.31\textwidth}
  \centering
  \includegraphics[width=1.0\linewidth]{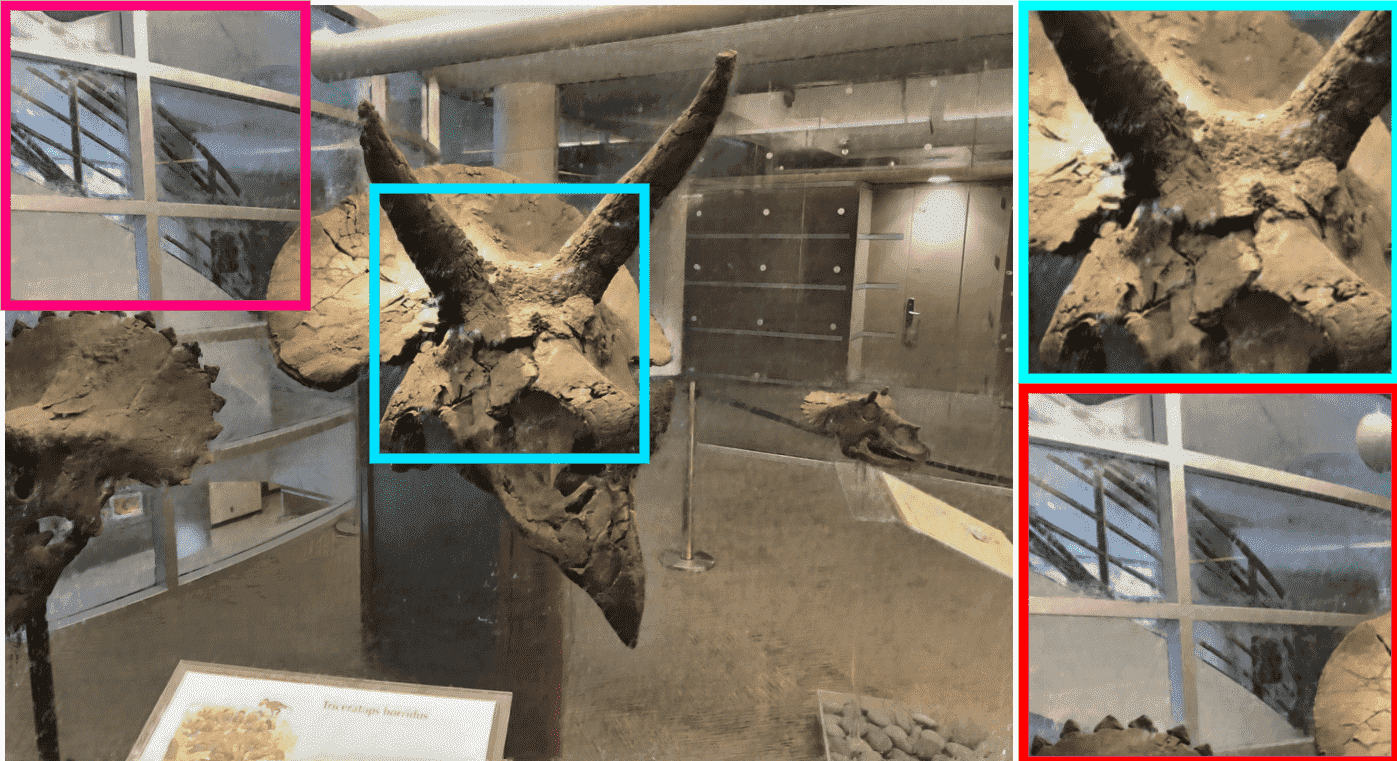}
\caption{GAURA}
\end{subfigure}%
\begin{subfigure}[t]{0.31\textwidth}
  \centering
  \includegraphics[width=1.0\linewidth]{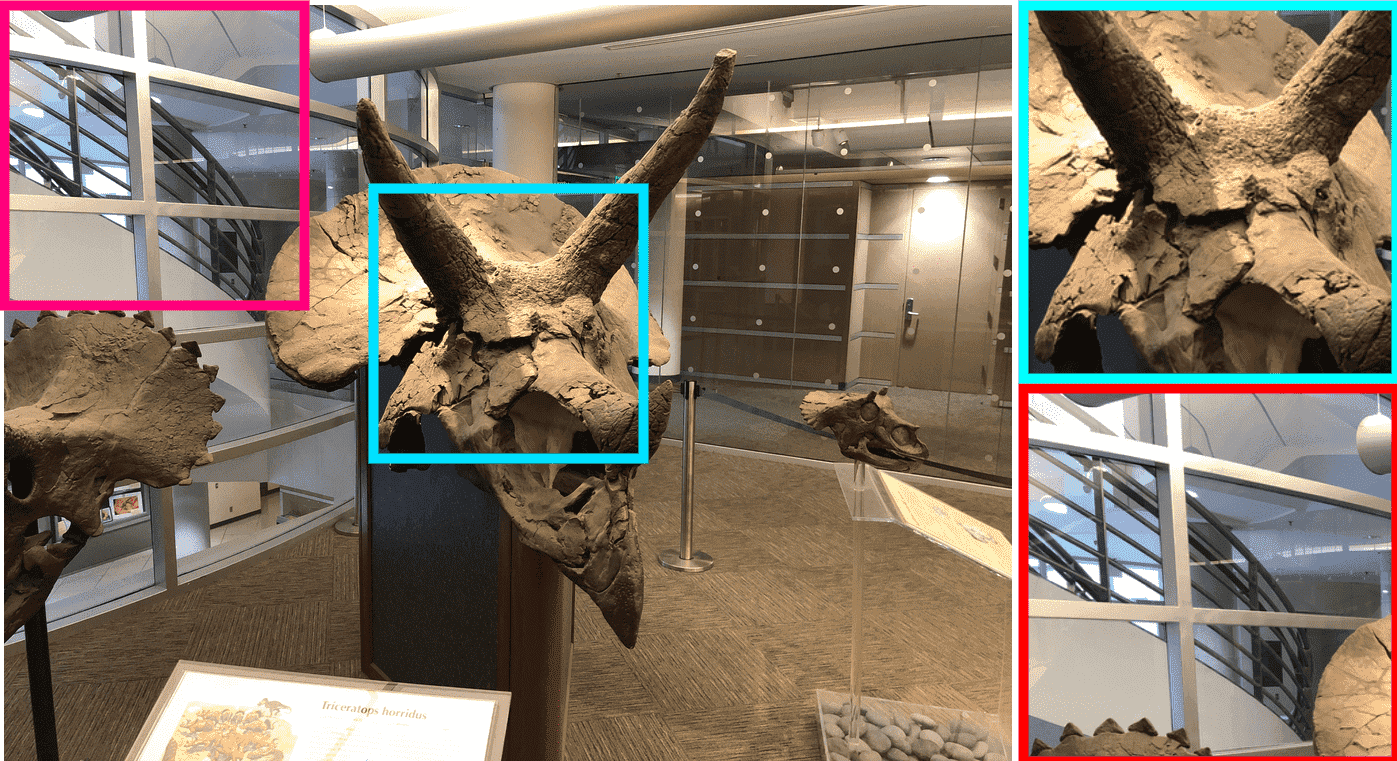}
\caption{Ground Truth}
\end{subfigure}%

\caption{Qualitative results on scenes containing multiple imperfections in the input image captures. Upon interpolating the latent codes corresponding to the individual imperfection types, GAURA can "combine" multiple image formation processes and recover the original clean scene effectively. 
% We visualize the results of our model on the multi-corruption task, i.e. scenes with more than one corruption. We achieve this by interpolating between 2 latent codes corresponding to the corruptions. Our method successfully eliminates both corruptions simultaneously from the scene and is closer to the ground truth. 
}
\vspace{-2em}
\label{fig:multicorruption}
\end{figure*}

\vspace{0.3em}
\subsection{Fine-tuning to New Degradations}
\label{sec:finetuning}

\begin{wraptable}{r}{0.45\linewidth}
% \begin{table}[t]
    \vspace{-3em}
 \caption{Quantitative results upon fine-tuning pretrained GAURA on two unseen restoration tasks - snow removal and defocus deblurring. }
  \centering
  \resizebox{0.44\columnwidth}{!}{
  \begin{tabular}{lccc|ccc}
    \toprule
    \multirow{2}{*}{Models} & \multicolumn{3}{c|}{Snow} & \multicolumn{3}{c}{Defocus}\\
    \cmidrule(r){2-7}
     & \hspace{0.2em}PSNR$\uparrow$\hspace{0.2em} & SSIM$\uparrow$\hspace{0.2em} & LPIPS$\downarrow$\hspace{0.2em} & \hspace{0.2em}PSNR$\uparrow$\hspace{0.2em} & SSIM$\uparrow$\hspace{0.2em} & LPIPS$\downarrow$\\
    \midrule
    Vanilla-GNT~\cite{gnt} & \cellcolor{black!15}21.96 & \cellcolor{black!15}0.688 & \cellcolor{black!30}0.295 & 20.95 & 0.754 & 0.283\\ 
    GNT--Restore & 20.24 & 0.608 & 0.434 & \cellcolor{black!15}21.13 & \cellcolor{black!15}0.762 & \cellcolor{black!15}0.257\\ 
    Ours & \cellcolor{black!30}22.61 & \cellcolor{black!30}0.710 & \cellcolor{black!15}0.342 & \cellcolor{black!30}21.34 & \cellcolor{black!30}0.768 & \cellcolor{black!30}0.251\\ 
    \bottomrule
  \end{tabular}
  }
  \label{tab:finetune}
    \vspace{-2em}
% \end{table}
\end{wraptable}
As discussed in Sec. \ref{sec:trainandinf}, our method allows for efficient fine-tuning to any new degradation type and we provide more details below.
We expand the Degradation Aware Latent Block for an incoming corruption type to contain an additional embedding for the given input imperfection, effectively adding less than 5\% parameters to the GAURA model. 
By freezing the remaining parameters, one can efficiently optimize GAURA on the new degradation with a limited number of scenes and training iterations. 
Simply put, our architecture design only requires a latent code to be learned that encodes degradation-specific information and can modulate the remaining parts of the network based on the input imperfection. Note that in this case, we do not utilize paired real-world data and rely only on synthetically corrupted data for supervision.  
To verify the same, we select two representative restoration tasks: defocus blur removal and desnowing and finetune on 8 scenes from LLFF's training dataset~\cite{mildenhall2019llff}, manually altered to contain these imperfections. 
For evaluation, we benchmark on the scenes from DeblurNeRF~\cite{ma2022deblur} (for defocus blur removal), synthetically altered scenes from LLFF containing snow imperfections (for desnowing), and report average scores in Table~\ref{tab:finetune}.
In addition to the baseline GNT--Restore (see Sec. \ref{sec:comparisons}), we compare against a Vanilla GNT model trained from scratch on the given degradation type. 
We observe that, indeed, the pre-trained priors learned by GAURA can be repurposed to learn any new degradation type with limited fine-tuning (in practice, as little as 10,000 steps). 
Due to the absence of benchmarks with ground truth clean views, we simply visualize our rendered outputs on real scenes containing snow and blur imperfections in Fig. \ref{fig:rain_snow_defocus} (rows 2 and 3) and our method faithfully reconstructs the clean novel view with high quality when compared to baselines. Notably, since the additional latent codes are only optimized during fine-tuning, GAURA can retain its original performance on the previous degradation types.

\vspace{5mm}
\begin{minipage}{\textwidth}
  \hspace*{-0.6cm}  
  \begin{minipage}[b]{0.6\textwidth}
    \centering
    \centering
    \includegraphics[width=1.0\linewidth]{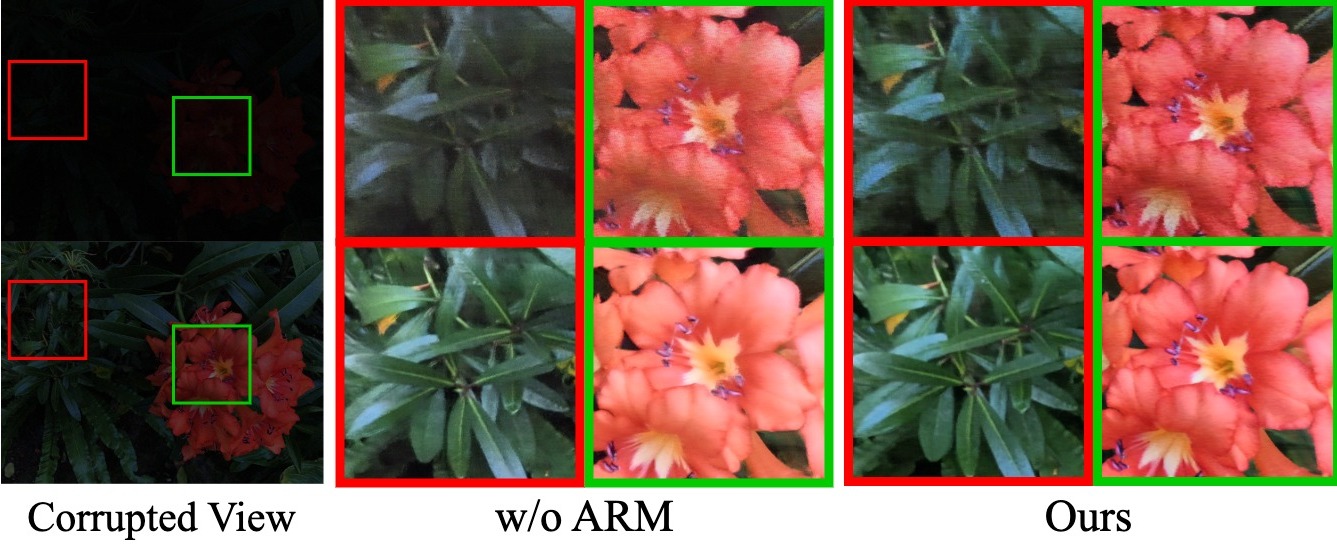}
    \vspace{-0.75cm}
    \label{fig:ablation}
    \captionof{figure}{Qualitative comparisons for ablation study.}
  \end{minipage}
  % \hfill
  \hspace*{0.1cm}  
  \begin{minipage}[b]{0.3\textwidth}
    \centering
    \captionof{table}{Ablation Study}
    % \vspace{-0.5cm}
    \resizebox{1.2\columnwidth}{!}{\begin{tabular}[b]{lccc}
    \toprule
    Models & PSNR$\uparrow$ & SSIM$\uparrow$ & LPIPS$\downarrow$\\
    \midrule
    Vanilla GNT & 17.69 & 0.704 & 0.406 \\
    \quad w/ DLM & & & \\
    \quad\quad in $\mathcal{F}_{\text{conv}}$ & 17.98 & 0.680 & 0.427 \\
    \quad\quad in $\mathcal{F}_{\text{conv}}$, $\mathcal{F}_{\text{view}}$ & 18.73 & \cellcolor{black!15}0.727 & 0.394 \\
    \quad\quad in $\mathcal{F}_{\text{conv}}$, $\mathcal{F}_{\text{view}}$, $\mathcal{F}_{\text{point}}$ & \cellcolor{black!15}19.37 & 0.714 & \cellcolor{black!15}0.363\\
    \midrule
    \quad\quad\quad w/ ARM (Ours) & \cellcolor{black!30}19.91 & \cellcolor{black!30}0.736 & \cellcolor{black!30}0.352\\
    \bottomrule
  \end{tabular}}
  \label{tab:ablation}
    \vspace{0.5cm}
    \end{minipage}
\end{minipage}

\subsection{Extension to Multiple Degradations}
\label{sec:multi-corruption}
In a practical scenario, there are often cases where multiple imperfections are present in the input image captures. 
Explicitly modeling such a complex image formation process is challenging, making it difficult to naively extend existing NeRF restoration techniques to such a scenario. 
However, GAURA simply embeds degradation-specific priors onto individual latent codes that can be ``interpolated'' to combine their individual image formation processes. 
Formally, given two degradation types $\Mat{D}_{1}$, and $\Mat{D}_{2}$ present in the input images, we query the DLM blocks with a new combined latent code $\Mat{L} = \alpha \times \Mat{L}_{\Mat{D}_{1}} + (1 - \alpha) \times \Mat{L}_{\Mat{D}_{2}}$, where $\alpha$ is a scaling term. 
We visualize qualitative results in Fig. \ref{fig:multicorruption}, where the input images of a scene are manually altered to contain more than one imperfection type. 
By manually tweaking $\alpha$, our pretrained GAURA can effectively remove multiple imperfections and synthesize clean novel views.
% \textcolor{blue}{Should we mention what are the best values of alpha we got for both the rain+dark and snow+haze?}

\subsection{Ablation Studies}
\label{sec:ablation}
We conduct the following ablations to validate our architectural design and provide quantitative results in Table~\ref{tab:ablation}.
In each case, we train the model following the details mentioned in Sec. \ref{sec:implementation}, and then evaluate performance on a single representative restoration task, i.e. low-light enhancement on the AlethNeRF dataset~\cite{cui_aleth_nerf}. 
First, we train the original GNT~\cite{gnt}, as-is without any degradation-aware modules to automatically render and restore novel views (indicated as Vanilla GNT). 
Without explicitly encoding degradation-specific information, GNT fails to restore novel views reasonably. 
By inserting Degradation-aware Latent Blocks (from Sec. \ref{sec:degradationmodule}) into several parts of the network ($\mathcal{F}_{\text{conv}}$, $\mathcal{F}_{\text{view}}$, $\mathcal{F}_{\text{point}}$), GAURA can now learn to implicitly embed degradation-specific information into its learned rendering process. 
Finally, our Adaptive Residual block captures variations within each degradation type that can further improve performance (note the color differences in the case w/o ARM in Fig. \ref{fig:ablation}).

\vspace{-1em}

\section{Discussion}
\label{sec:discussion}

\vspace{-0.2em}
\paragraph{\textbf{Limitations and Future Work. }}
While GAURA can successfully perform novel view synthesis using input images captured in degraded environments, prior information about the imperfection type is still required. 
While this is not difficult to obtain, an ideal pipeline should be able to automatically restore any corruption type observed in the input images (formally called blind image restoration).
Our method relies on epipolar-based rendering techniques, and therefore, its potential is capped by its flaws; for example, our method may not handle sparse 360-degree scenes or objects with complex light transport where epipolar geometry no longer holds. 
The recent advancements in 3D reconstruction use an explicit representation, e.g., 3D Gaussians, that can encode high-frequency details and enable faster rendering. 
Therefore, incorporating our degradation-aware modules into such a pipeline could be an interesting direction. 
Lastly, (and more importantly) the need for benchmarks to evaluate methods on different real-world restoration tasks is glaring, and future work must focus on this direction.

\vspace{-0.4em}
\paragraph{\textbf{Conclusion. }}
We propose GAURA, a pipeline that combines degradation-aware modules with epipolar-based rendering to perform novel view synthesis from imperfect input images, showcasing generalization to any scene and corruption type. 
Our method demonstrates that several of the inductive biases that are necessary for rendering and restoration (e.g. explicit modeling of physical degradation process, hard-coded rendering equation) can be implicitly encoded into neural networks. 
We believe this insight could open up new possibilities for a universal 3D restoration and rendering model. 

\clearpage  % TODO REVIEW/FINAL: This \clearpage needs to be removed from both review and camera-ready versions.

\bibliographystyle{splncs04}
\bibliography{main}

\begin{thebibliography}{10}
\providecommand{\url}[1]{\texttt{#1}}
\providecommand{\urlprefix}{URL }
\providecommand{\doi}[1]{https://doi.org/#1}

\bibitem{aharon2023hypernetwork}
Aharon, S., Ben-Artzi, G.: Hypernetwork-based adaptive image restoration. In: ICASSP 2023-2023 IEEE International Conference on Acoustics, Speech and Signal Processing (ICASSP). pp.~1--5. IEEE (2023)

\bibitem{barron2022mip}
Barron, J.T., Mildenhall, B., Verbin, D., Srinivasan, P.P., Hedman, P.: Mip-nerf 360: Unbounded anti-aliased neural radiance fields. In: Proceedings of the IEEE/CVF Conference on Computer Vision and Pattern Recognition. pp. 5470--5479 (2022)

\bibitem{DerainRLNet2021}
Chen, C., Li, H.: Robust representation learning with feedback for single image deraining. In: IEEE/CVF Conference on Computer Vision and Pattern Recognition (CVPR). pp. 7742--7751 (2021)

\bibitem{chen2019learning}
Chen, Z., Zhang, H.: Learning implicit fields for generative shape modeling. In: Proceedings of the IEEE/CVF Conference on Computer Vision and Pattern Recognition. pp. 5939--5948 (2019)

\bibitem{cheng2023drm}
Cheng, Y., Shao, M., Wan, Y., Wang, C., Zuo, W.: Drm-ir: Task-adaptive deep unfolding network for all-in-one image restoration. arXiv preprint arXiv:2307.07688  (2023)

\bibitem{cho2021rethinking}
Cho, S.J., Ji, S.W., Hong, J.P., Jung, S.W., Ko, S.J.: Rethinking coarse-to-fine approach in single image deblurring. In: Proceedings of the IEEE/CVF international conference on computer vision. pp. 4641--4650 (2021)

\bibitem{cui2023selective}
Cui, Y., Tao, Y., Bing, Z., Ren, W., Gao, X., Cao, X., Huang, K., Knoll, A.: Selective frequency network for image restoration. In: The Eleventh International Conference on Learning Representations (2023)

\bibitem{cui_aleth_nerf}
Cui, Z., Gu, L., Sun, X., Ma, X., Qiao, Y., Harada, T.: Aleth-nerf: Illumination adaptive nerf with concealing field assumption. In: Proceedings of the AAAI Conference on Artificial Intelligence (2024)

\bibitem{genova2020local}
Genova, K., Cole, F., Sud, A., Sarna, A., Funkhouser, T.: Local deep implicit functions for 3d shape. In: Proceedings of the IEEE/CVF Conference on Computer Vision and Pattern Recognition. pp. 4857--4866 (2020)

\bibitem{gupta2024gsn}
Gupta, V., Goel, R., Dhawal, S., Narayanan, P.: Gsn: Generalisable segmentation in neural radiance field. In: Proceedings of the AAAI Conference on Artificial Intelligence. vol.~38, pp. 2013--2021 (2024)

\bibitem{ha2016hypernetworks}
Ha, D., Dai, A., Le, Q.V.: Hypernetworks. arXiv preprint arXiv:1609.09106  (2016)

\bibitem{jiang2023autodir}
Jiang, Y., Zhang, Z., Xue, T., Gu, J.: Autodir: Automatic all-in-one image restoration with latent diffusion. arXiv preprint arXiv:2310.10123  (2023)

\bibitem{johari2022geonerf}
Johari, M.M., Lepoittevin, Y., Fleuret, F.: Geonerf: Generalizing nerf with geometry priors. In: Proceedings of the IEEE/CVF Conference on Computer Vision and Pattern Recognition. pp. 18365--18375 (2022)

\bibitem{koschmieder1924theorie}
Koschmieder, H.: Theorie der horizontalen sichtweite. Beitrage zur Physik der freien Atmosphare pp. 33--53 (1924)

\bibitem{lamba2023real}
Lamba, M., Kumar, M., Mitra, K.: Real-time restoration of dark stereo images. In: Proceedings of the IEEE/CVF Winter Conference on Applications of Computer Vision. pp. 4914--4924 (2023)

\bibitem{land1977retinex}
Land, E.H.: The retinex theory of color vision. Scientific american  \textbf{237}(6),  108--129 (1977)

\bibitem{levy2023seathru}
Levy, D., Peleg, A., Pearl, N., Rosenbaum, D., Akkaynak, D., Korman, S., Treibitz, T.: Seathru-nerf: Neural radiance fields in scattering media. In: Proceedings of the IEEE/CVF Conference on Computer Vision and Pattern Recognition. pp. 56--65 (2023)

\bibitem{li2021you}
Li, B., Gou, Y., Gu, S., Liu, J.Z., Zhou, J.T., Peng, X.: You only look yourself: Unsupervised and untrained single image dehazing neural network. International Journal of Computer Vision  \textbf{129},  1754--1767 (2021)

\bibitem{airnet}
Li, B., Liu, X., Hu, P., Wu, Z., Lv, J., Peng, X.: All-in-one image restoration for unknown corruption. In: Proceedings of the IEEE/CVF Conference on Computer Vision and Pattern Recognition. pp. 17452--17462 (2022)

\bibitem{li2020all}
Li, R., Tan, R.T., Cheong, L.F.: All in one bad weather removal using architectural search. In: Proceedings of the IEEE/CVF conference on computer vision and pattern recognition. pp. 3175--3185 (2020)

\bibitem{liu2018learning}
Liu, R., He, Y., Cheng, S., Fan, X., Luo, Z.: Learning collaborative generation correction modules for blind image deblurring and beyond. In: Proceedings of the 26th ACM international conference on Multimedia. pp. 1921--1929 (2018)

\bibitem{lou2023simhaze}
Lou, Z., Xu, H., Mu, F., Liu, Y., Zhang, X., Shang, L., Li, J., Guan, B., Li, Y., Hu, Y.H.: Simhaze: game engine simulated data for real-world dehazing. arXiv preprint arXiv:2305.16481  (2023)

\bibitem{daclip}
Luo, Z., Gustafsson, F.K., Zhao, Z., Sj{\"o}lund, J., Sch{\"o}n, T.B.: Controlling vision-language models for universal image restoration. arXiv preprint arXiv:2310.01018  (2023)

\bibitem{lv2021attention}
Lv, F., Li, Y., Lu, F.: Attention guided low-light image enhancement with a large scale low-light simulation dataset. International Journal of Computer Vision  \textbf{129}(7),  2175--2193 (2021)

\bibitem{ma2022deblur}
Ma, L., Li, X., Liao, J., Zhang, Q., Wang, X., Wang, J., Sander, P.V.: Deblur-nerf: Neural radiance fields from blurry images. In: Proceedings of the IEEE/CVF Conference on Computer Vision and Pattern Recognition. pp. 12861--12870 (2022)

\bibitem{ma2022toward}
Ma, L., Ma, T., Liu, R., Fan, X., Luo, Z.: Toward fast, flexible, and robust low-light image enhancement. In: Proceedings of the IEEE/CVF Conference on Computer Vision and Pattern Recognition. pp. 5637--5646 (2022)

\bibitem{Mildenhall_2022_CVPR}
Mildenhall, B., Hedman, P., Martin-Brualla, R., Srinivasan, P.P., Barron, J.T.: Nerf in the dark: High dynamic range view synthesis from noisy raw images. In: Proceedings of the IEEE/CVF Conference on Computer Vision and Pattern Recognition (CVPR). pp. 16190--16199 (June 2022)

\bibitem{mildenhall2019llff}
Mildenhall, B., Srinivasan, P.P., Ortiz-Cayon, R., Kalantari, N.K., Ramamoorthi, R., Ng, R., Kar, A.: Local light field fusion: Practical view synthesis with prescriptive sampling guidelines. ACM Transactions on Graphics (TOG)  (2019)

\bibitem{mildenhall2020nerf}
Mildenhall, B., Srinivasan, P.P., Tancik, M., Barron, J.T., Ramamoorthi, R., Ng, R.: Nerf: Representing scenes as neural radiance fields for view synthesis. In: Computer Vision--ECCV 2020: 16th European Conference, Glasgow, UK, August 23--28, 2020, Proceedings, Part I. pp. 405--421 (2020)

\bibitem{niemeyer2020differentiable}
Niemeyer, M., Mescheder, L., Oechsle, M., Geiger, A.: Differentiable volumetric rendering: Learning implicit 3d representations without 3d supervision. In: Proceedings of the IEEE/CVF Conference on Computer Vision and Pattern Recognition. pp. 3504--3515 (2020)

\bibitem{potlapalli2023promptir}
Potlapalli, V., Zamir, S.W., Khan, S., Khan, F.S.: Promptir: Prompting for all-in-one blind image restoration. arXiv preprint arXiv:2306.13090  (2023)

\bibitem{rim_2020_ECCV}
Rim, J., Lee, H., Won, J., Cho, S.: Real-world blur dataset for learning and benchmarking deblurring algorithms. In: Proceedings of the European Conference on Computer Vision (ECCV) (2020)

\bibitem{sitzmann2019scene}
Sitzmann, V., Zollh{\"o}fer, M., Wetzstein, G.: Scene representation networks: Continuous 3d-structure-aware neural scene representations. Advances in Neural Information Processing Systems  \textbf{32} (2019)

\bibitem{suhail2022generalizable}
Suhail, M., Esteves, C., Sigal, L., Makadia, A.: Generalizable patch-based neural rendering. In: European Conference on Computer Vision. pp. 156--174. Springer (2022)

\bibitem{sun2013edge}
Sun, L., Cho, S., Wang, J., Hays, J.: Edge-based blur kernel estimation using patch priors. In: IEEE international conference on computational photography (ICCP). pp.~1--8. IEEE (2013)

\bibitem{tian2020image}
Tian, C., Xu, Y., Zuo, W.: Image denoising using deep cnn with batch renormalization. Neural Networks  \textbf{121},  461--473 (2020)

\bibitem{gnt}
Varma, M., Wang, P., Chen, X., Chen, T., Venugopalan, S., Wang, Z.: Is attention all that nerf needs? In: The Eleventh International Conference on Learning Representations (2023)

\bibitem{verbin2021ref}
Verbin, D., Hedman, P., Mildenhall, B., Zickler, T., Barron, J.T., Srinivasan, P.P.: Ref-nerf: Structured view-dependent appearance for neural radiance fields. arXiv preprint arXiv:2112.03907  (2021)

\bibitem{wang2022generalizable}
Wang, D., Cui, X., Salcudean, S., Wang, Z.J.: Generalizable neural radiance fields for novel view synthesis with transformer. arXiv preprint arXiv:2206.05375  (2022)

\bibitem{wang2023lighting}
Wang, H., Xu, X., Xu, K., Lau, R.W.: Lighting up nerf via unsupervised decomposition and enhancement. In: Proceedings of the IEEE/CVF International Conference on Computer Vision. pp. 12632--12641 (2023)

\bibitem{wang2021neus}
Wang, P., Liu, L., Liu, Y., Theobalt, C., Komura, T., Wang, W.: Neus: Learning neural implicit surfaces by volume rendering for multi-view reconstruction. arXiv preprint arXiv:2106.10689  (2021)

\bibitem{wang2023badnerf}
Wang, P., Zhao, L., Ma, R., Liu, P.: {BAD-NeRF: Bundle Adjusted Deblur Neural Radiance Fields}. In: Proceedings of the IEEE/CVF Conference on Computer Vision and Pattern Recognition (CVPR). pp. 4170--4179 (June 2023)

\bibitem{wang2021ibrnet}
Wang, Q., Wang, Z., Genova, K., Srinivasan, P.P., Zhou, H., Barron, J.T., Martin-Brualla, R., Snavely, N., Funkhouser, T.: Ibrnet: Learning multi-view image-based rendering. In: Proceedings of the IEEE/CVF Conference on Computer Vision and Pattern Recognition. pp. 4690--4699 (2021)

\bibitem{wang2022removing}
Wang, Y., Ma, C., Liu, J.: Removing rain streaks via task transfer learning. arXiv preprint arXiv:2208.13133  (2022)

\bibitem{wang2022low}
Wang, Y., Wan, R., Yang, W., Li, H., Chau, L.P., Kot, A.: Low-light image enhancement with normalizing flow. In: Proceedings of the AAAI conference on artificial intelligence. vol.~36, pp. 2604--2612 (2022)

\bibitem{Chen2018Retinex}
Wei, Chen, W., Wenjing, Y., Wenhan, L., Jiaying: Deep retinex decomposition for low-light enhancement. In: British Machine Vision Conference (2018)

\bibitem{Wizadwongsa2021NeX}
Wizadwongsa, S., Phongthawee, P., Yenphraphai, J., Suwajanakorn, S.: Nex: Real-time view synthesis with neural basis expansion. In: Proceedings of the IEEE/CVF Conference on Computer Vision and Pattern Recognition. pp. 8534--8543 (2021)

\bibitem{yang2017deep}
Yang, W., Tan, R.T., Feng, J., Liu, J., Guo, Z., Yan, S.: Deep joint rain detection and removal from a single image. In: Proceedings of the IEEE conference on computer vision and pattern recognition. pp. 1357--1366 (2017)

\bibitem{zhang2018densely}
Zhang, H., Patel, V.M.: Densely connected pyramid dehazing network. In: Proceedings of the IEEE conference on computer vision and pattern recognition. pp. 3194--3203 (2018)

\bibitem{zhang2020deblurring}
Zhang, K., Luo, W., Zhong, Y., Ma, L., Stenger, B., Liu, W., Li, H.: Deblurring by realistic blurring. In: Proceedings of the IEEE/CVF Conference on Computer Vision and Pattern Recognition. pp. 2737--2746 (2020)

\bibitem{zhang2018perceptual}
Zhang, R., Isola, P., Efros, A.A., Shechtman, E., Wang, O.: The unreasonable effectiveness of deep features as a perceptual metric. In: CVPR (2018)

\bibitem{zhang2021learning}
Zhang, X., Dong, H., Pan, J., Zhu, C., Tai, Y., Wang, C., Li, J., Huang, F., Wang, F.: Learning to restore hazy video: A new real-world dataset and a new method. In: Proceedings of the IEEE/CVF Conference on Computer Vision and Pattern Recognition. pp. 9239--9248 (2021)

\end{thebibliography}

\appendix
% \title{GAURA: Generalizable Approach for Unified Restoration and Rendering of Arbitrary Views\\\em Supplementary Material}
% \titlerunning{GAURA}

% \maketitle
% \begin{itemize}
%     \item Supplementary for "" Write Paper Name and Paper ID - Done
%     \item In synthetic gen, bold the degradations - Done
%     \item In results, make the degradations name bold or italics. - Done 
%     \item In the main paper, we just choose the best method but here we give results for all the all-in-one methods 
%     \item refer to figures properly
%     \item Mention that it is real data in figure caption and synthetic in LLFF-Corrupted Figure
%     \item Remove Multi-Corruption Results and make a seperate figure
%     \item In the left make the subcaptions on the left and label properly. 
% \end{itemize}

% \section{Preliminaries}
% \paragraph{Self Attention and Transformers. }

% \paragraph{Neural Radiance Fields. }

% \paragraph{Generalizable NeRF Transformer (GNT)}

% \section{Video Results}

% This supplementary material comprises video results demonstrating scenes affected by various forms of degradation. We conduct a side-by-side comparison with the comprehensive all-in-one baseline methods. 
\newpage

\section{More Details on Model Architecture}

\begin{figure}
\centering
\begin{subfigure}{.5\textwidth}
  \centering
  \includegraphics[width=\linewidth]{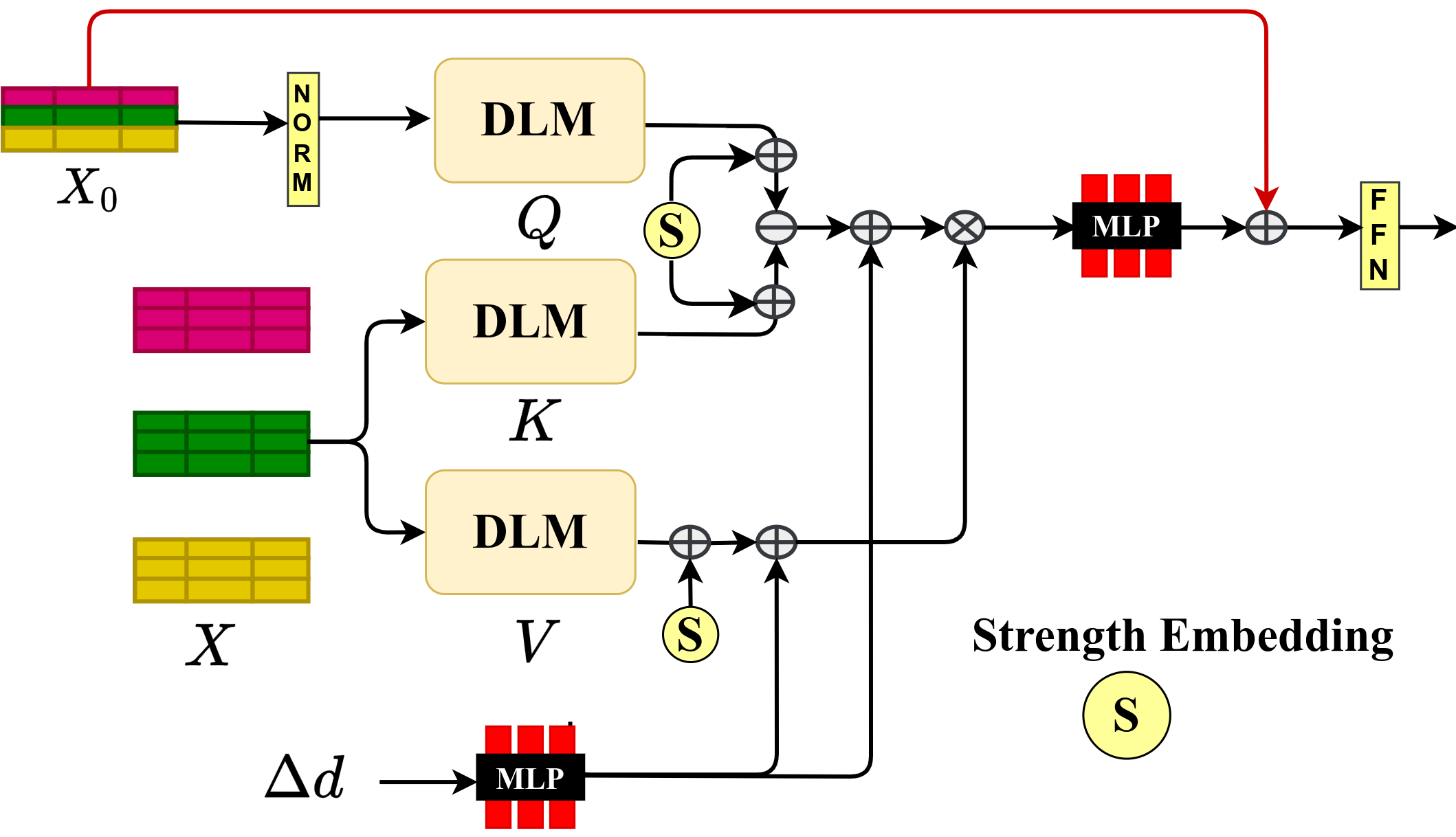}
  \caption{View Transformer}
  \label{fig:sub1}
\end{subfigure}%
\begin{subfigure}{.5\textwidth}
  \centering
  \includegraphics[width=\linewidth]{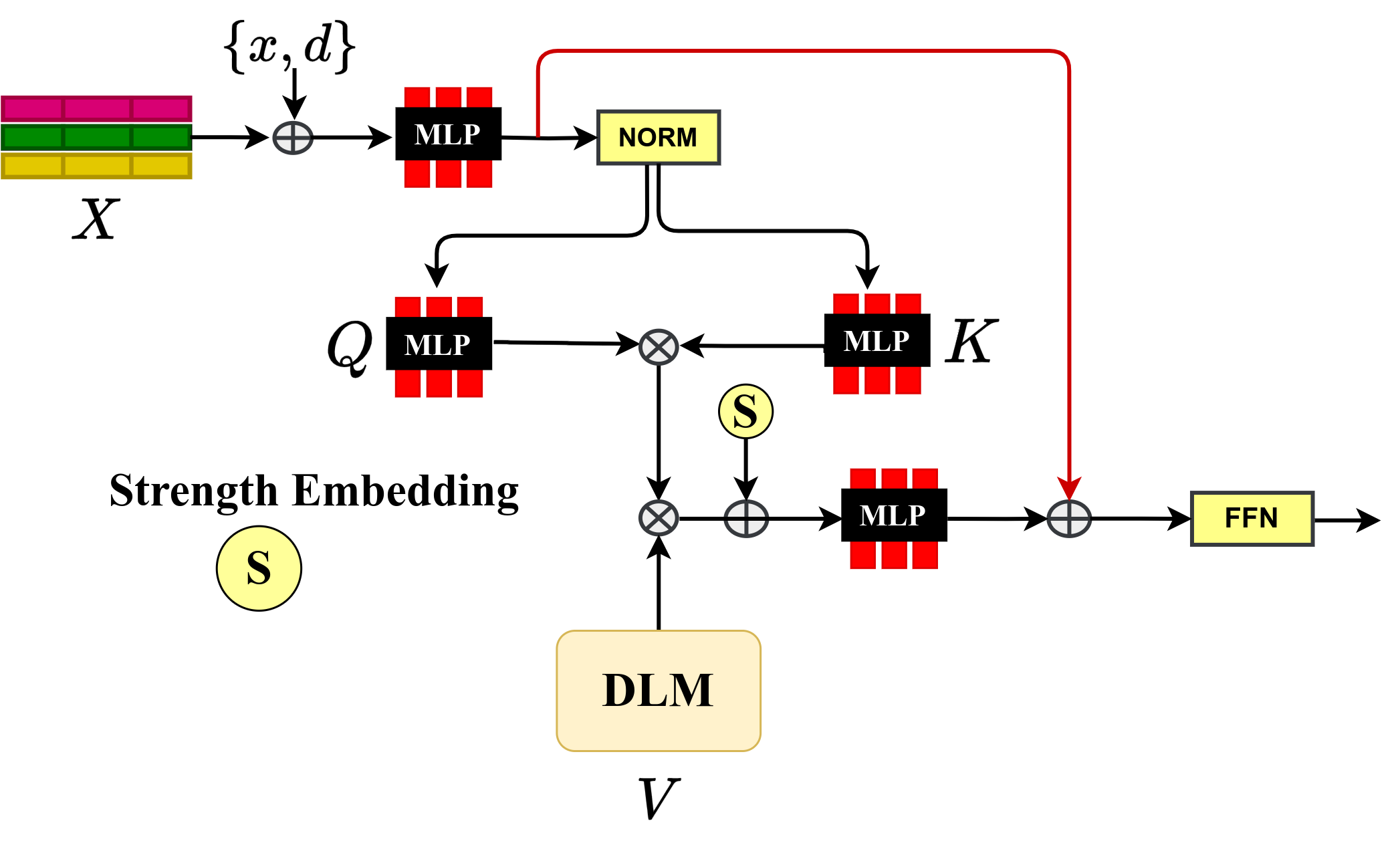}
  \caption{Ray Transformer}
  \label{fig:sub2}
\end{subfigure}
\caption{A more detailed representation of the DLM blocks in the View and Ray transformer of GAURA}
\label{fig:view}
\end{figure}

\noindent In Figure \ref{fig:view}, we elaborate on our approach, GAURA. As mentioned in the main paper, specifically, within the View Transformer, we substitute the conventional query, key, and value Multi-Layer Perceptrons (MLPs) with the Degradation Latent Module. This substitution is crucial as the inputs to these MLPs are degradation-dependent, necessitating specific degradation priors for effective restoration. However, in the ray transformer, we only tailor the value matrix to be degradation-specific, as the query-key attention mechanism inherently captures depth-information while lacking detailed appearance features crucial for restoration. Additionally, it is noteworthy that in the View Transformer, subtraction attention is employed due to computational constraints, whereas in the ray transformer, dot-product attention is utilized, as mentioned in \cite{gnt}.

\section{Synthetic Data Generation. }

In this section, we provide a comprehensive overview of the generation process utilized to synthetically introduce corruption to clean images. For the \textit{Low-Light} task, we implement the methodology outlined in \cite{lamba2023real}. Initially, the RGB image is converted to a Linear RGB scale, and all subsequent manipulations are performed within this color space. Subsequently, the image is appropriately downscaled, followed by the addition of Heteroscedastic noise. We vary the scaling factor within the range of 8 to 30 to encompass a broad spectrum of low-light scenarios. For the generation of \textit{Haze}, we employ the straightforward Koschmieder model \cite{koschmieder1924theorie}, which mimics image degradation caused by scattering and ambient light. The equation governing this model is provided as follows:
\begin{align}
    \label{eqn:physics}
    I(i) = J(i)T_{B} (i) + (1 - T_{B} (i))A
\end{align}
In the equation, $J$, $T_{B}$, and $A$ represent the scene radiance, backscatter transmission map, and global background light, respectively. We employ this equation for its simplicity, facilitating faster inference. Beta and global background light are set within the ranges of 1 to 5 and 125 to 200, respectively. To simulate motion blur and defocus blur, we utilize basic OpenCV functions. For \textit{Motion Blurring}, we adjust the kernel size to modulate the intensity of blur while simultaneously varying the direction of blur to simulate motion in different directions. Accordingly, we set kernel values ranging from 2 to 6 and angles for direction ranging from 0 to 180 degrees. In \textit{Defocus Blur}, we manipulate kernel values from 5 to 11 and randomly apply defocusing to either the foreground or background. It is important to note that we follow a similar approach as outlined in \cite{ma2022deblur}, which introduces varying degrees of blur in each captured image. 

\begin{figure*}[t]
\centering
\captionsetup[subfigure]{labelformat=empty}
% \rotatebox[origin=lc]{90}{\scriptsize{Low-Light}}
\rotatebox{90}{\quad\,\, \tiny{Low-Light}}
\begin{subfigure}[t]{0.24\textwidth}
  \centering
  \includegraphics[width=1.0\linewidth]{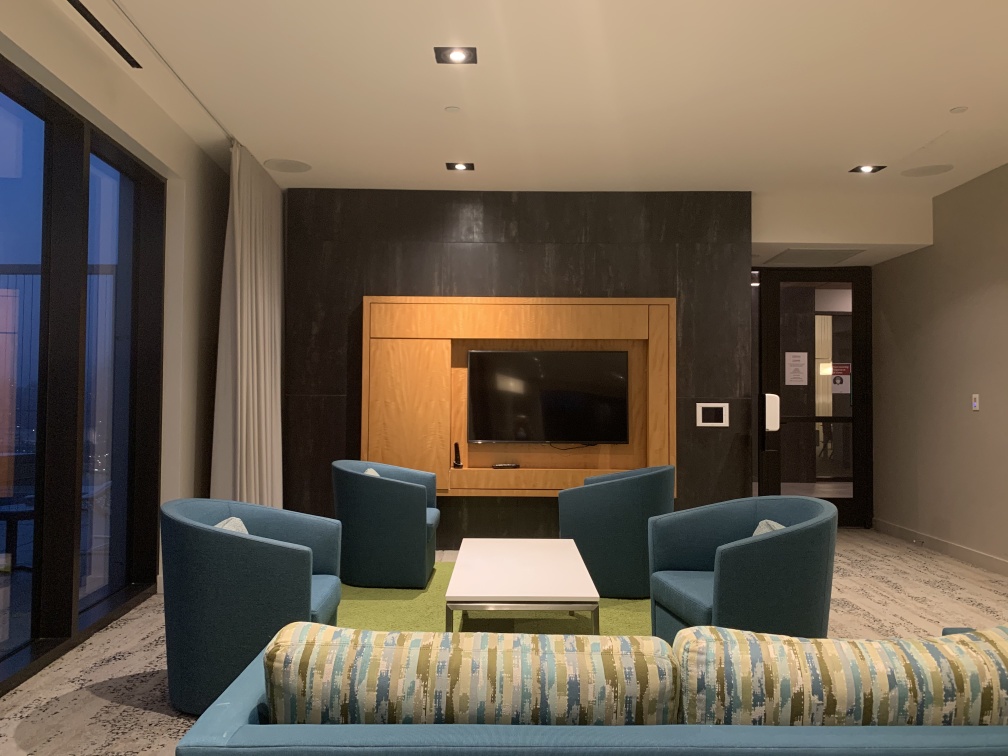}
\end{subfigure}%
\begin{subfigure}[t]{0.24\textwidth}
  \centering
  \includegraphics[width=1.0\linewidth]{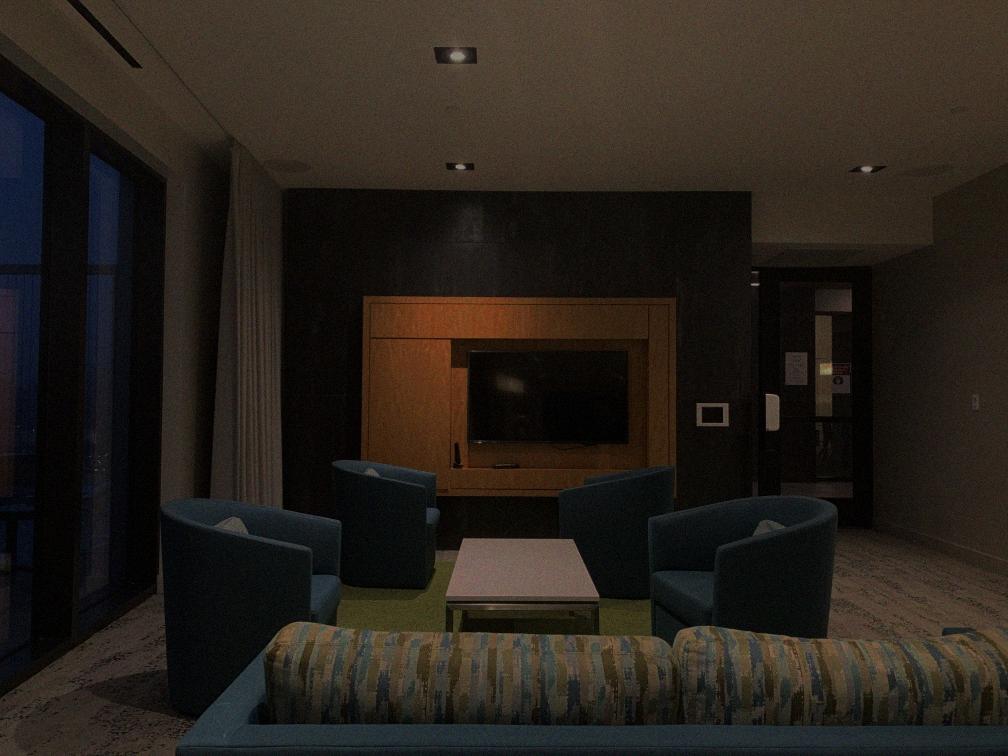}
\end{subfigure}%
\begin{subfigure}[t]{0.24\textwidth}
  \centering
  \includegraphics[width=1.0\linewidth]{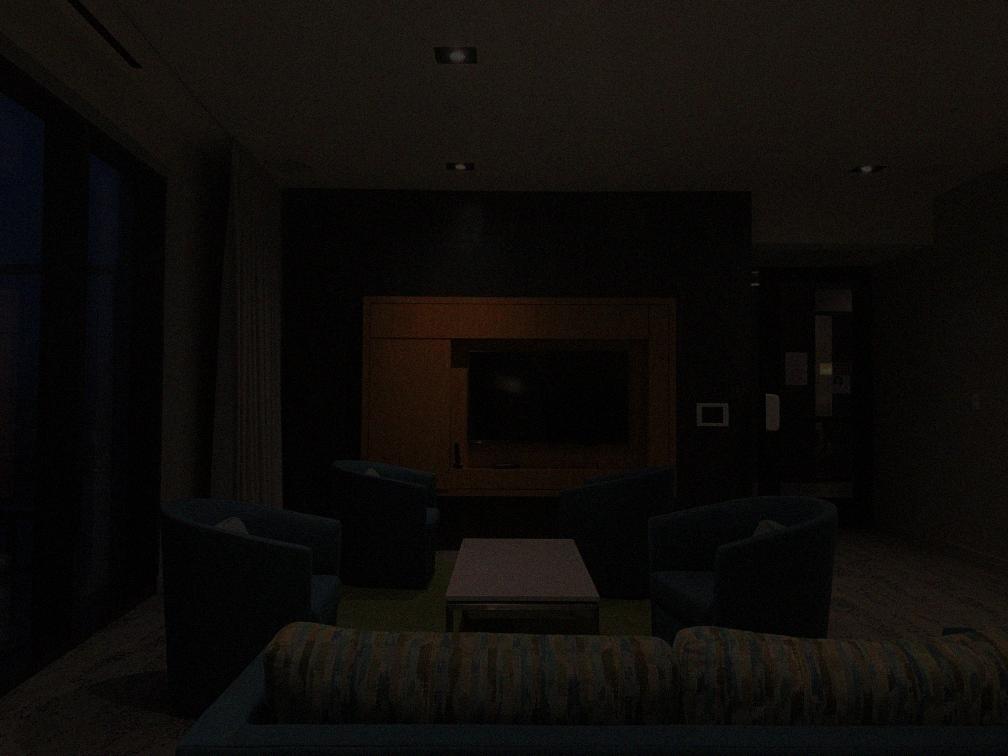}
\end{subfigure}%
\begin{subfigure}[t]{0.24\textwidth}
  \centering
  \includegraphics[width=1.0\linewidth]{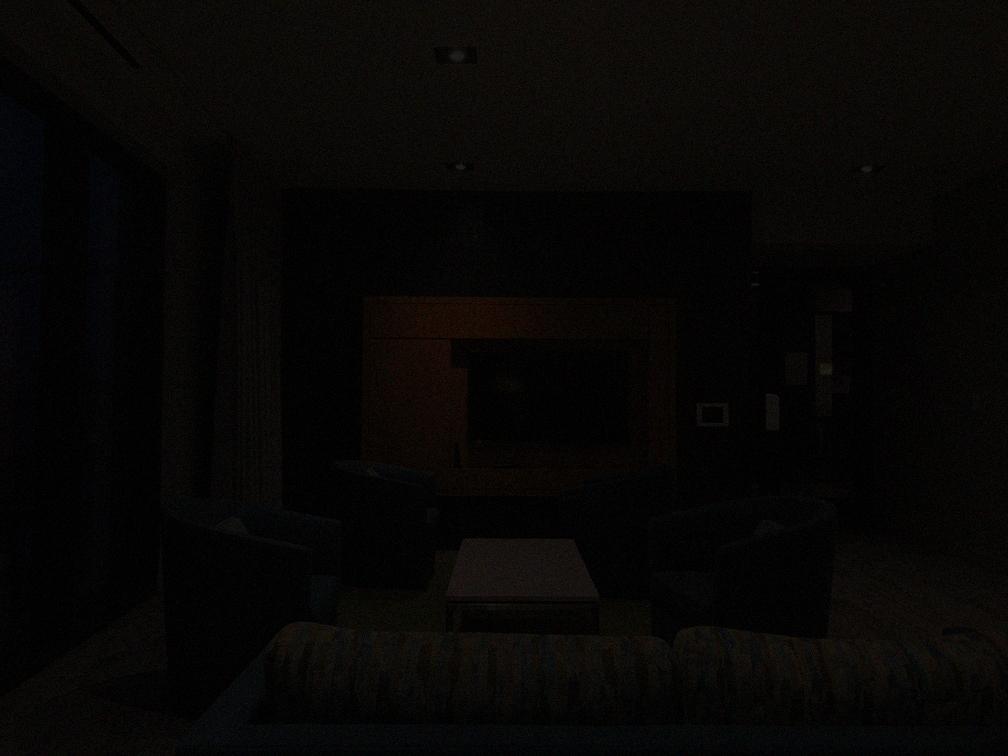}
\end{subfigure}%

\setcounter{subfigure}{0}

\rotatebox{90}{\quad\,\, \tiny{Motion-Blur}}
\begin{subfigure}[t]{0.24\textwidth}
  \centering
  \includegraphics[width=1.0\linewidth]{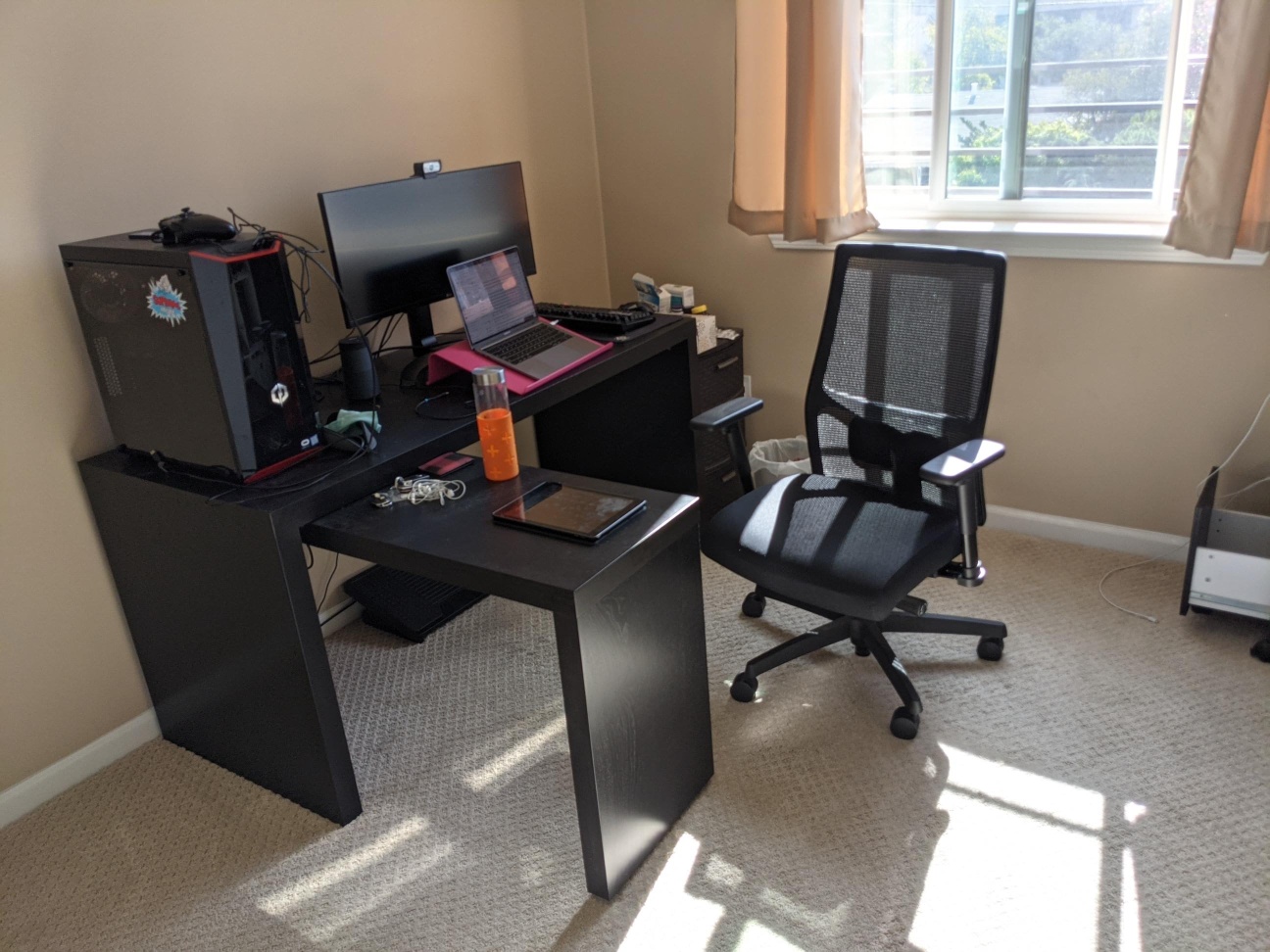}
\end{subfigure}%
\begin{subfigure}[t]{0.24\textwidth}
  \centering
  \includegraphics[width=1.0\linewidth]{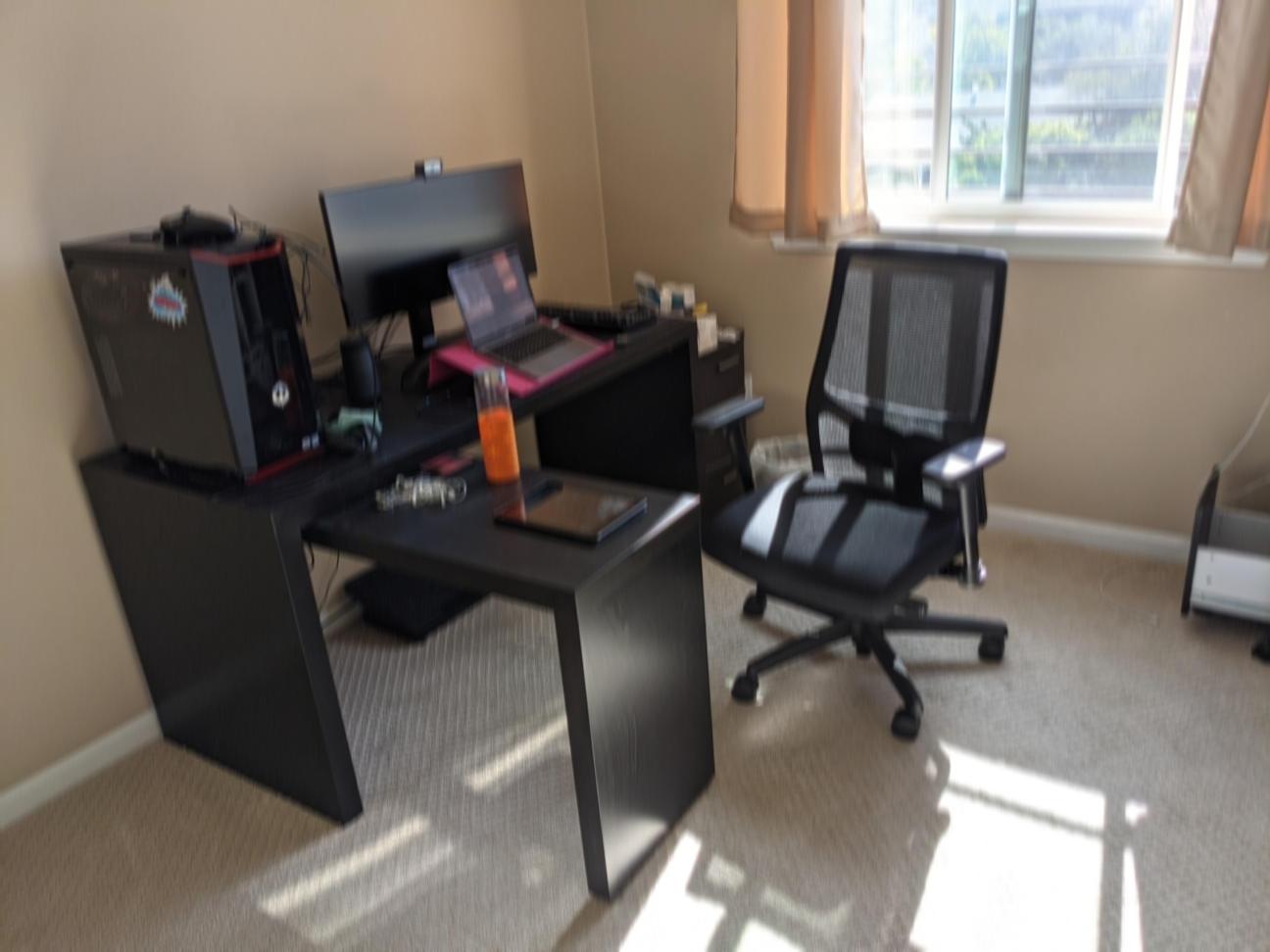}
\end{subfigure}%
\begin{subfigure}[t]{0.24\textwidth}
  \centering
  \includegraphics[width=1.0\linewidth]{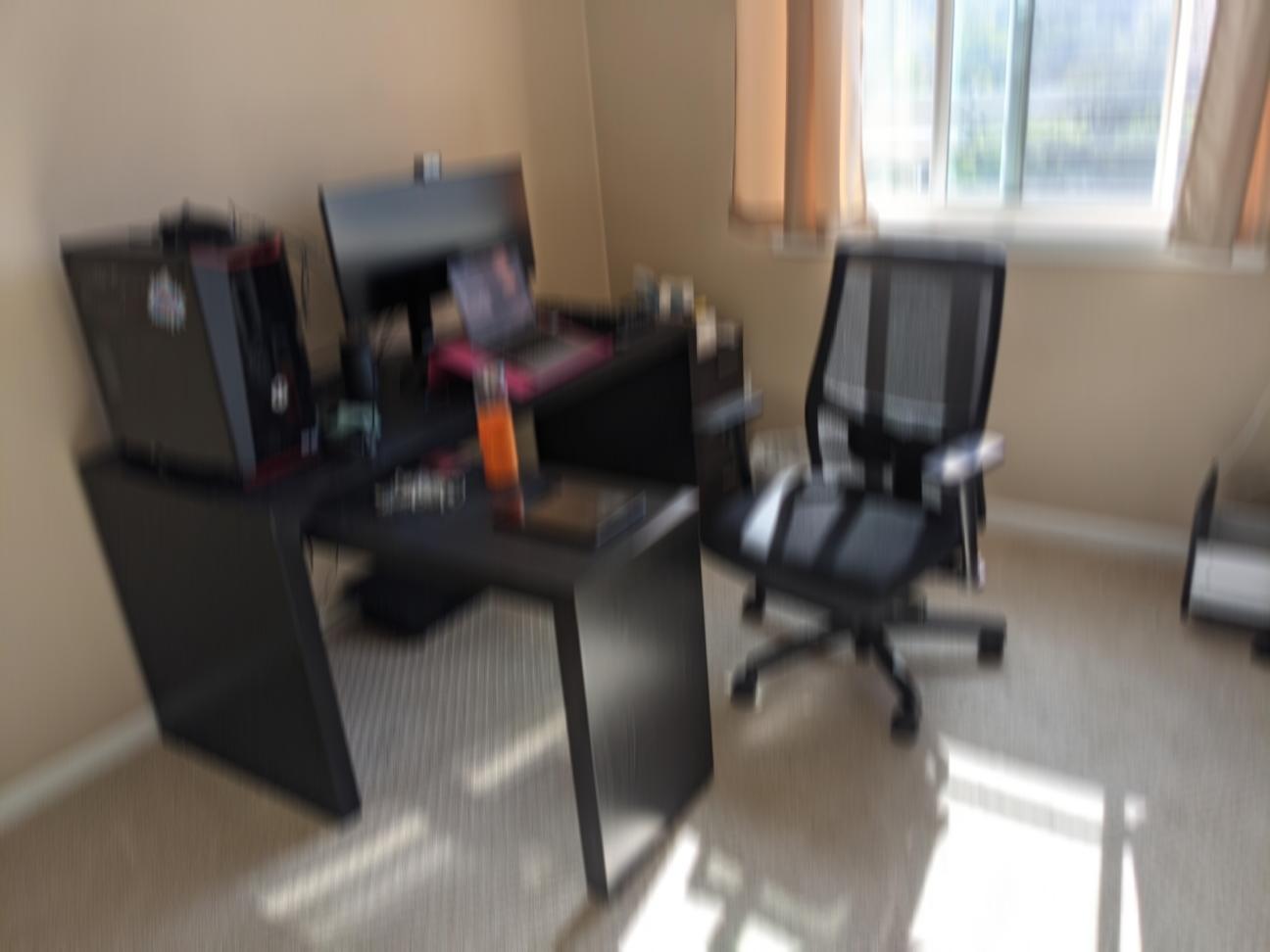}
\end{subfigure}%
\begin{subfigure}[t]{0.24\textwidth}
  \centering
  \includegraphics[width=1.0\linewidth]{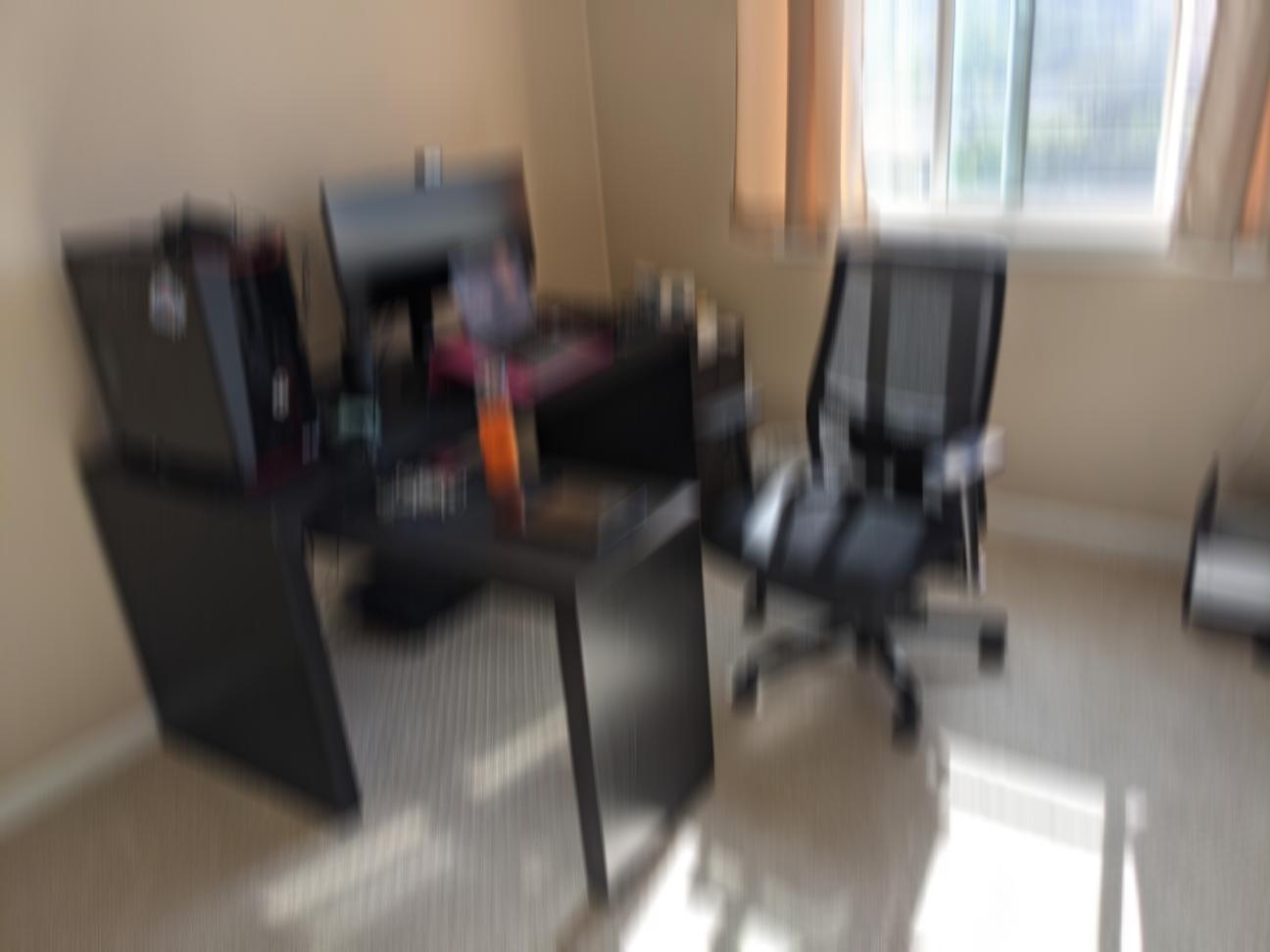}
\end{subfigure}%
\setcounter{subfigure}{0}

\rotatebox{90}{\quad\quad\,\, \tiny{Haze}}
\begin{subfigure}[t]{0.24\textwidth}
  \centering
  \includegraphics[width=1.0\linewidth]{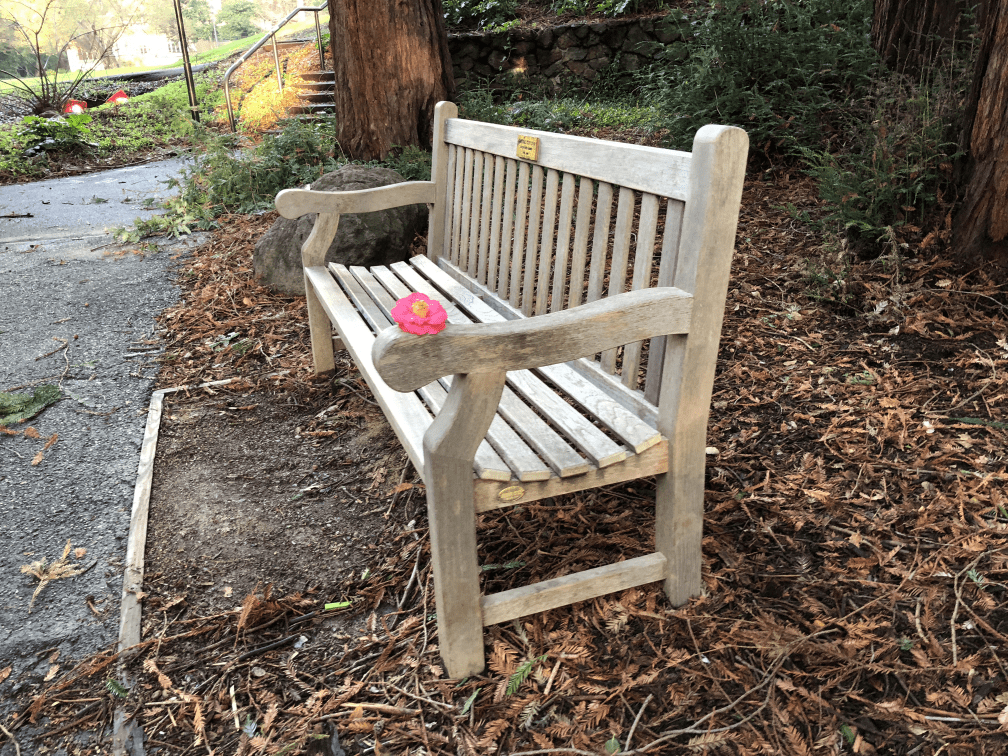}
\end{subfigure}%
\begin{subfigure}[t]{0.24\textwidth}
  \centering
  \includegraphics[width=1.0\linewidth]{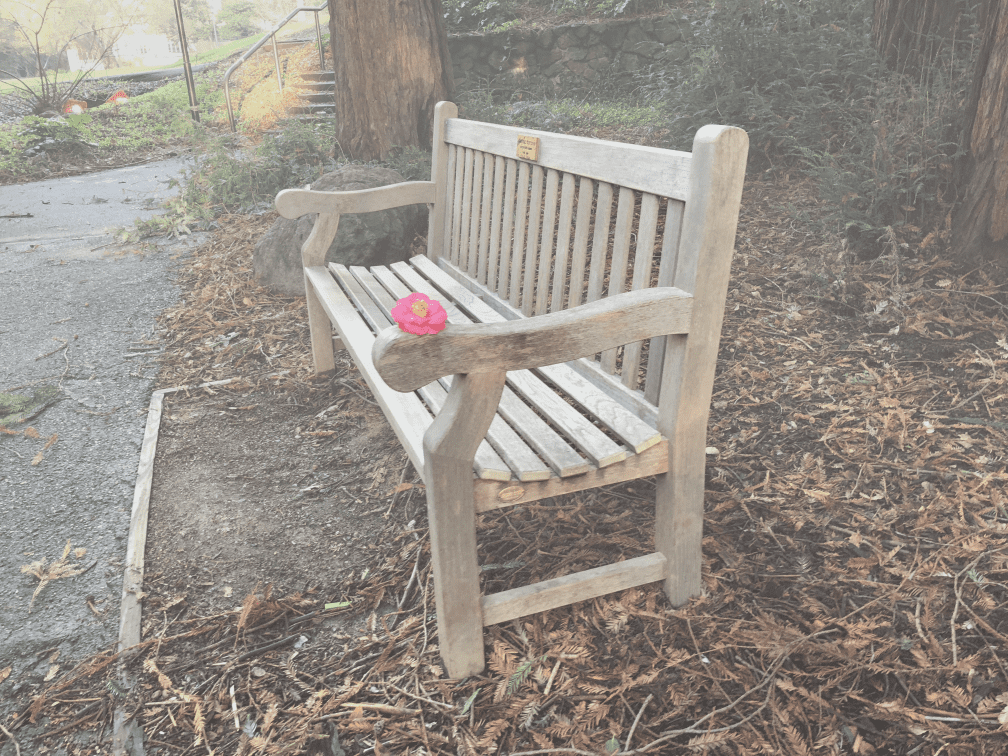}
\end{subfigure}%
\begin{subfigure}[t]{0.24\textwidth}
  \centering
  \includegraphics[width=1.0\linewidth]{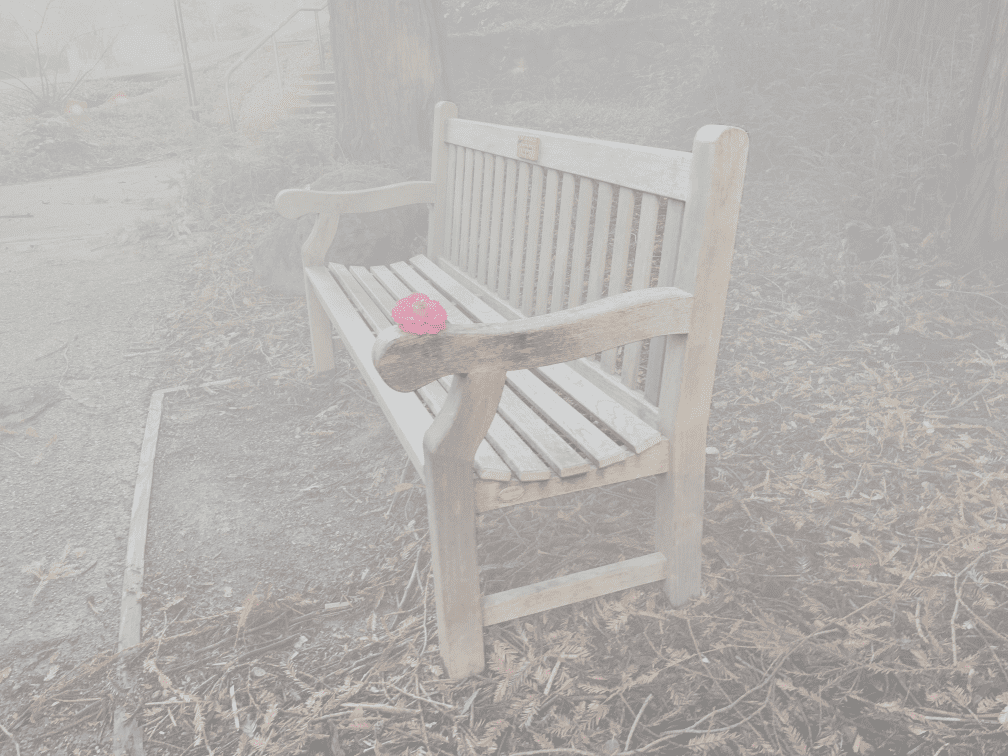}
\end{subfigure}%
\begin{subfigure}[t]{0.24\textwidth}
  \centering
  \includegraphics[width=1.0\linewidth]{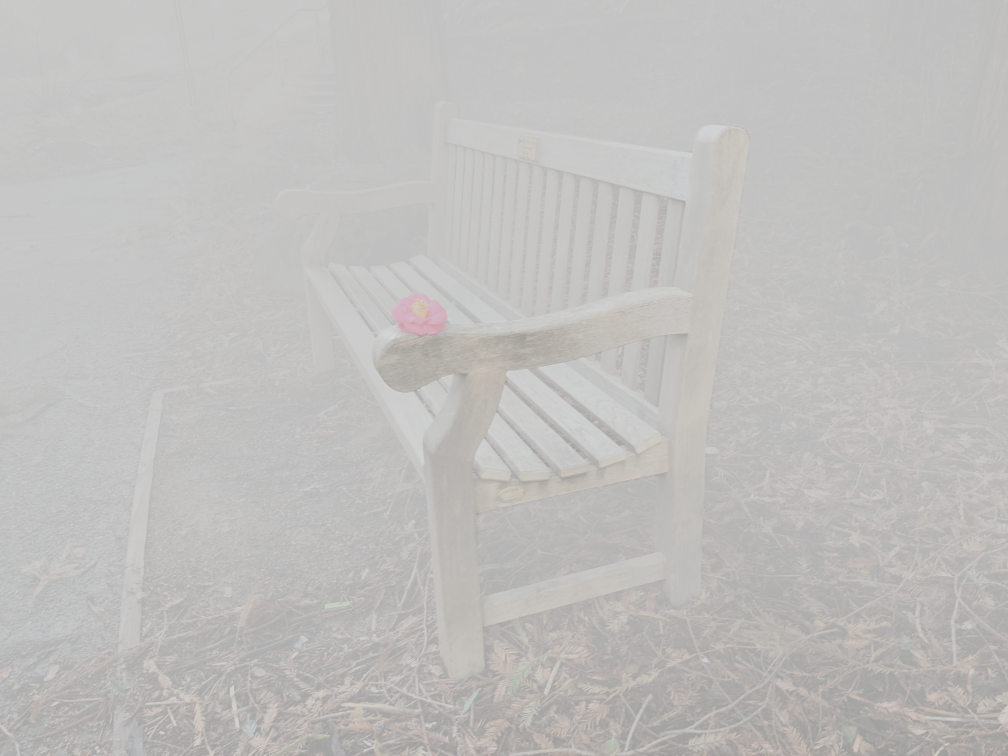}
\end{subfigure}%
\setcounter{subfigure}{0}

\rotatebox{90}{\quad\quad\,\, \tiny{Rain}}
\begin{subfigure}[t]{0.24\textwidth}
  \centering
  \includegraphics[width=1.0\linewidth]{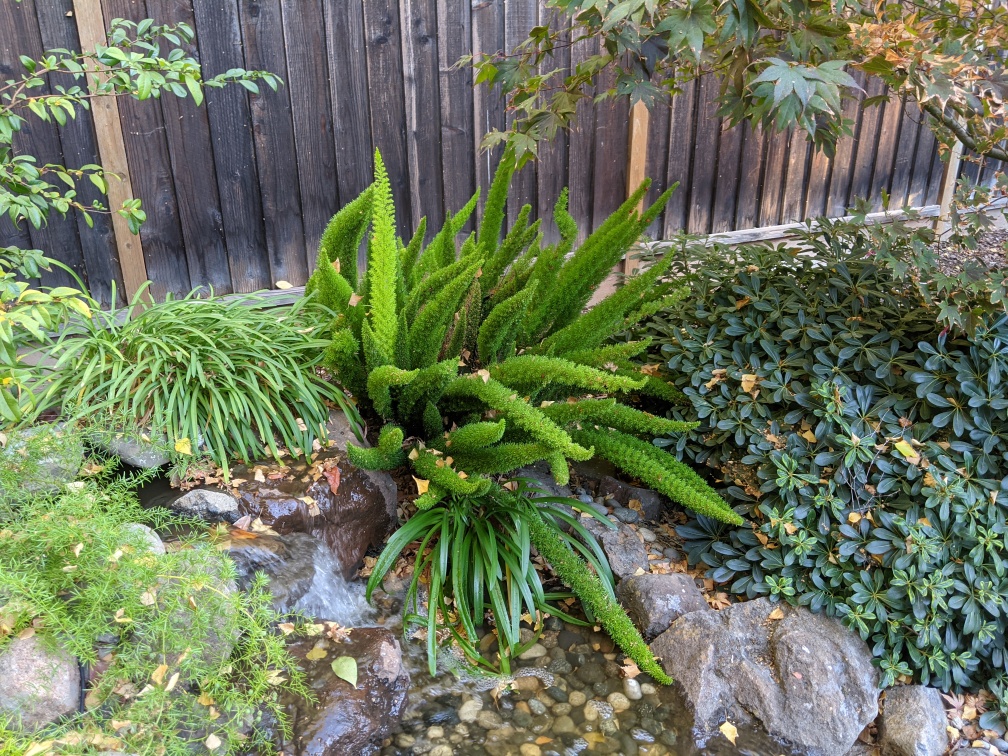}
\end{subfigure}%
\begin{subfigure}[t]{0.24\textwidth}
  \centering
  \includegraphics[width=1.0\linewidth]{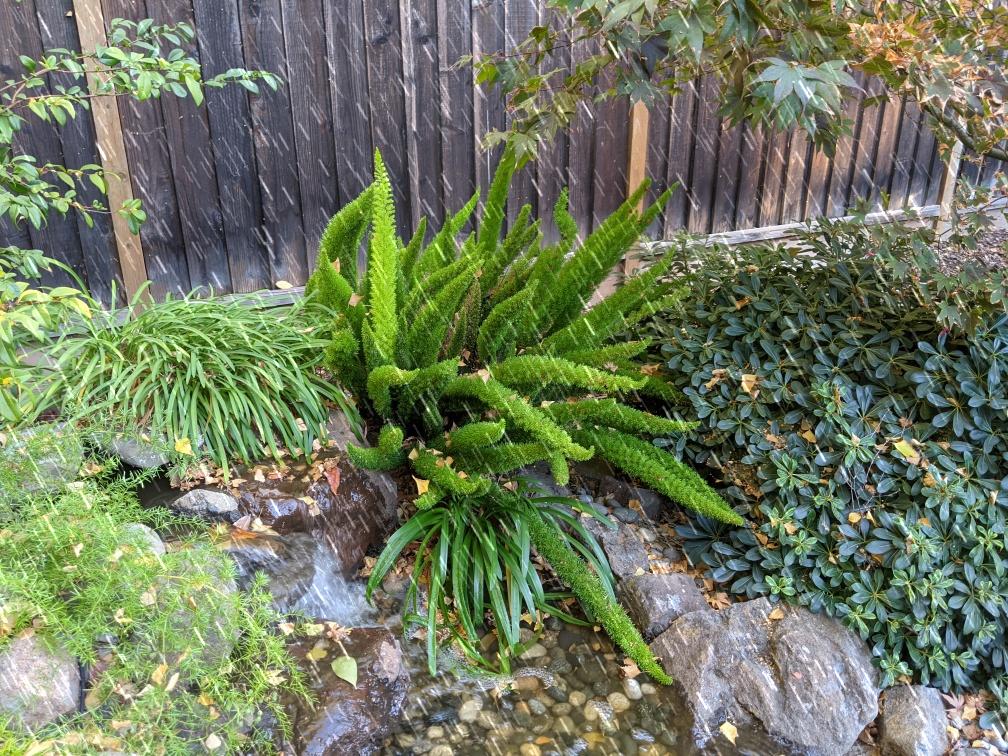}
\end{subfigure}%
\begin{subfigure}[t]{0.24\textwidth}
  \centering
  \includegraphics[width=1.0\linewidth]{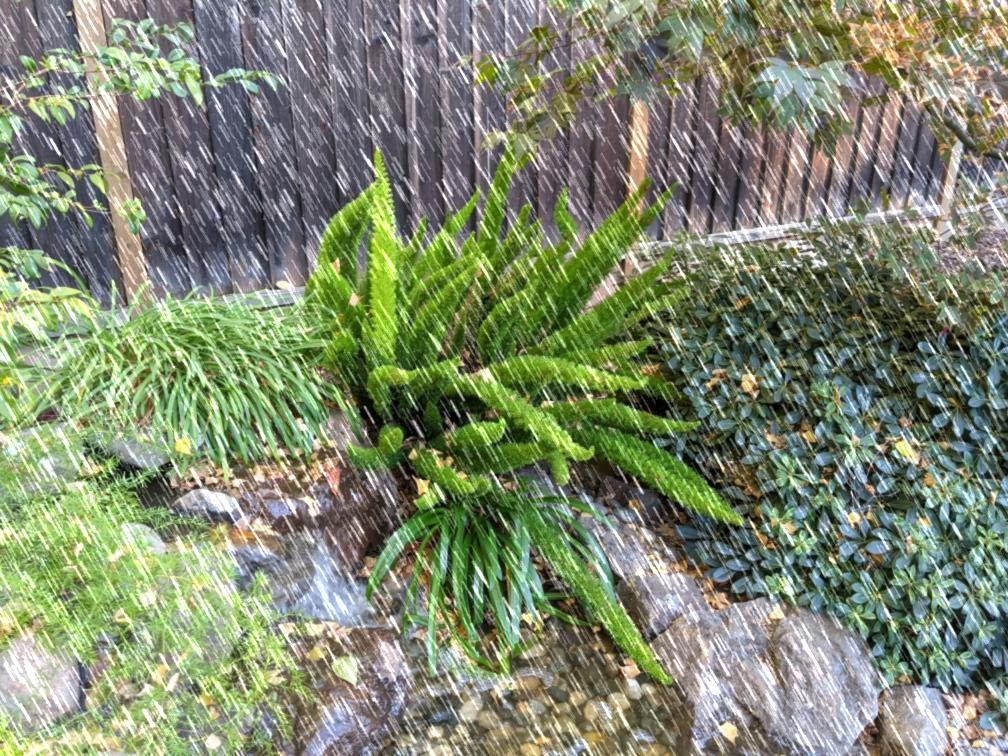}
\end{subfigure}%
\begin{subfigure}[t]{0.24\textwidth}
  \centering
  \includegraphics[width=1.0\linewidth]{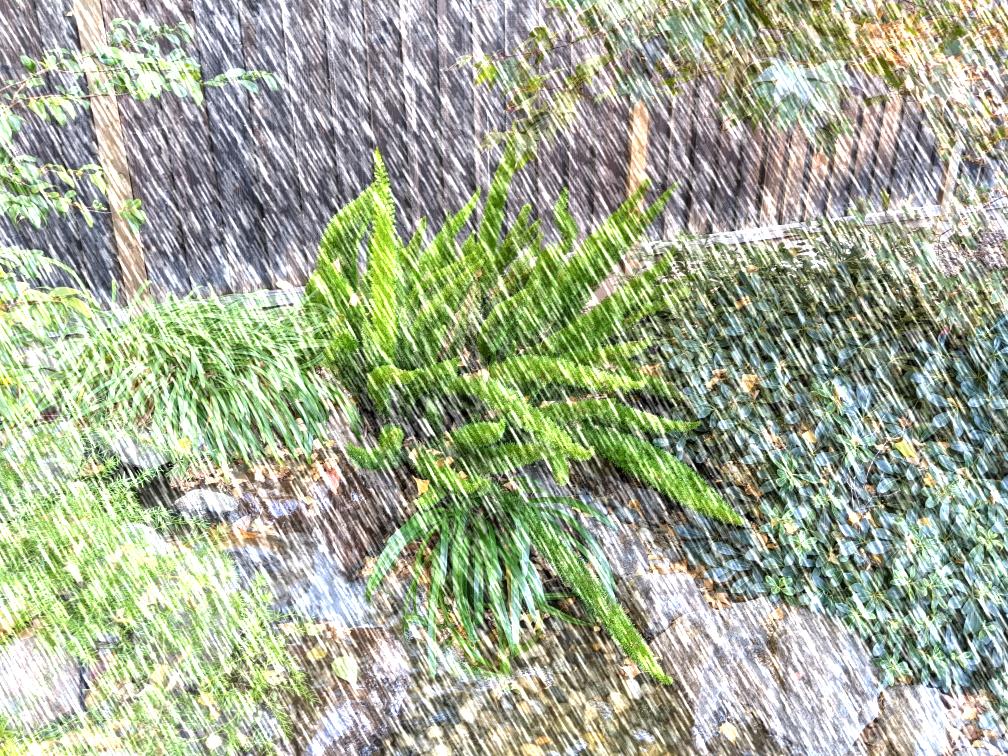}
\end{subfigure}%
\setcounter{subfigure}{0}

\rotatebox{90}{\quad\quad\,\, \tiny{Snow}}
\begin{subfigure}[t]{0.24\textwidth}
  \centering
  \includegraphics[width=1.0\linewidth]{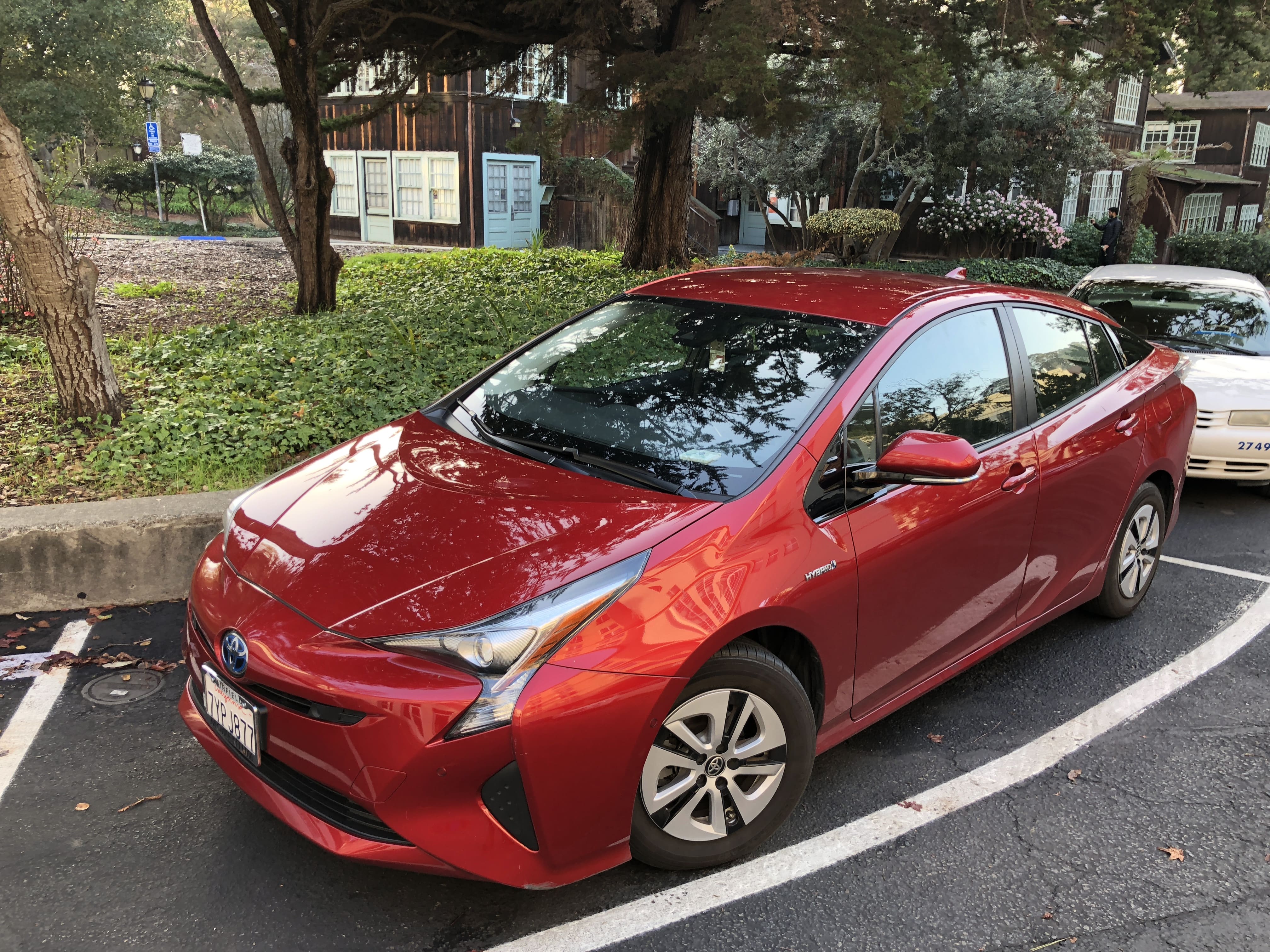}
\end{subfigure}%
\begin{subfigure}[t]{0.24\textwidth}
  \centering
  \includegraphics[width=1.0\linewidth]{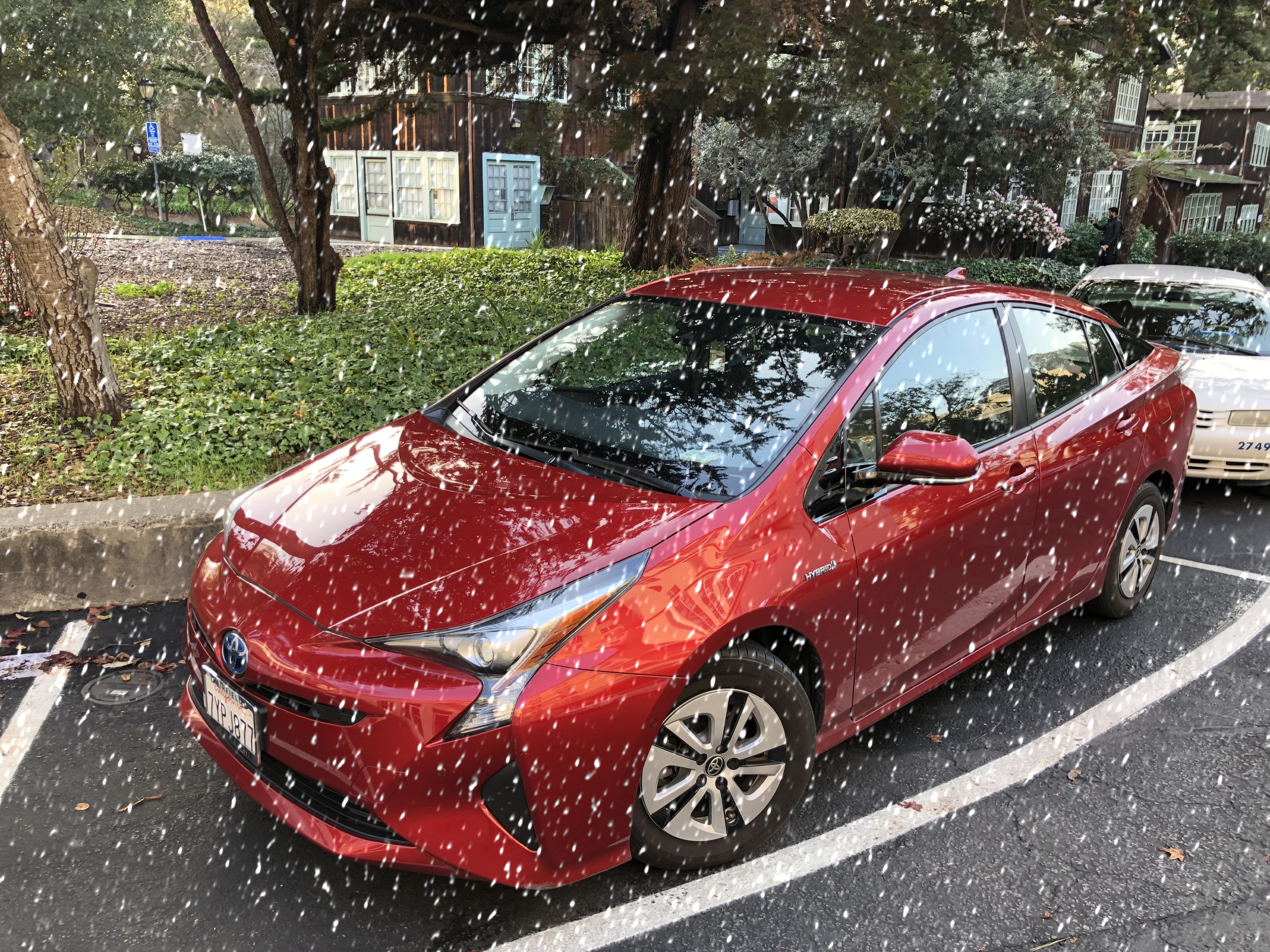}
\end{subfigure}%
\begin{subfigure}[t]{0.24\textwidth}
  \centering
  \includegraphics[width=1.0\linewidth]{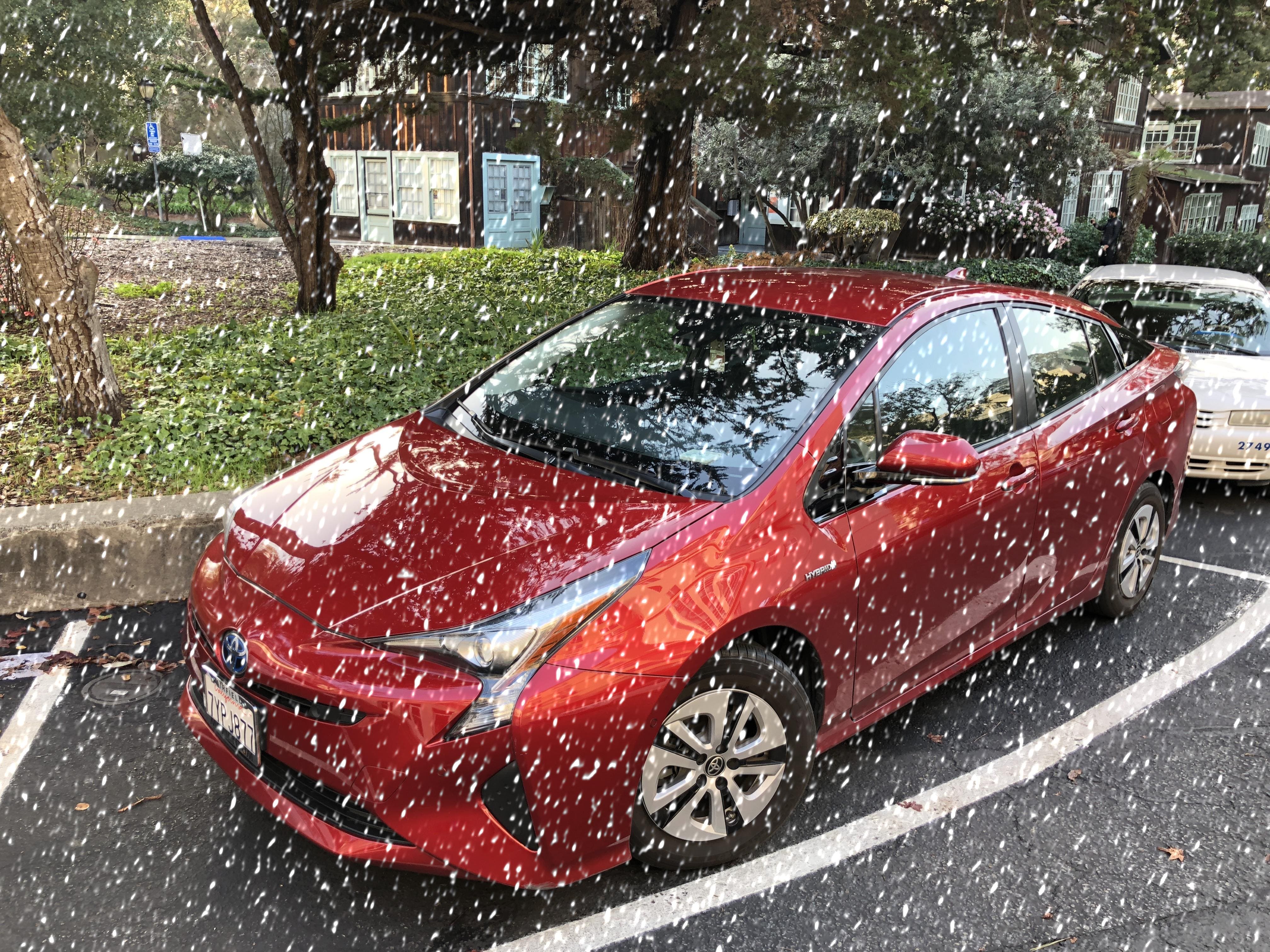}
\end{subfigure}%
\begin{subfigure}[t]{0.24\textwidth}
  \centering
  \includegraphics[width=1.0\linewidth]{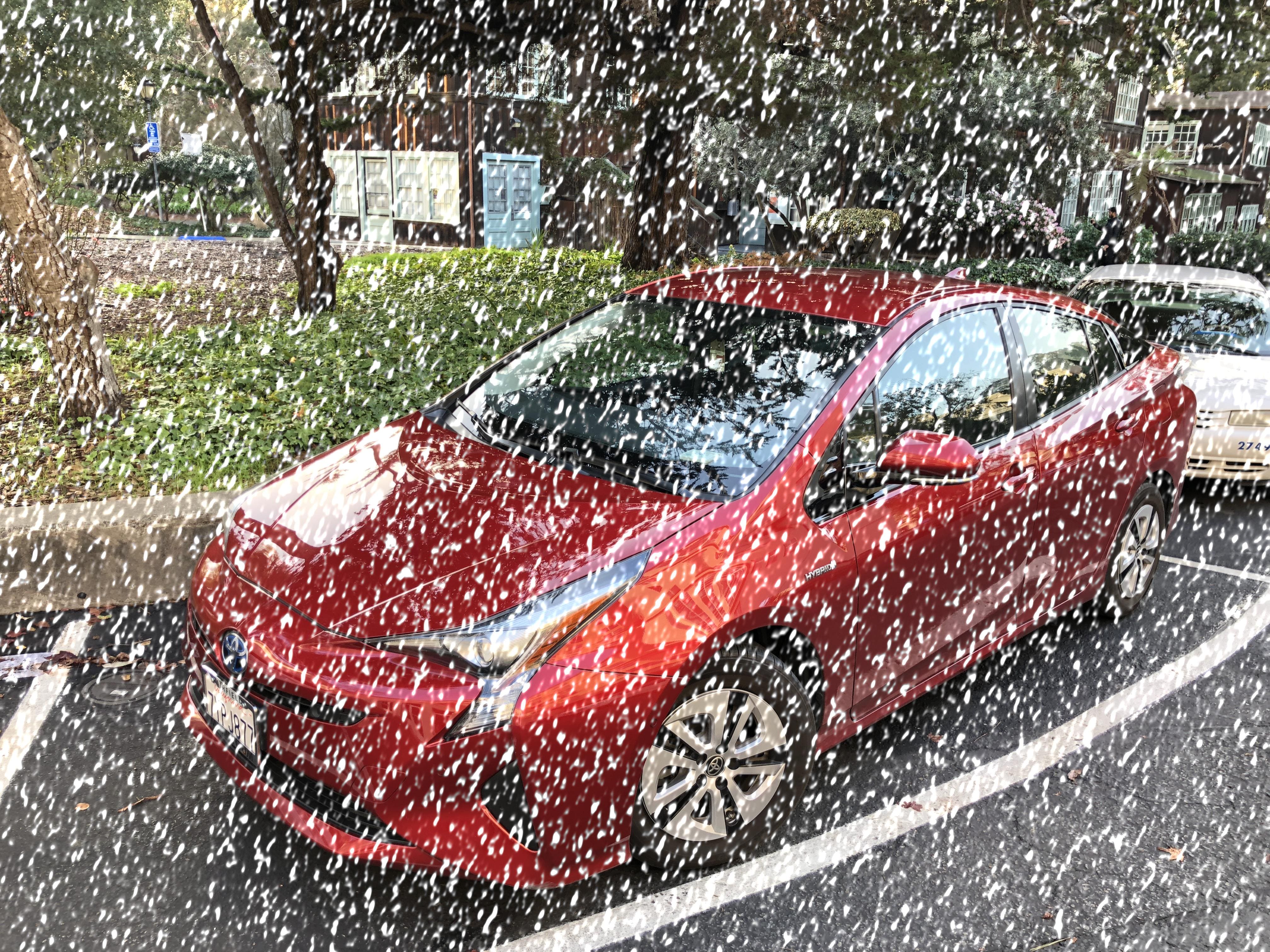}
\end{subfigure}%
\setcounter{subfigure}{0}

\rotatebox{90}{\quad\quad \tiny{Defocus}}
\begin{subfigure}[t]{0.24\textwidth}
  \centering
  \includegraphics[width=1.0\linewidth]{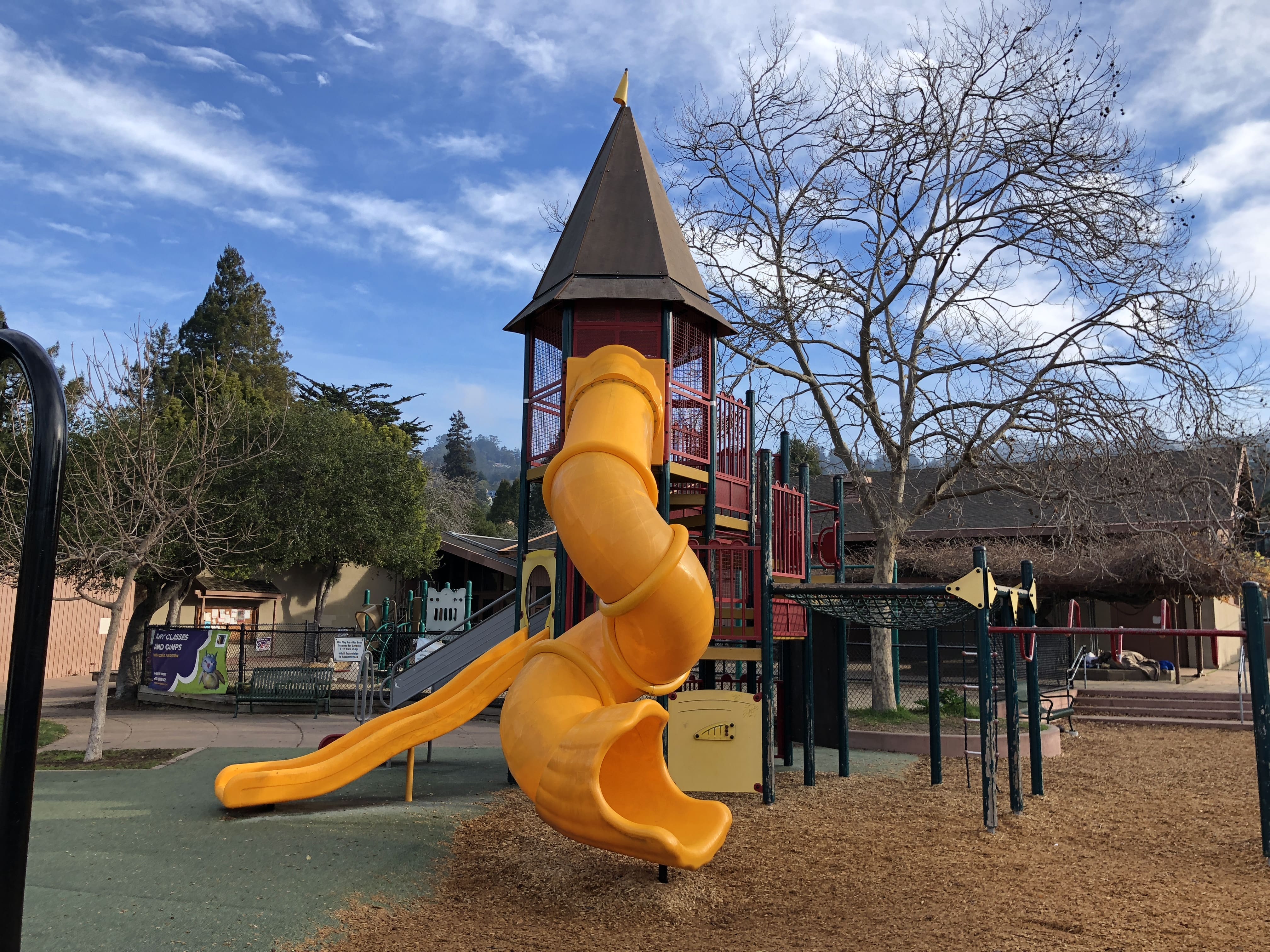}
\end{subfigure}%
\begin{subfigure}[t]{0.24\textwidth}
  \centering
  \includegraphics[width=1.0\linewidth]{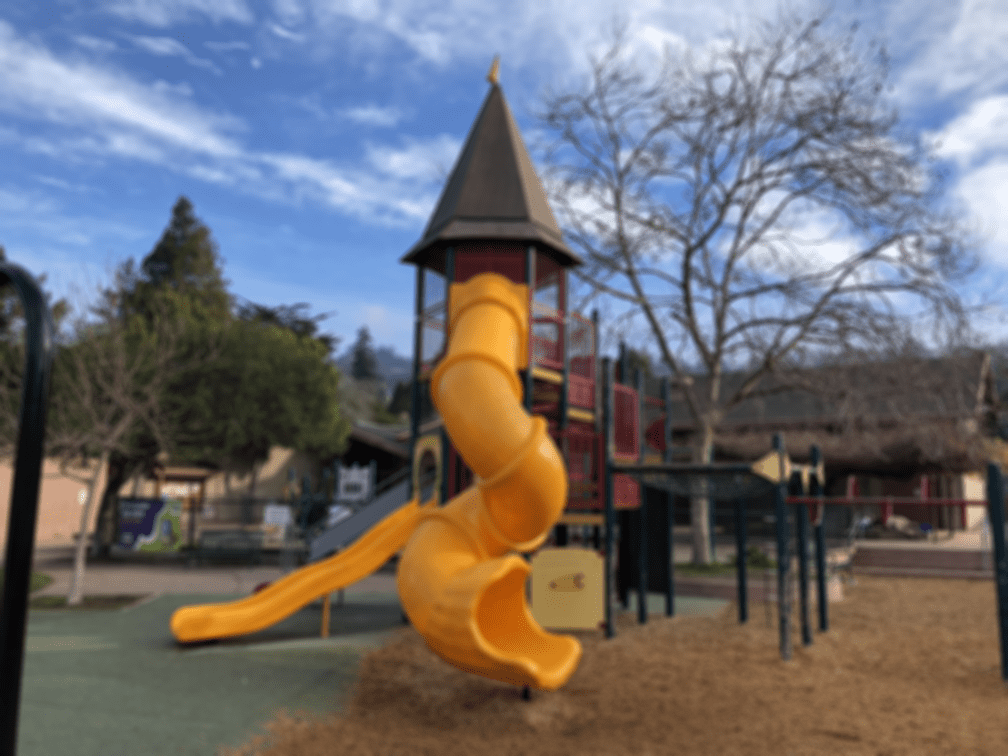}
\end{subfigure}%
\begin{subfigure}[t]{0.24\textwidth}
  \centering
  \includegraphics[width=1.0\linewidth]{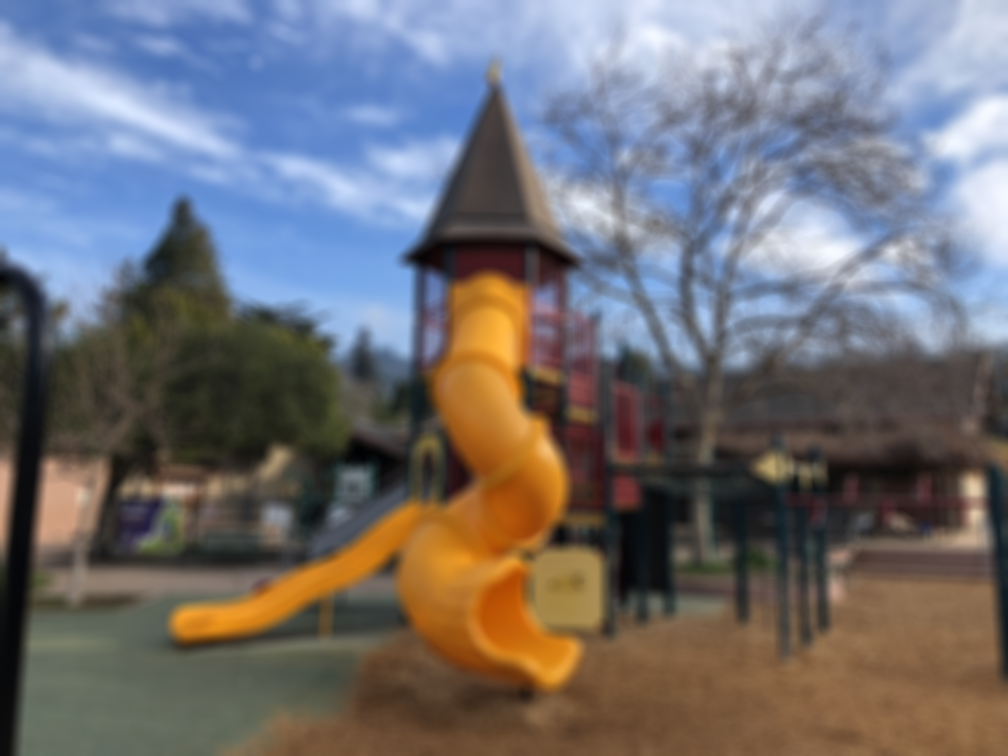}
\end{subfigure}%
\begin{subfigure}[t]{0.24\textwidth}
  \centering
  \includegraphics[width=1.0\linewidth]{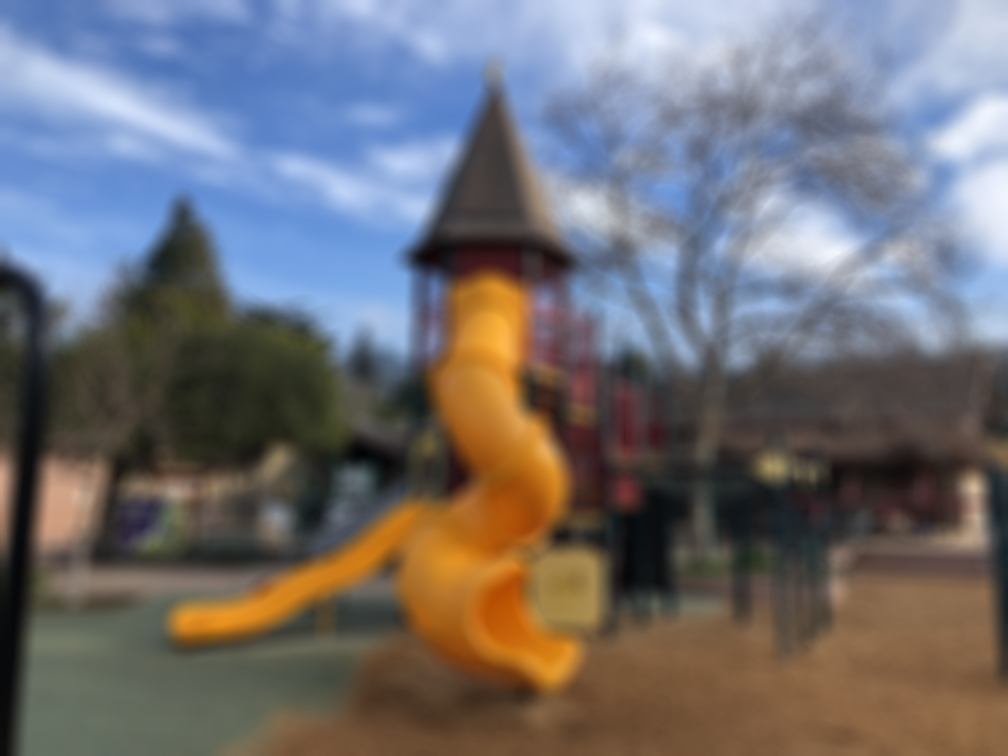}
\end{subfigure}%

\caption{We show visualisation of images degraded with varying intensities. The first column corresponds to the clean image, and starting from the second column, the images are degraded from low to high intensities. This approach enables us to capture a range of randomness in the degradation of the scene, facilitating easy generalization to real-world data. }
\label{fig:strength}
\vspace{-1em}
\end{figure*}

For simulating \textit{Rain}, we adopt a similar methodology and utilize OpenCV to generate rain particles. Three key parameters are varied during the rain generation process: the intensity of rain, the size of the rain streaks, and the direction of rain. In the case of \textit{Snow} generation, we utilize the imgaug library to introduce snowflakes onto the image. Unlike rain particles, we adjust a range of parameters including the direction of fall, density of the snowflakes, size of the snowflakes, and speed of the snowflakes.  
For all these corruptions, these parameters are carefully selected to encompass a wide range of variations, enhancing the transferability of synthetic data to real-world multi-view scenarios. We provide visualizations demonstrating the different intensities utilized in our training data across all corruption types in Figure \ref{fig:strength}.

\begin{table}[t]
  \centering
  \caption{
  Quantitative results on scenes containing \textit{real-world} degradations - specifically low-light enhancement, motion blur removal, and dehazing. All the baselines compared against are all-in-one baselines which generalise both to scene and corruption. The \sethlcolor{black!30}\hl{best} scores and \sethlcolor{black!15}\hl{second best} scores are highlighted.
  }
  \resizebox{1.0\columnwidth}{!}{
  \begin{tabular}{lccccc|ccc|ccc}
  
    \toprule
    \multirow{2}{*}{Models} & \multicolumn{2}{c}{Generalize to} & \multicolumn{3}{c|}{Low-Light} & \multicolumn{3}{c|}{Motion Blur} & \multicolumn{3}{c}{Haze}\\
    \cmidrule(r){2-12}
    & Scene & Corr. & \hspace{0.2em}PSNR$\uparrow$\hspace{0.2em} & SSIM$\uparrow$\hspace{0.2em} & LPIPS$\downarrow$\hspace{0.2em} & \hspace{0.2em}PSNR$\uparrow$\hspace{0.2em} & SSIM$\uparrow$\hspace{0.2em} & LPIPS$\downarrow$\hspace{0.2em} & \hspace{0.2em}PSNR$\uparrow$\hspace{0.2em} & SSIM$\uparrow$\hspace{0.2em} & LPIPS$\downarrow$\\
    \midrule
    GNT--(Airnet) & \xmark & \xmark & 17.73 & 0.577 & 0.367 & 20.73 & 0.612 & 0.406 & 15.70 & 0.590 & 0.351\\ 
    GNT--(PromptIR) & \xmark & \xmark & \cellcolor{black!15}17.90 & 0.573 & \cellcolor{black!15}0.354 & 19.33 & 0.552 & \cellcolor{black!15}0.398 & 16.10 & 0.632 & \cellcolor{black!15}0.293\\ 
    % \midrule
    GNT--(DA-CLIP) & \cmark & \xmark & 14.28 & \cellcolor{black!15}0.615 & 0.424 & \cellcolor{black!15}20.88 & \cellcolor{black!15}0.632 & 0.410 & \cellcolor{black!15}16.68 & \cellcolor{black!15}0.729 & 0.300\\ 
    GNT--(AutoDIR) & \cmark & \cmark & 11.64 & 0.570 & 0.436 & 20.20 & 0.601 & 0.404 & 14.29 & 0.678 & 0.316\\ 
    % \midrule
    Ours & \cmark & \cmark & \cellcolor{black!30}19.91 & \cellcolor{black!30}0.736 & \cellcolor{black!30}0.352 & \cellcolor{black!30}22.12 & \cellcolor{black!30}0.712 & \cellcolor{black!30}0.346 & \cellcolor{black!30}16.82 & \cellcolor{black!30}0.759 & \cellcolor{black!30}0.288\\ 
    \bottomrule
  \end{tabular}
  }
  
  \label{tab:realscenes}
  \vspace{-1em}
\end{table}

\begin{table}
\centering
\caption{\footnotesize Ablation}
\label{tab:addn_abl}
\vspace{-0.2em}
\resizebox{0.5\columnwidth}{!}{\begin{tabular}{lc}
\toprule
Method & PSNR / SSIM / LPIPS\\
\midrule
Simple Conditioning & 21.34 / 0.689 / 0.394\\
ARM+DLM (Ours) & \cellcolor{black!30}22.12 / 0.712 / 0.346\\
\bottomrule
\end{tabular}}
\label{tab:ablviews}
% \vspace{-2em}
\end{table}

\begin{table}
  \centering
\caption{\footnotesize Effect of the number of input views for low-light enhancement. Metrics are ordered as PSNR / SSIM / LPIPS}
\vspace{-0.2em}
  \resizebox{\columnwidth}{!}{
  \begin{tabular}{lccc}
    \toprule
    3D Restore & \multicolumn{3}{c}{GAURA}\\
    \midrule
    N.A. & 3 views & 6 views & 10 views\\
    \midrule
     17.64 / 0.736 / 0.415 & 19.18 / 0.697 / 0.387 & 19.48 / 0.720 / 0.364 & 19.91 / 0.738 / 0.352\\
    \bottomrule
  \end{tabular}}
  \label{tab:effectofviews}
  % \vspace{-2em}
\end{table}

\begin{table}
  \centering
\caption{\footnotesize Quantitative results to measure view consistency. Metrics are ordered as RMSE / LPIPS}
\vspace{-0.2em}
  \resizebox{\columnwidth}{!}{
  \begin{tabular}{lcccccc}
    \toprule
    Method & \multicolumn{2}{c}{Short Range Consistency$\downarrow$} & \multicolumn{2}{c}{Long Range Consistency$\downarrow$}\\
    & Rain & Snow & Rain & Snow \\
    \midrule
    GNT--(All-in-one) Restore & 0.138 / 0.253 & 0.099 / 0.211 & 0.249 / 0.402 & 0.188 / 0.312\\
     GAURA & \cellcolor{black!30}0.115 / 0.215 & \cellcolor{black!30}0.083 / 0.189 & \cellcolor{black!30}0.204 / 0.333 & \cellcolor{black!30}0.153 / 0.299\\
    \bottomrule
  \end{tabular}}
  \label{tab:viewconsis}
  \vspace{-2em}
\end{table}

\section{Results}

\begin{figure*}[!ht]
\centering
\includegraphics[width=\textwidth]{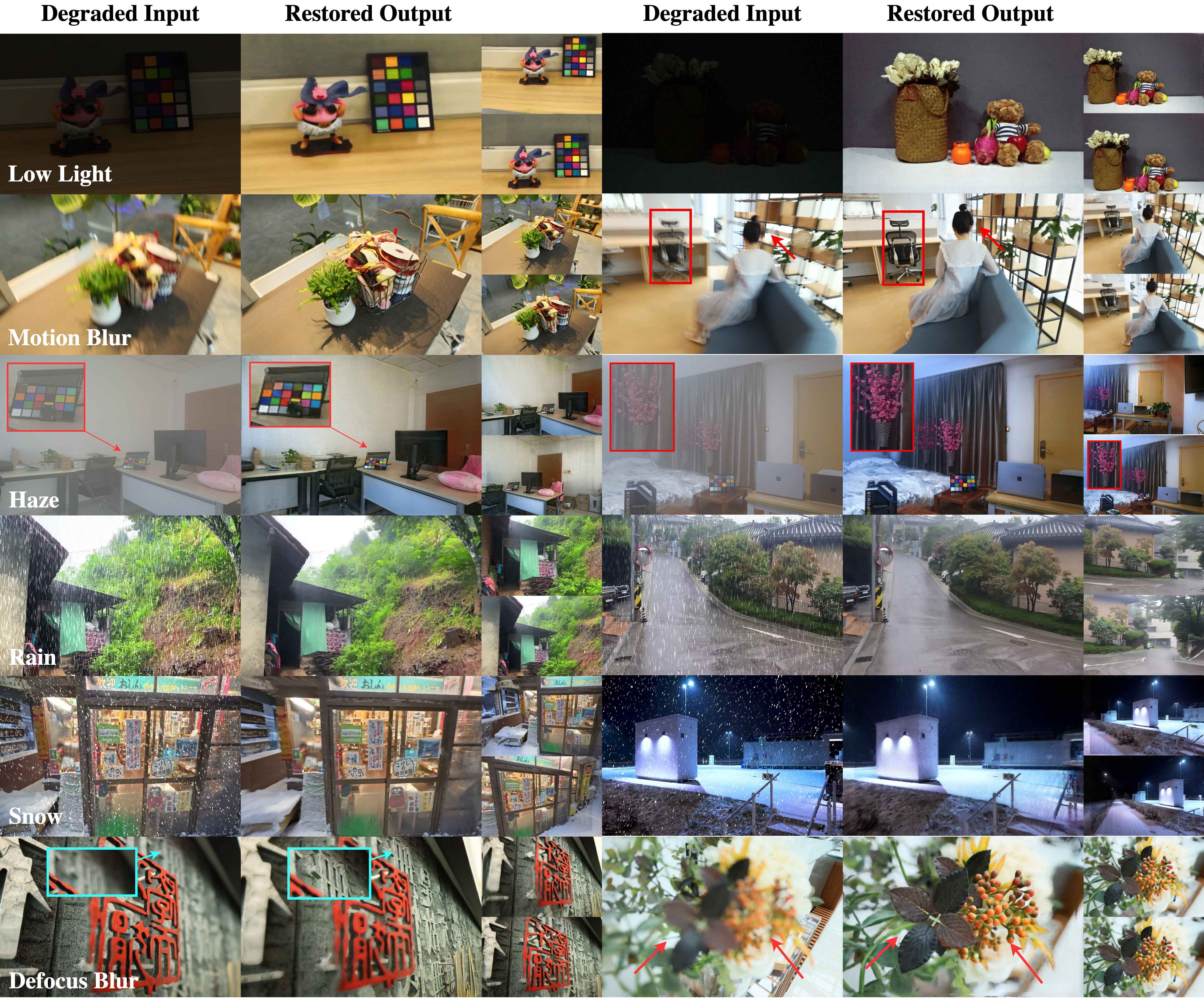}
\caption{Below are the gallery results showcasing our method's performance on various corruptions on real-world data. Each row presents results of our method on a single corruption across two scenes. The degraded image and its corresponding restored output are visualized side-by-side. Additionally, renders from two other novel viewpoints are provided on the right. }
\label{fig:gallery}
\vspace{3em}
\includegraphics[width=\textwidth]{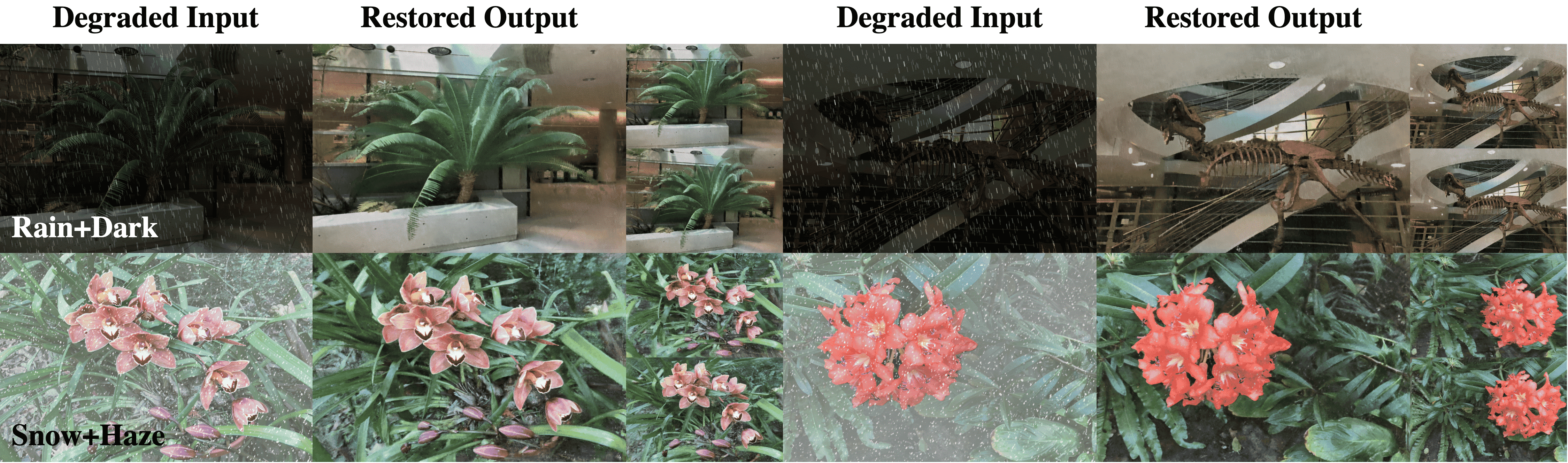}
\caption{Results on scenes corrupted with more than 1 degradations. We show results on scenes corrupted with Rain+Dark and Snow+Haze on the synthetically corrupted LLFF dataset. }
\label{fig:multi-corr}
\end{figure*}

\paragraph{\textbf{Real-World Data.}} We provide a summary of the quantitative results against the GNT-all-in-one baselines in Table \ref{tab:realscenes}. In the main paper, we select the best-performing all-in-one model and present both qualitative and quantitative results accordingly. In this section, we delve into further details regarding the selected best method. For the \textit{low-light} enhancement task, we utilize AirNet+GNT as the baseline. In the cases of \textit{motion deblurring} and \textit{dehazing} tasks, DA-CLIP+GNT serves as the baseline for comparison. Similarly, for rain and snow removal, we employ DA-CLIP+GNT as the comparative baseline for both qualitative and quantitative analyses. Regarding defocus, we utilize the state-of-the-art Single Pixel Defocus Deblur model \cite{cui2023selective}. We present a collection of results showcasing our method's performance on real-world data across several types of corruptions in Fig. \ref{fig:gallery}

\begin{figure*}[t]
\centering
\captionsetup[subfigure]{labelformat=empty}
% \rotatebox[origin=lc]{90}{\scriptsize{Low-Light}}
\rotatebox{90}{\quad \tiny{Low-Light}}
\begin{subfigure}[t]{0.19\textwidth}
  \centering
  \includegraphics[width=1.0\linewidth]{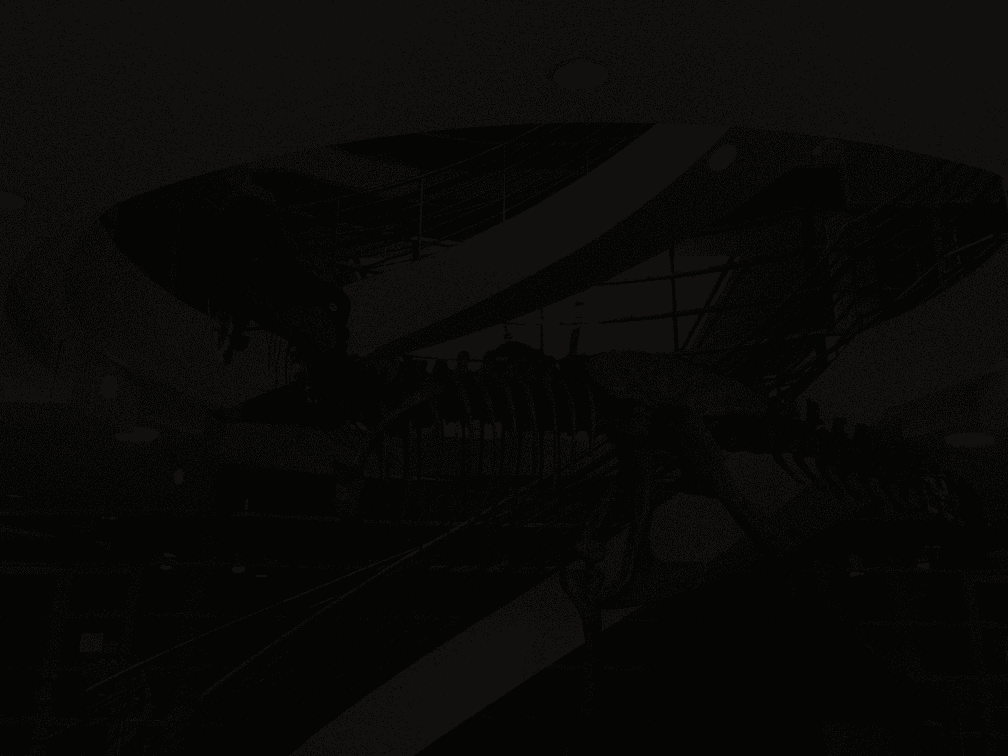}
\end{subfigure}%
\begin{subfigure}[t]{0.19\textwidth}
  \centering
  \includegraphics[width=1.0\linewidth]{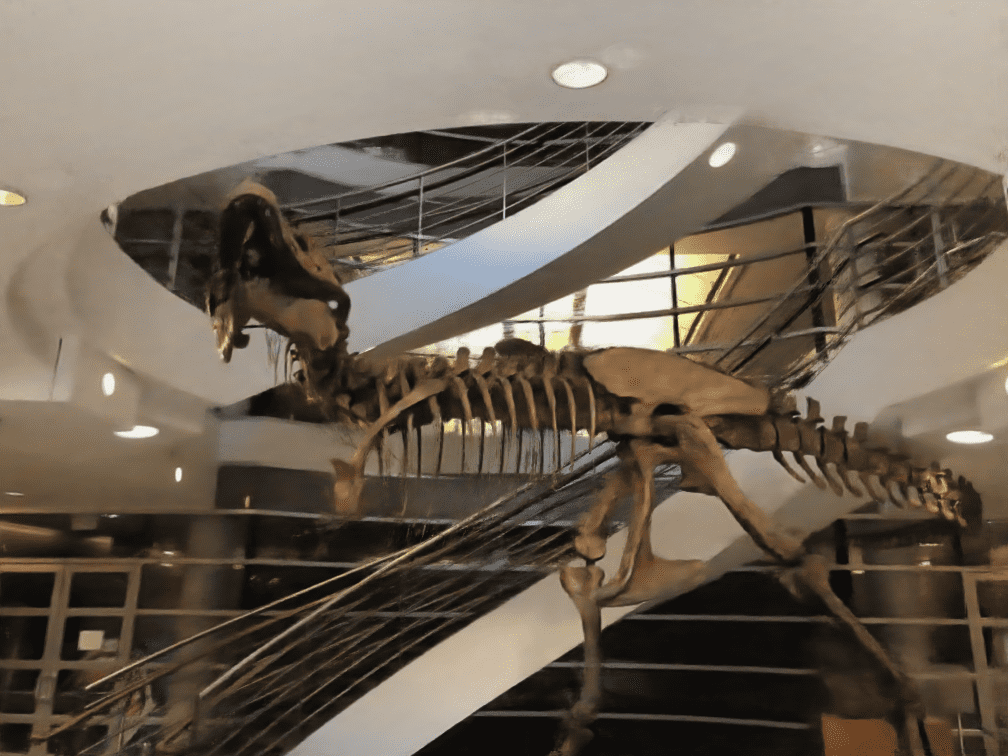}
\end{subfigure}%
\begin{subfigure}[t]{0.19\textwidth}
  \centering
  \includegraphics[width=1.0\linewidth]{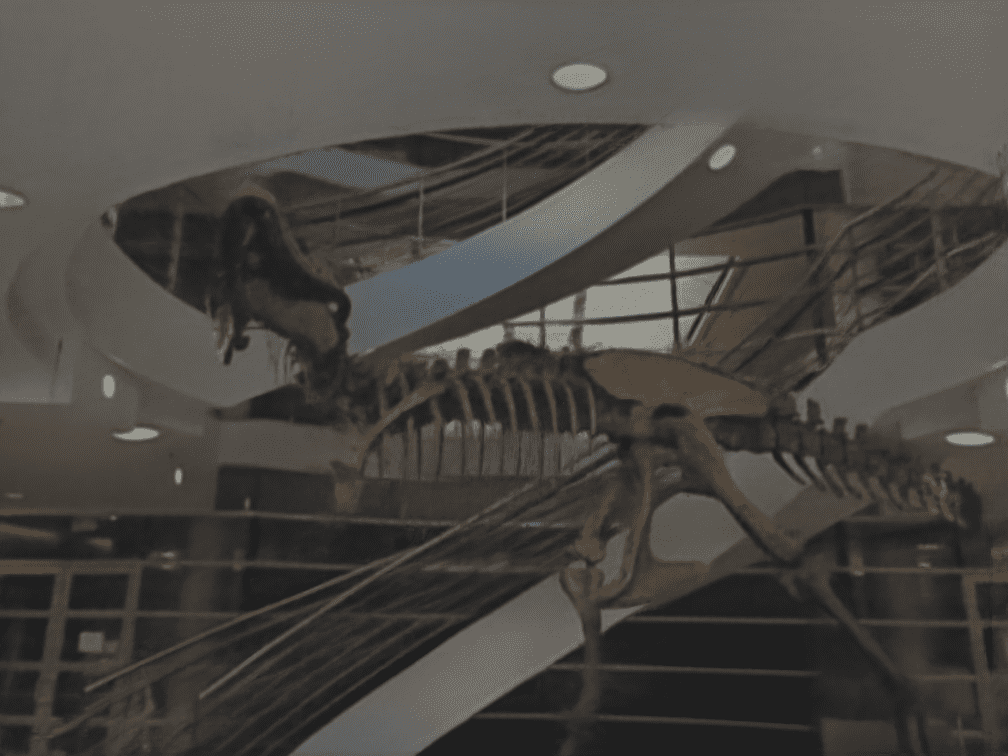}
\end{subfigure}%
\begin{subfigure}[t]{0.19\textwidth}
  \centering
  \includegraphics[width=1.0\linewidth]{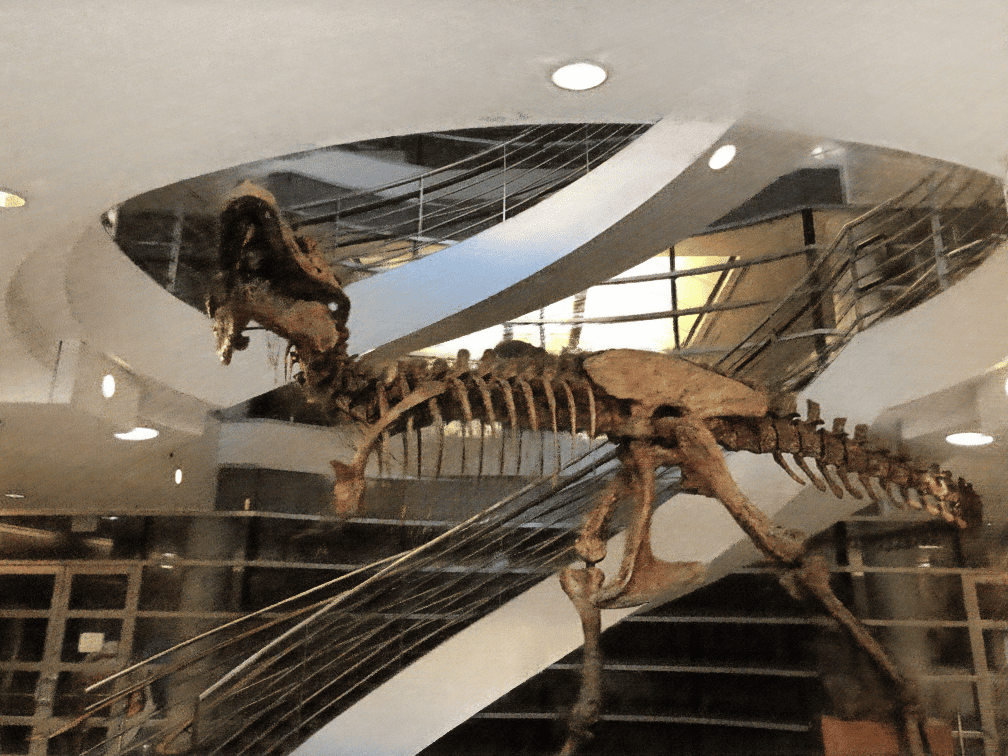}
\end{subfigure}%
\begin{subfigure}[t]{0.19\textwidth}
  \centering
  \includegraphics[width=1.0\linewidth]{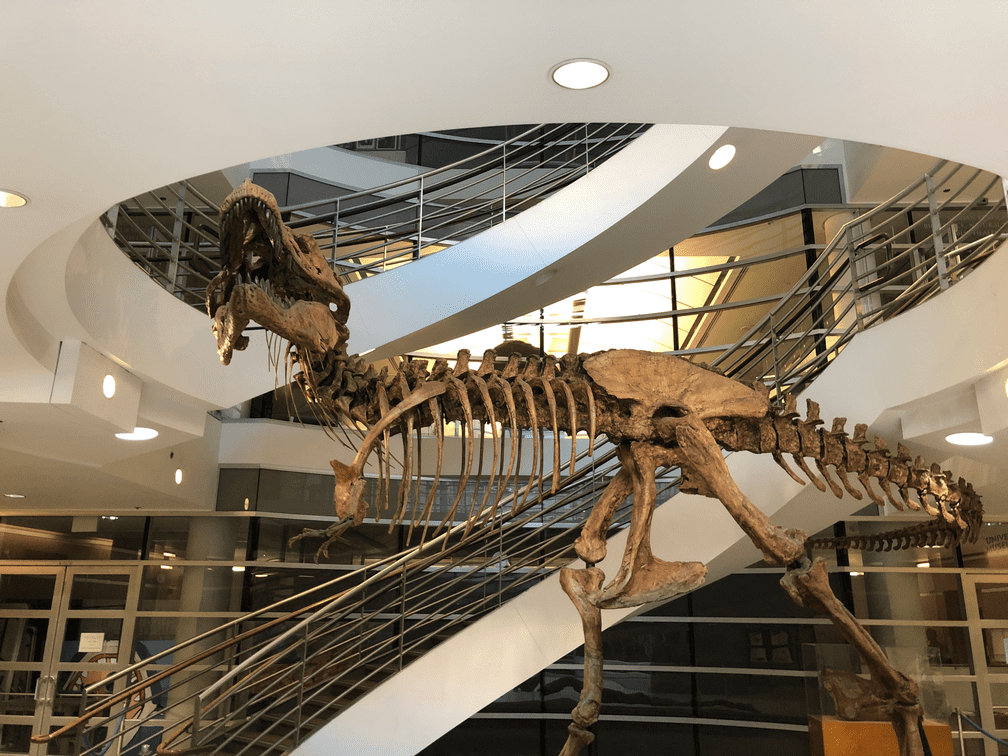}
\end{subfigure}%
\setcounter{subfigure}{0}

\rotatebox{90}{\quad \tiny{Motion Blur}}
\begin{subfigure}[t]{0.19\textwidth}
  \centering
  \includegraphics[width=1.0\linewidth]{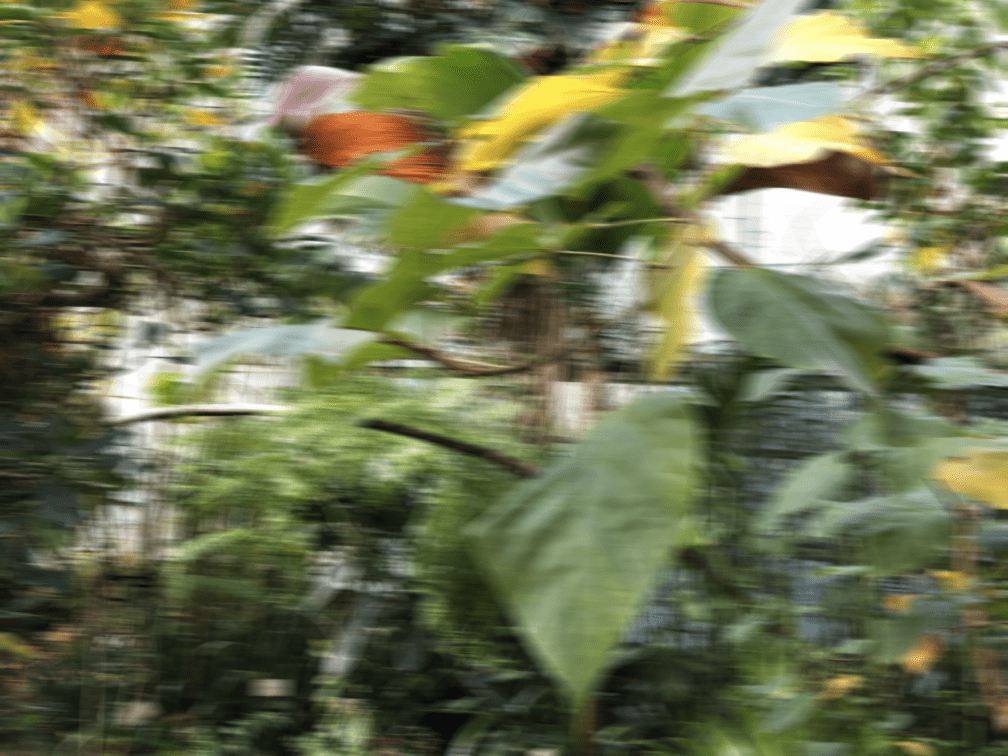}
\end{subfigure}%
\begin{subfigure}[t]{0.19\textwidth}
  \centering
  \includegraphics[width=1.0\linewidth]{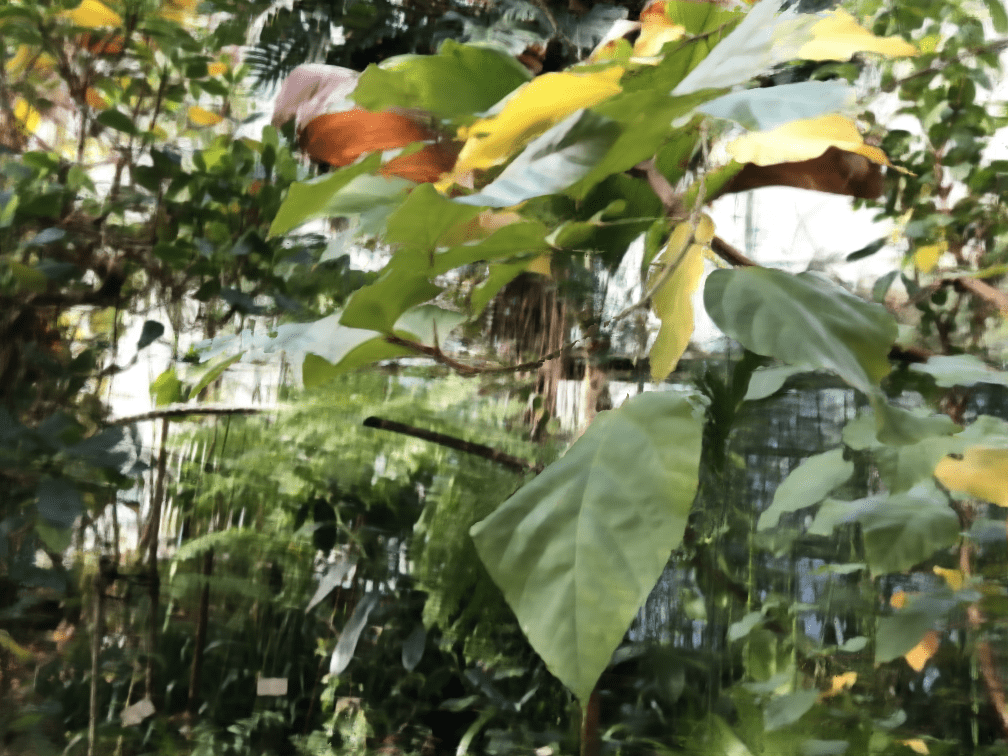}
\end{subfigure}%
\begin{subfigure}[t]{0.19\textwidth}
  \centering
  \includegraphics[width=1.0\linewidth]{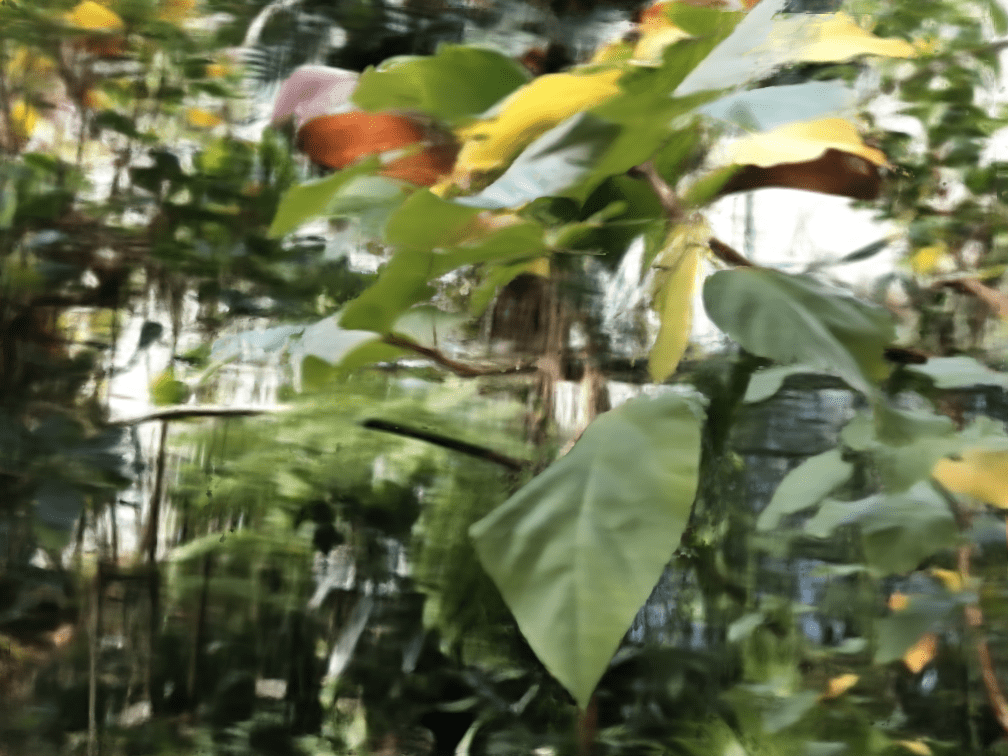}
\end{subfigure}%
\begin{subfigure}[t]{0.19\textwidth}
  \centering
  \includegraphics[width=1.0\linewidth]{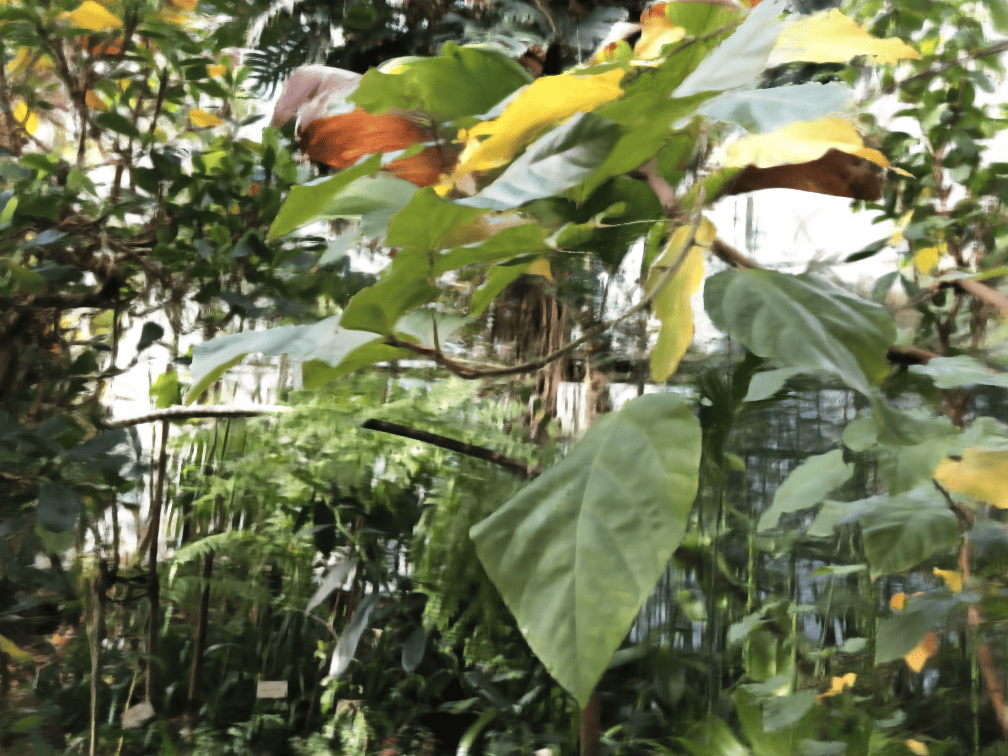}
\end{subfigure}%
\begin{subfigure}[t]{0.19\textwidth}
  \centering
  \includegraphics[width=1.0\linewidth]{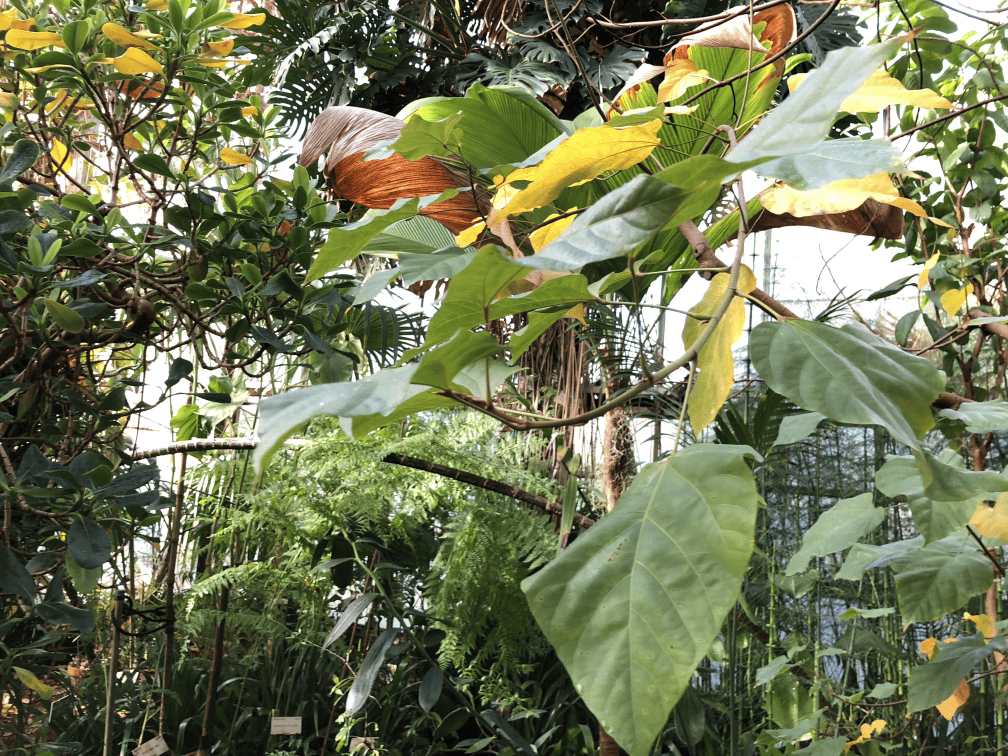}
\end{subfigure}%
\setcounter{subfigure}{0}

\rotatebox{90}{\quad\quad \tiny{Haze}}
\begin{subfigure}[t]{0.19\textwidth}
  \centering
  \includegraphics[width=1.0\linewidth]{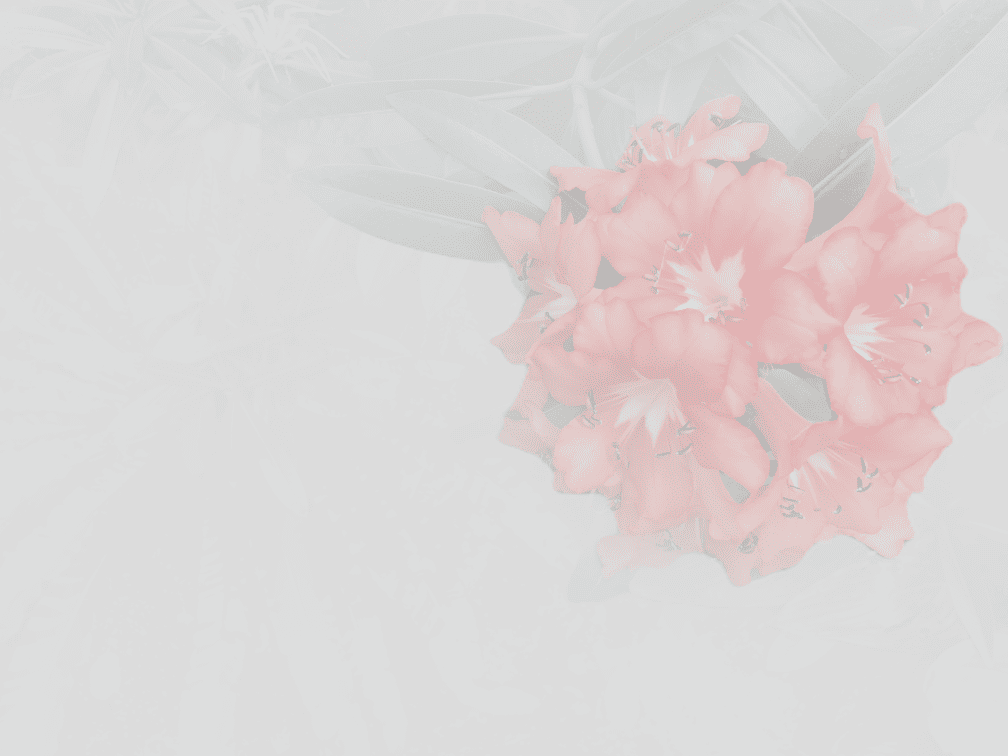}
\end{subfigure}%
\begin{subfigure}[t]{0.19\textwidth}
  \centering
  \includegraphics[width=1.0\linewidth]{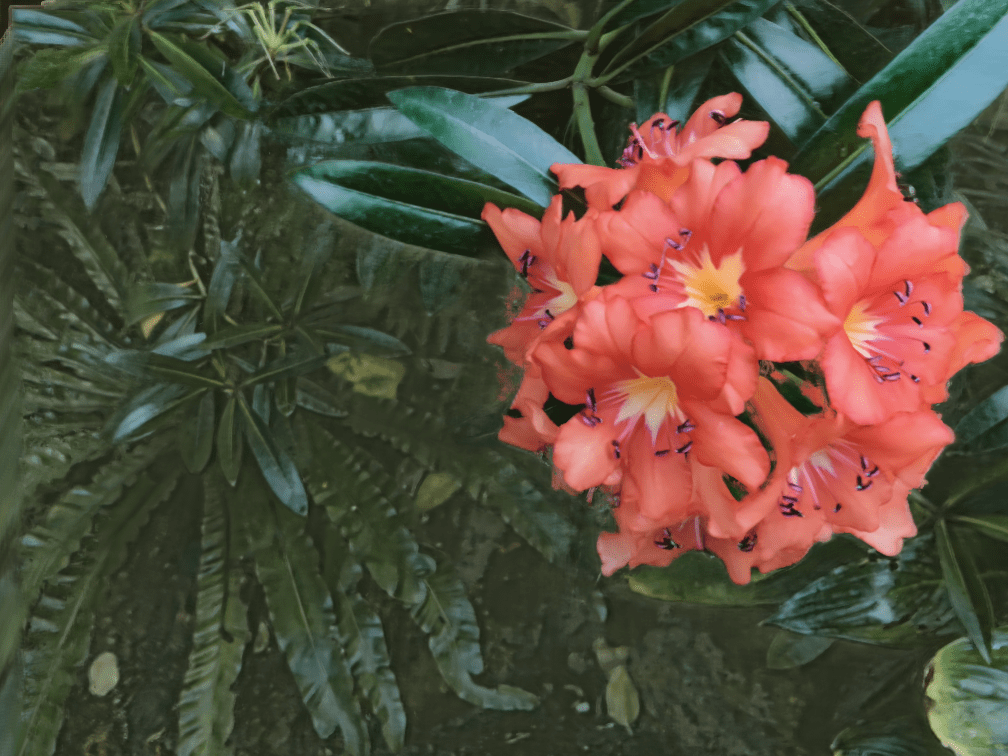}
\end{subfigure}%
\begin{subfigure}[t]{0.19\textwidth}
  \centering
  \includegraphics[width=1.0\linewidth]{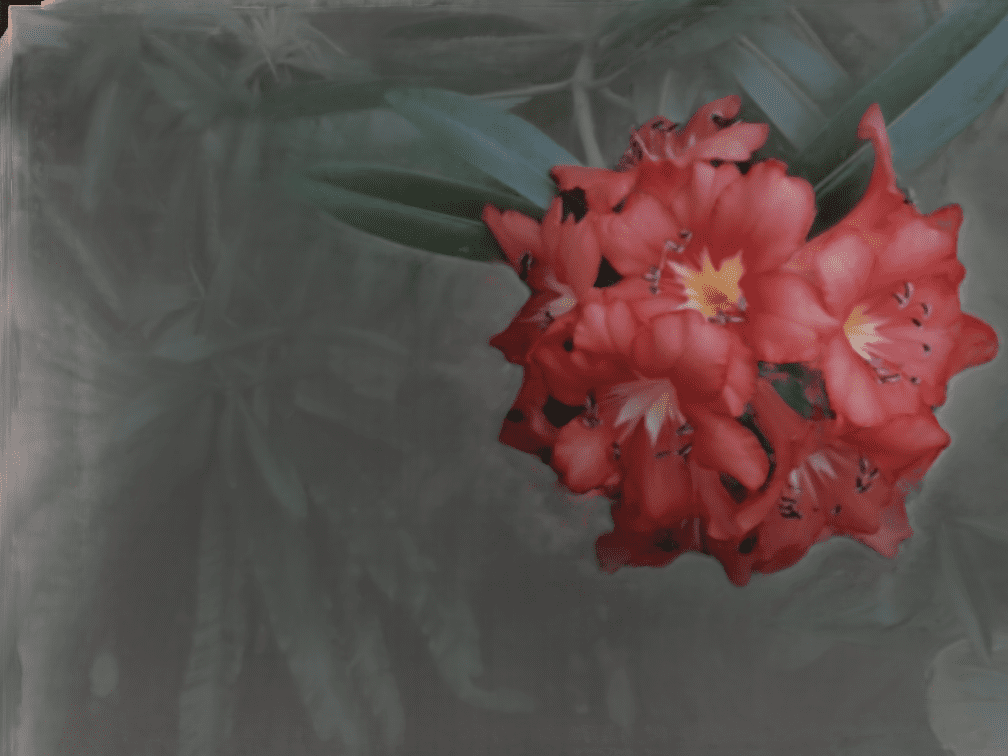}
\end{subfigure}%
\begin{subfigure}[t]{0.19\textwidth}
  \centering
  \includegraphics[width=1.0\linewidth]{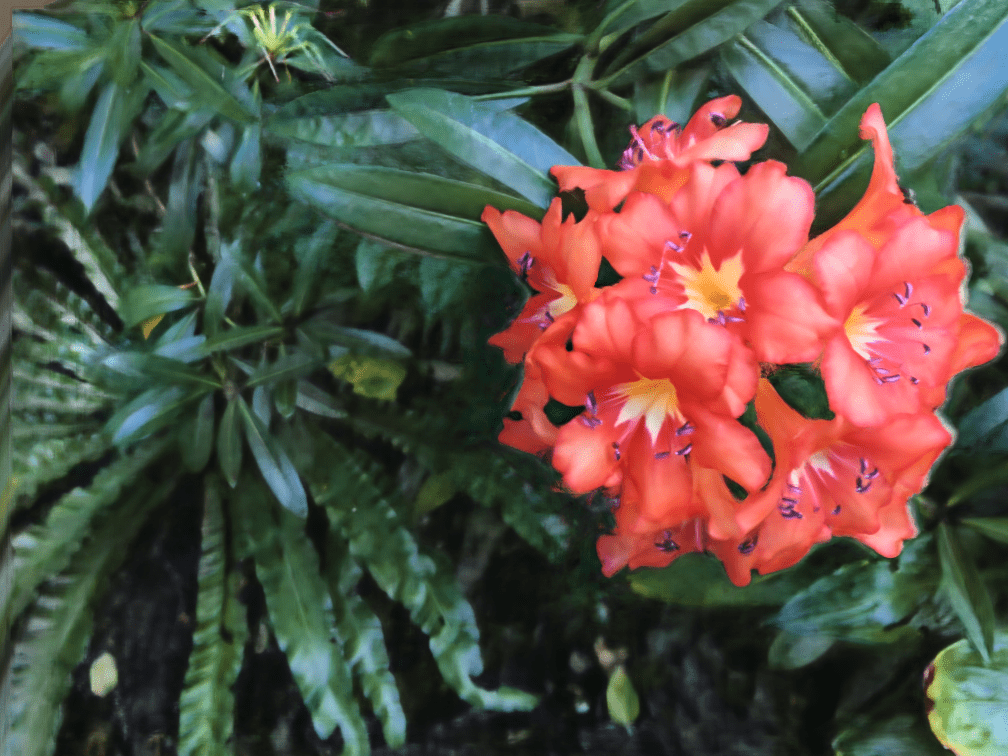}
\end{subfigure}%
\begin{subfigure}[t]{0.19\textwidth}
  \centering
  \includegraphics[width=1.0\linewidth]{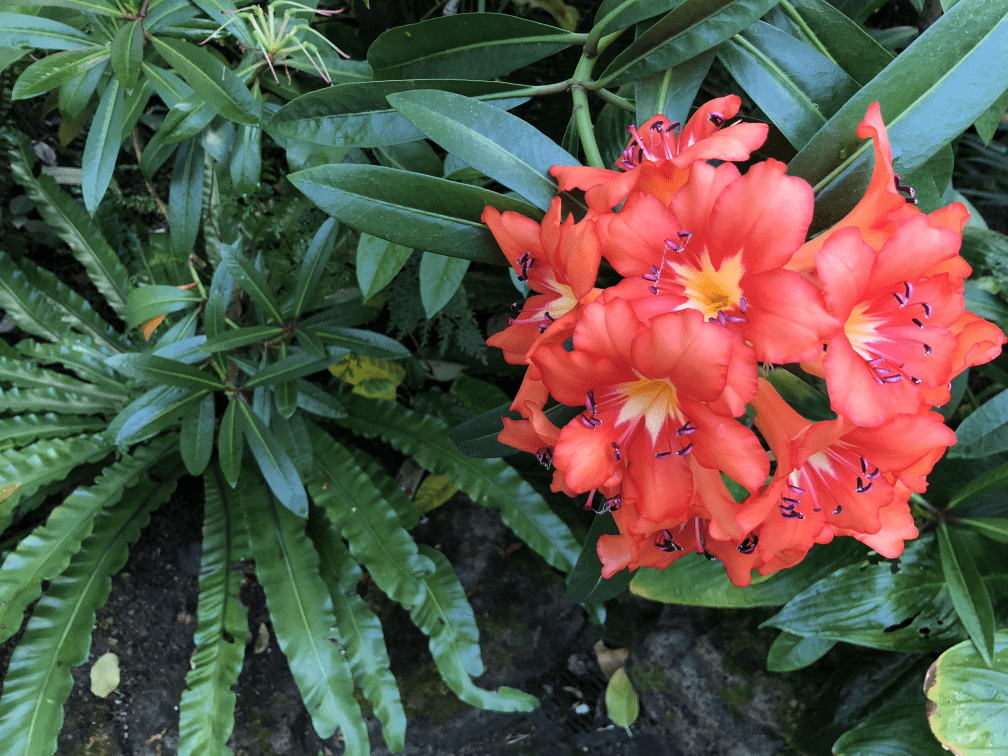}
\end{subfigure}%
\setcounter{subfigure}{0}

\rotatebox{90}{\quad\quad \tiny{Rain}}
\begin{subfigure}[t]{0.19\textwidth}
  \centering
  \includegraphics[width=1.0\linewidth]{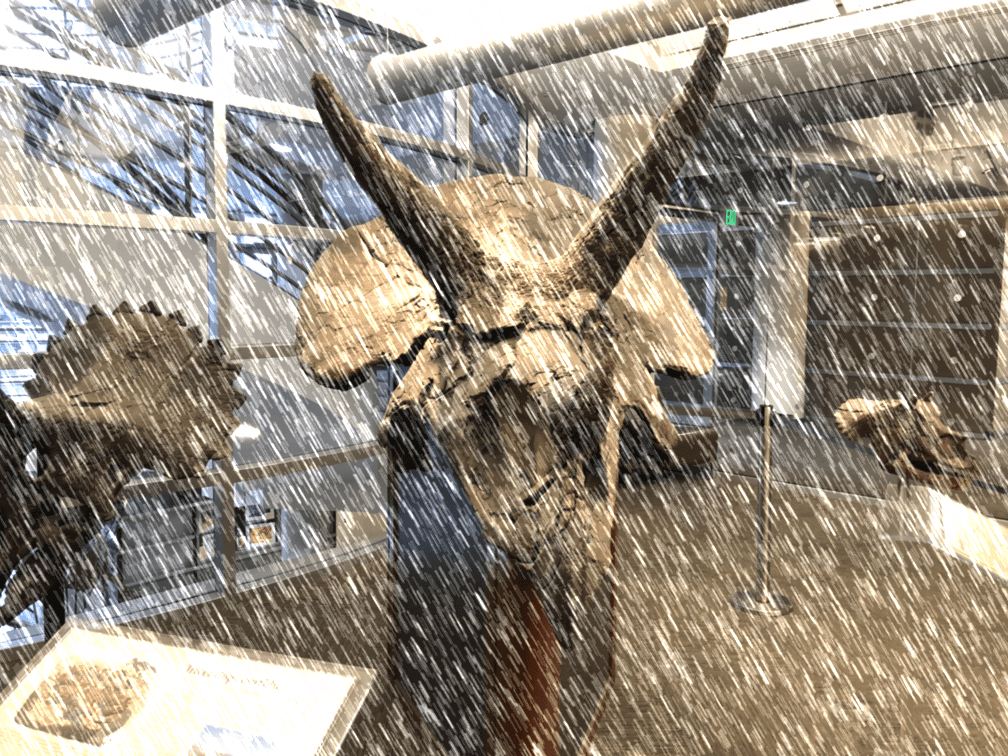}
\end{subfigure}%
\begin{subfigure}[t]{0.19\textwidth}
  \centering
  \includegraphics[width=1.0\linewidth]{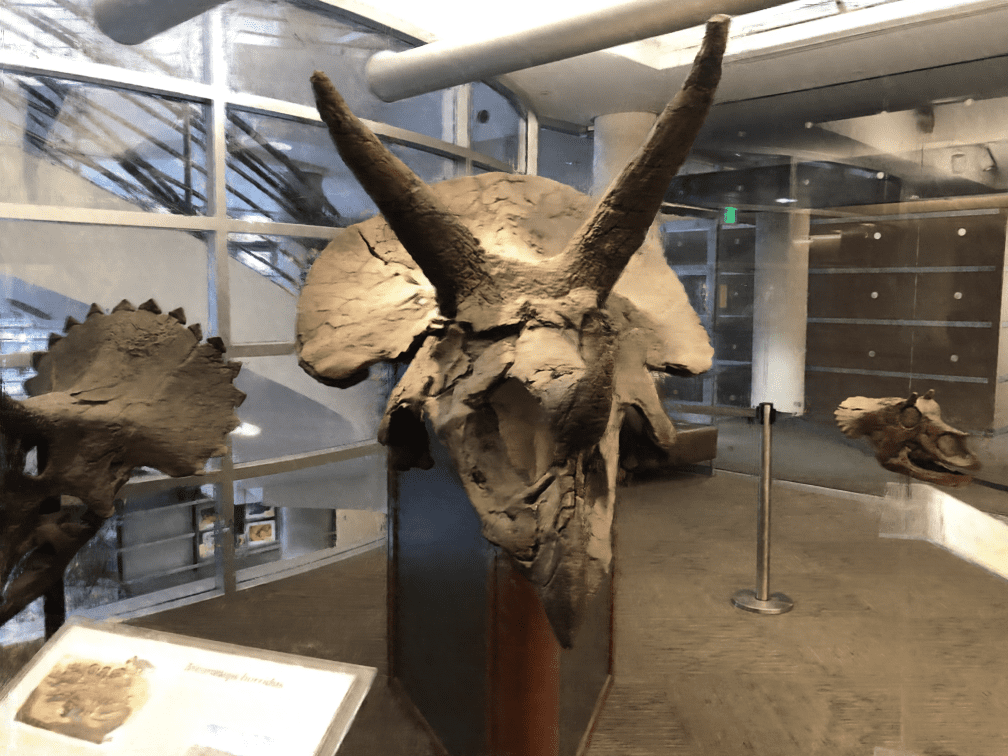}
\end{subfigure}%
\begin{subfigure}[t]{0.19\textwidth}
  \centering
  \includegraphics[width=1.0\linewidth]{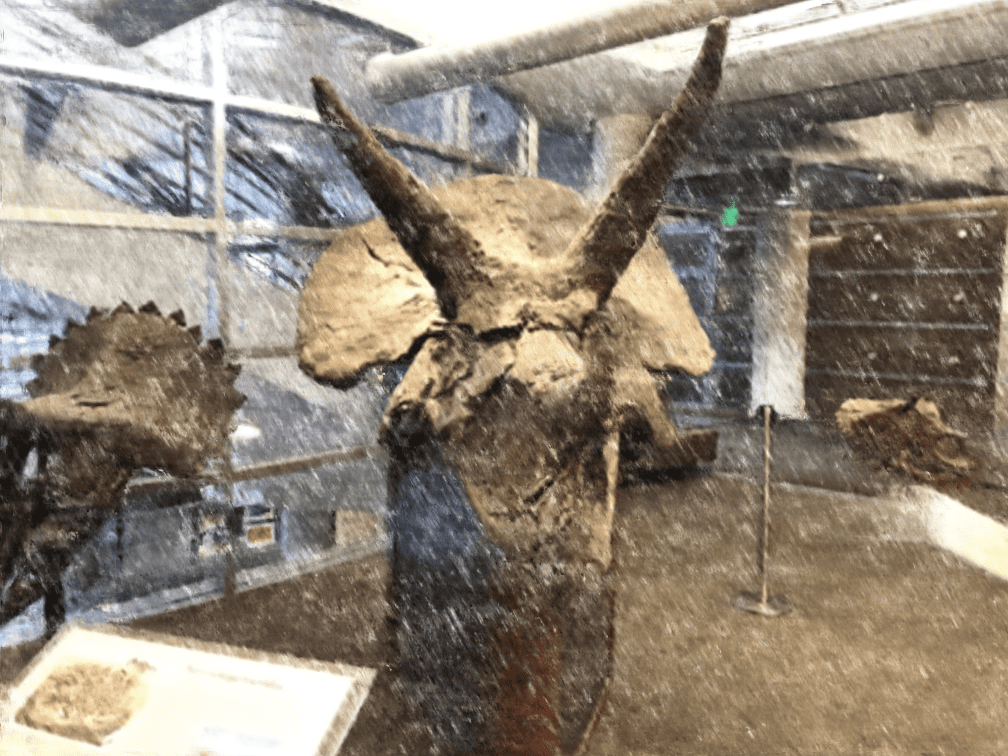}
\end{subfigure}%
\begin{subfigure}[t]{0.19\textwidth}
  \centering
  \includegraphics[width=1.0\linewidth]{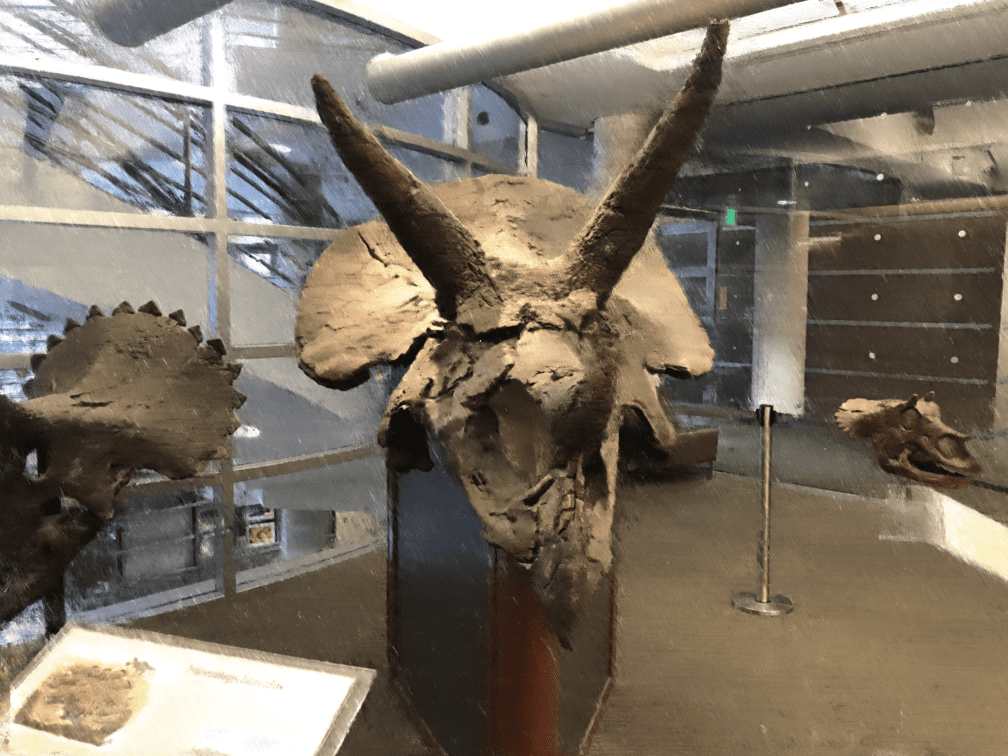}
\end{subfigure}%
\begin{subfigure}[t]{0.19\textwidth}
  \centering
  \includegraphics[width=1.0\linewidth]{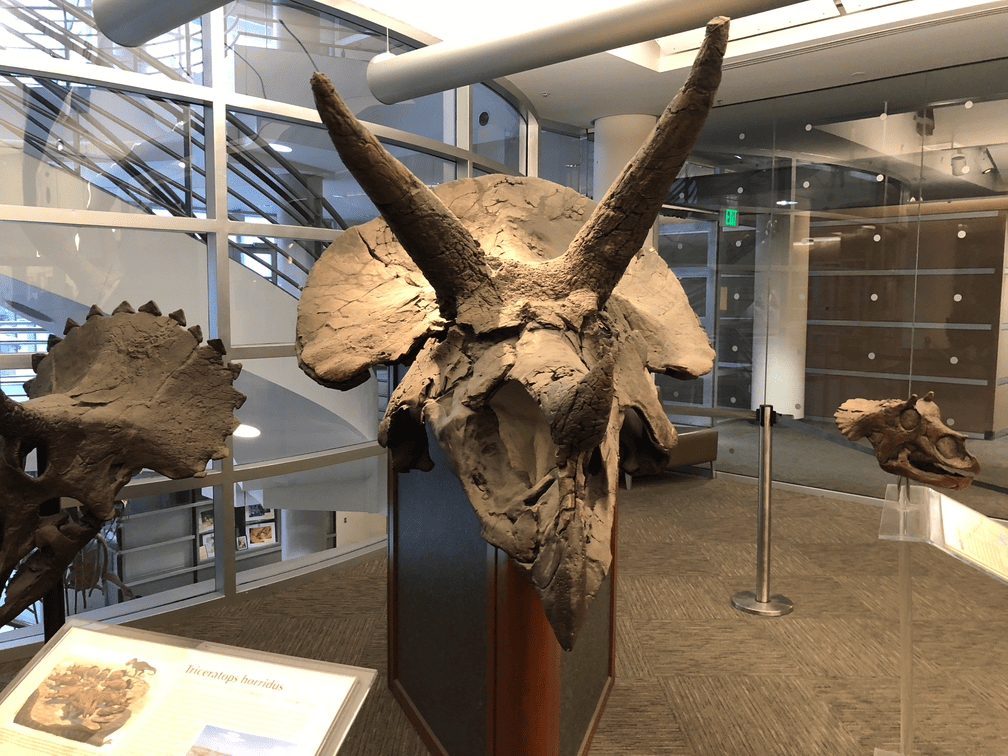}
\end{subfigure}%
\setcounter{subfigure}{0}

\rotatebox{90}{\quad\quad \tiny{Snow}}
\begin{subfigure}[t]{0.19\textwidth}
  \centering
  \includegraphics[width=1.0\linewidth]{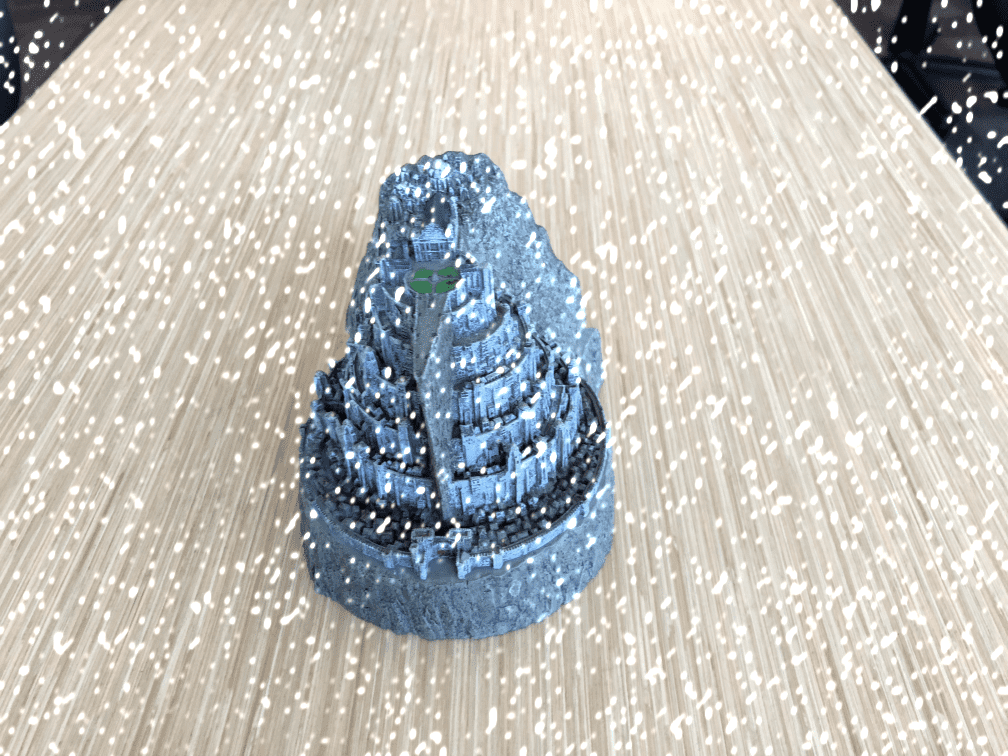}
  \caption{\tiny Corrupted View}
\end{subfigure}%
\begin{subfigure}[t]{0.19\textwidth}
  \centering
  \includegraphics[width=1.0\linewidth]{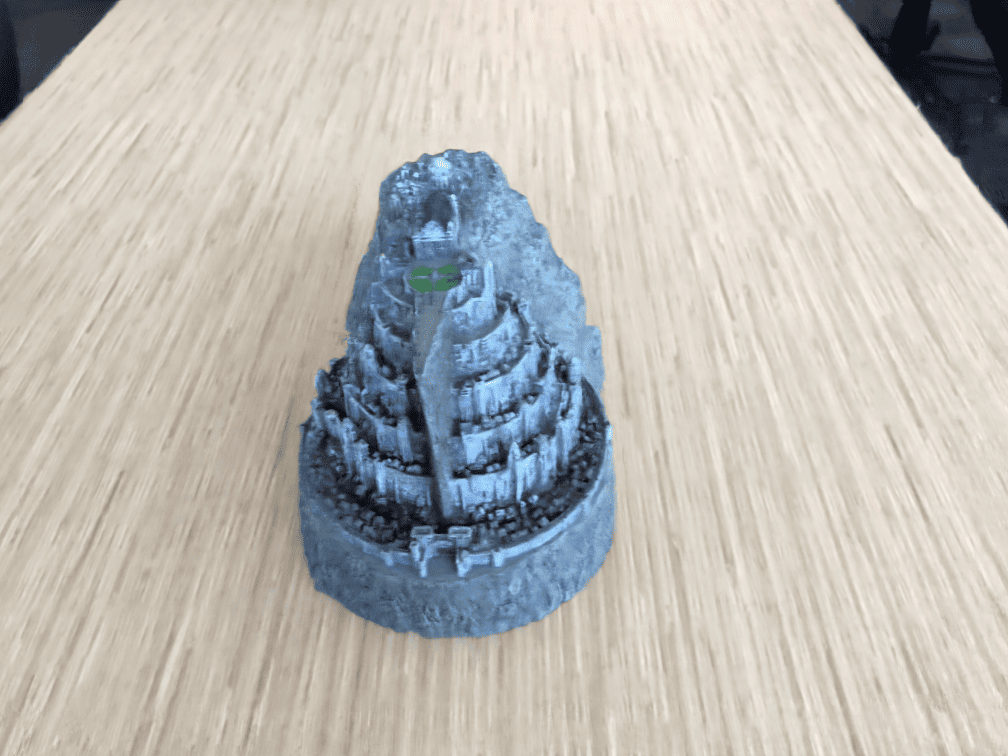}
  \caption{\tiny GNT--PromptIR}
\end{subfigure}%
\begin{subfigure}[t]{0.19\textwidth}
  \centering
  \includegraphics[width=1.0\linewidth]{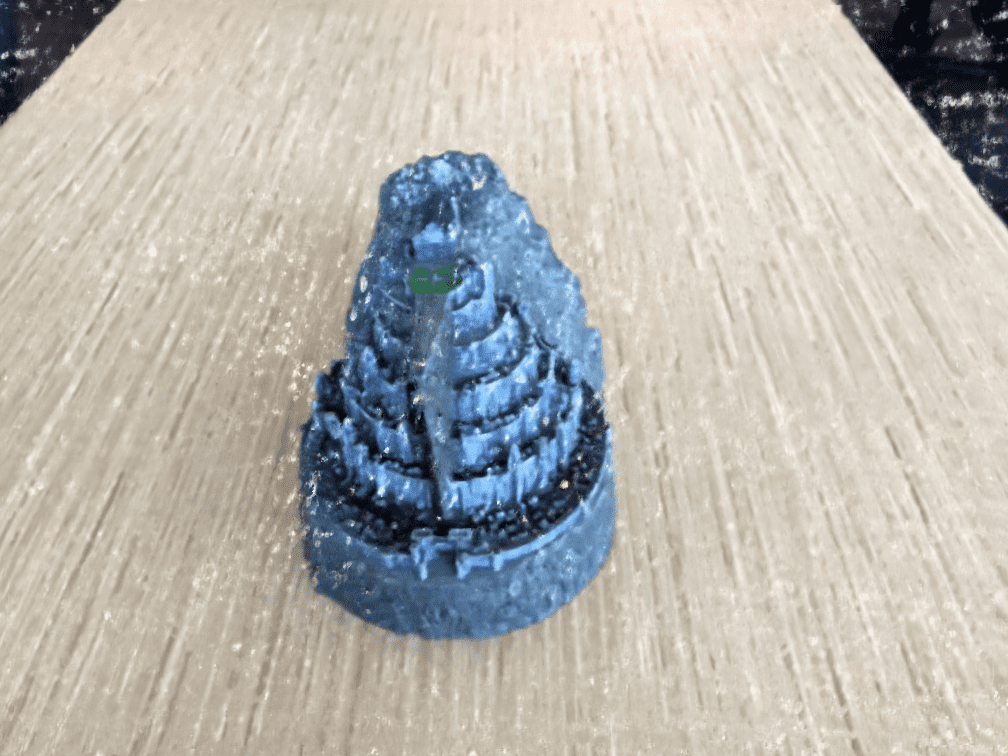}
  \caption{\tiny GNT--AutoDIR}
\end{subfigure}%
\begin{subfigure}[t]{0.19\textwidth}
  \centering
  \includegraphics[width=1.0\linewidth]{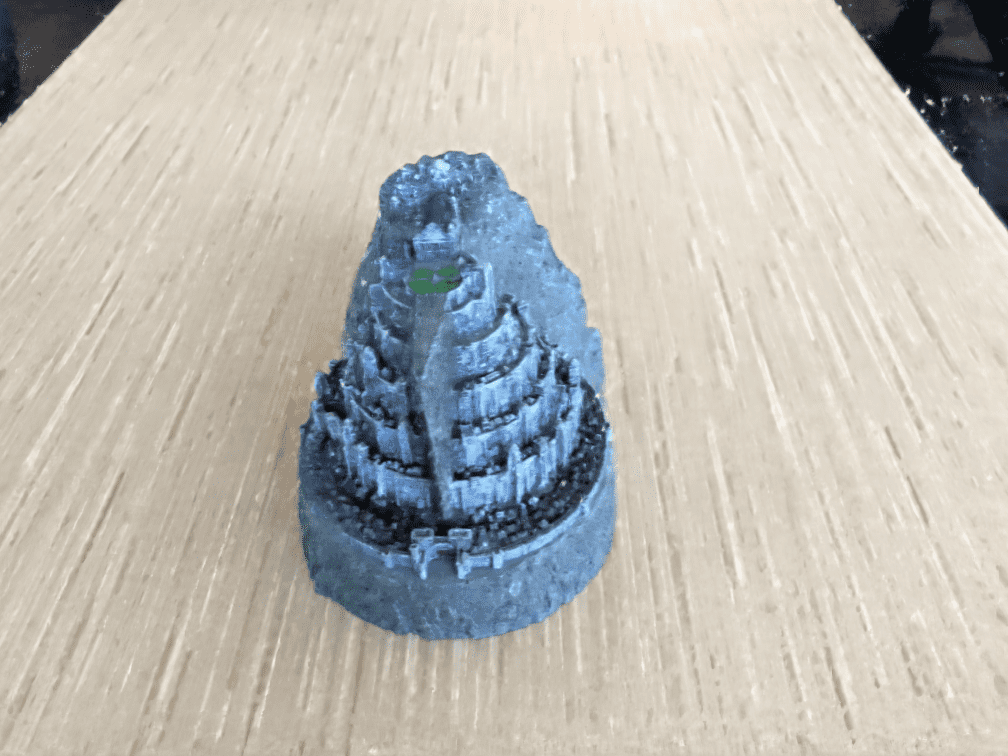}
  \caption{\tiny Ours}
\end{subfigure}%
\begin{subfigure}[t]{0.19\textwidth}
  \centering
  \includegraphics[width=1.0\linewidth]{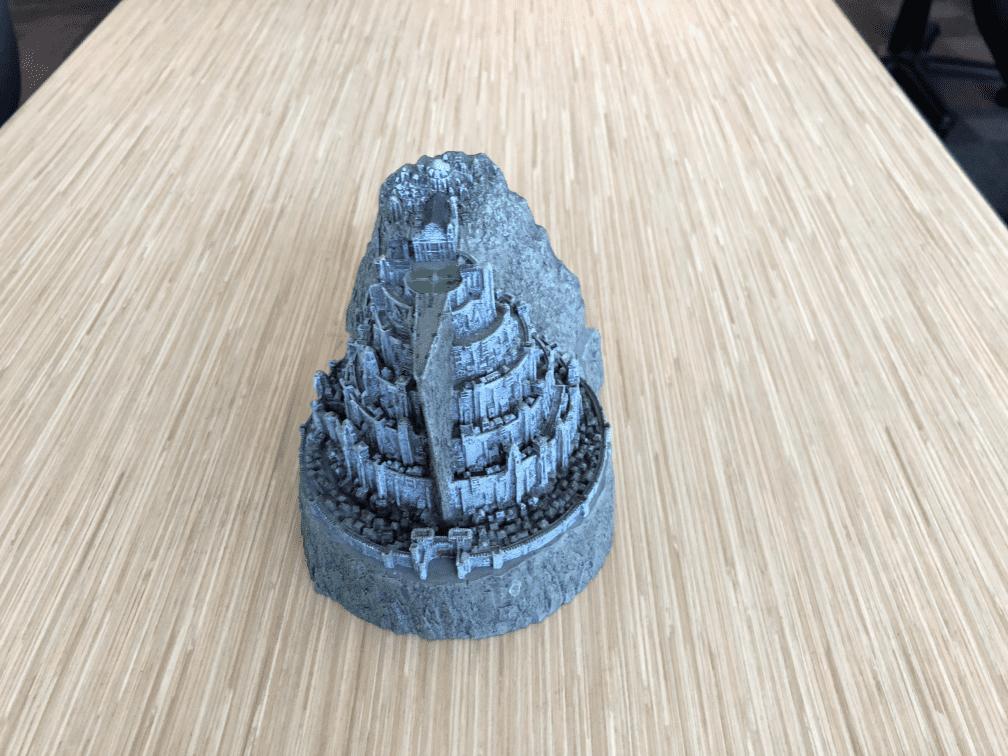}
  \caption{\tiny Ground Truth}
\end{subfigure}%

\caption{Results on the Synthetic LLFF-Corrupted data are presented, comparing against GNT-PromptIR \cite{potlapalli2023promptir} and GNT-AutoDIR \cite{jiang2023autodir}. Qualitative comparisons across five corruptions are shown, illustrating our method's consistent ability to restore the scene across all degradation types. }
\label{fig:llff-corrupted}
\vspace{-1em}
\end{figure*}

\paragraph{\textbf{LLFF-Corrupted. }} Results pertaining to LLFF-Corrupted scenes are presented in Figure \ref{fig:llff-corrupted}. We compare our method against two all-in-one approaches across the five degradations. It is evident that our method effectively restores the appearance details of the scenes while preserving their geometry simultaneously.

\section{Blind Restoration}
Despite our method's current requirement for the degradation type as input, it is feasible to extend it to achieve complete independence from the degradation type. This form of restoration, without the need for specifying the degradation type, is referred to as Blind Restoration.
To accomplish this, we propose training a convolutional network capable of taking degraded images as input and classifying the type of degradation. This network can be supervised using a cross-entropy loss function. For instance, utilizing a ResNet-18 backbone, we achieved 99.5\% accuracy in degradation type classification.
Once trained, this network can be utilized to predict the degradation type directly from input images, rendering the restoration process independent of user input regarding degradation type. 

\section{Additional Ablation Studies}
\subsection{Simple Conditioning vs ARM+DLM(Ours)}
In Table \ref{tab:ablviews}, we compare our ARM+DLM module against a simpler variant that concatenates the latent conditioning to the inputs of the cross and self-attention blocks in the view and ray transformers respectively. Our proposed modules outperform the baseline across all metrics.

\subsection{Effect of Number of Input Source Views}
In Table \ref{tab:effectofviews}, we measure the effect of the number of input views and observe a minimal drop in performance (< 4\%) with as little as 3 views. This indicates that our learned transformer modules are sufficiently robust to noisy epipolar input.

\section{Multi-View Consistency}
\begin{figure*}[t]
\centering
\captionsetup[subfigure]{labelformat=empty}

\rotatebox{90}{\quad {Snow}}
\begin{subfigure}[t]{0.19\textwidth}
  \centering
  \includegraphics[width=1.0\linewidth]{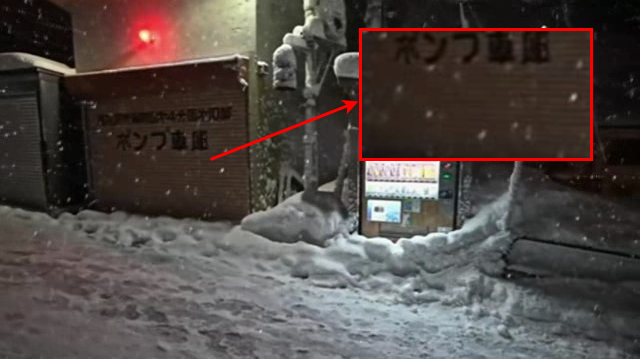}
\caption{AutoDIR (View 1)}
\end{subfigure}%
\begin{subfigure}[t]{0.19\textwidth}
  \centering
  \includegraphics[width=1.0\linewidth]{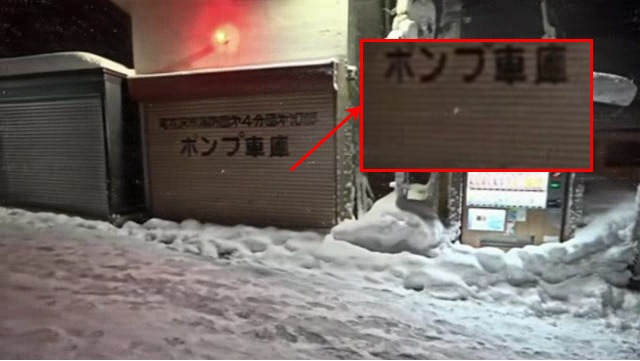}
\caption{AutoDIR (View 2)}
\end{subfigure}%
\begin{subfigure}[t]{0.19\textwidth}
  \centering
  \includegraphics[width=1.0\linewidth]{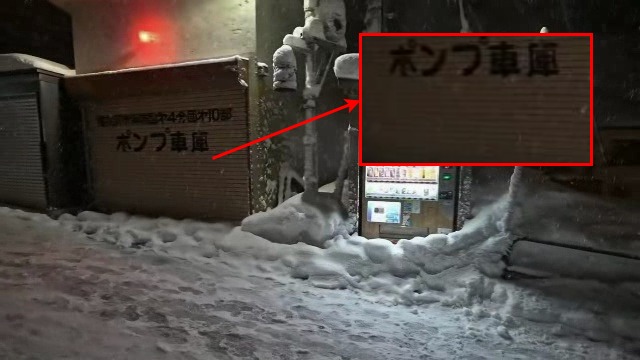}
\caption{GAURA (View 1)}
\end{subfigure}%
\begin{subfigure}[t]{0.19\textwidth}
  \centering
  \includegraphics[width=1.0\linewidth]{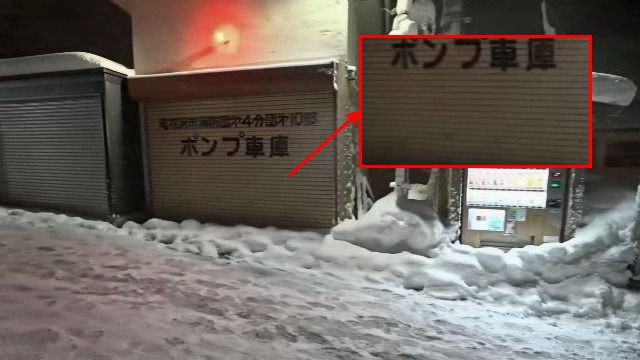}
\caption{GAURA (View 2)}
\end{subfigure}%

\rotatebox{90}{\quad {Rain}}
\begin{subfigure}[t]{0.19\textwidth}
  \centering
  \includegraphics[width=1.0\linewidth]{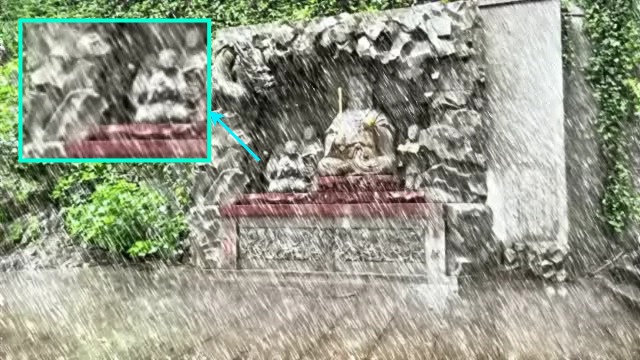}
\caption{DACLIP (View 1)}
\end{subfigure}%
\begin{subfigure}[t]{0.19\textwidth}
  \centering
  \includegraphics[width=1.0\linewidth]{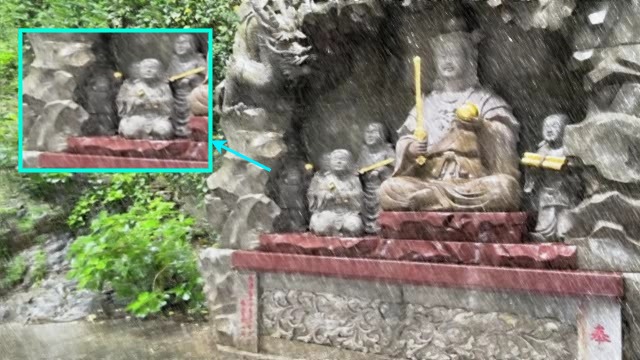}
\caption{DACLIP (View 2)}
\end{subfigure}%
\begin{subfigure}[t]{0.19\textwidth}
  \centering
  \includegraphics[width=1.0\linewidth]{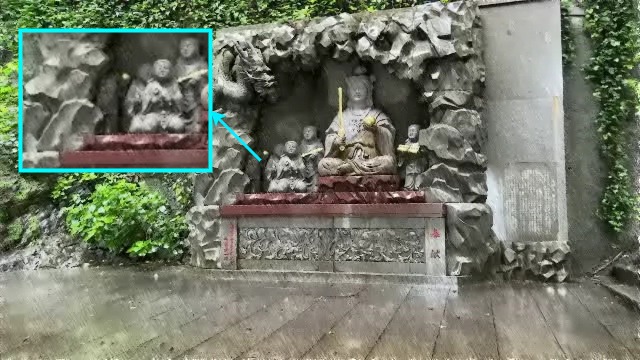}
\caption{GAURA (View 1)}
\end{subfigure}%
\begin{subfigure}[t]{0.19\textwidth}
  \centering
  \includegraphics[width=1.0\linewidth]{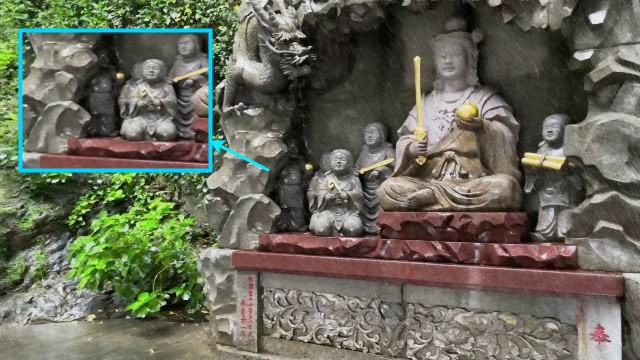}
\caption{GAURA (View 2)}
\end{subfigure}%

\caption{We compare several restoration techniques' multi-view consistency against GAURA. We observe that in the rain and snow scene, the restoration from our model results in view consistent restoration, while the baseline restores the scene inconsistently.
}
\label{fig:multiviewconsis}
\vspace{-1em}
\end{figure*}

In Table \ref{tab:viewconsis}, we present the short range and long range consistency evaluated across the generated multi-views. We see that GAURA can render view-consistent clear images from arbitrary viewing angles, superior to other baselines. Along with quantitative result, we present qualitative results in Fig. \ref{fig:multiviewconsis} which clearly shows the superiority of GAURA over other baselines in terms of multi-view consistent restoration.

\newpage

% \begin{figure*}[!ht]
%   \centering
%   \includegraphics[width=\textwidth]{figures/supplementary/gallery.jpg}
%   \caption{Below are the gallery results showcasing our method's performance on various corruptions on real-world data. Each row presents results of our method on a single corruption across two scenes. The degraded image and its corresponding restored output are visualized side-by-side. Additionally, renders from two other novel viewpoints are provided on the right. }
%   \label{fig:gallery}
%   % \vspace{-4em}
% \end{figure*}

% \begin{figure*}[!ht]
%   \centering
%   \includegraphics[width=\textwidth]{figures/supplementary/multi-corr.png}
%   \caption{Results on scenes corrupted with more than 1 degradations. We show results on scenes corrupted with Rain+Dark and Snow+Haze on the synthetically corrupted LLFF dataset. }
%   \label{fig:multi-corr}
%   % \vspace{-4em}
% \end{figure*}

\newpage

\newpage

\end{document}